\documentclass[10pt,twocolumn,letterpaper]{article} 

\usepackage{iccv}

\usepackage{times}
\usepackage{relsize}
\usepackage[T1]{fontenc} 
\usepackage[latin1]{inputenc} 
\usepackage[english]{babel}
\usepackage{ragged2e}

\usepackage{epsfig}
\usepackage{graphicx}
\usepackage{wrapfig}
\usepackage[belowskip=-7pt,aboveskip=5pt,font=small]{caption}
\usepackage[belowskip=-2pt,aboveskip=0pt,font=small]{subcaption}
\usepackage{amsmath, amsthm, amssymb}
\usepackage{textcomp}
\usepackage{stmaryrd}
\usepackage{upgreek}
\usepackage{bm}
\usepackage{cases}
\usepackage{mathtools}

\usepackage{cite}
\usepackage[pagebackref=true,breaklinks=true,letterpaper=true,colorlinks,bookmarks=false,citecolor=red]{hyperref}
\usepackage{bibspacing}
\usepackage{algorithm}
\usepackage{algpseudocode}

\usepackage{multirow}
\usepackage{rotating}
\usepackage{booktabs}
\usepackage{tabularx}

\usepackage{enumitem}
\usepackage[olditem,oldenum]{paralist}

\usepackage{alltt}
\usepackage{listings}

\belowdisplayskip \abovedisplayskip

\newlength{\sectionReduceTop}
\newlength{\sectionReduceBot}
\newlength{\subsectionReduceTop}
\newlength{\subsectionReduceBot}
\newlength{\abstractReduceTop}
\newlength{\abstractReduceBot}
\newlength{\captionReduceTop}
\newlength{\captionReduceBot}
\newlength{\subsubsectionReduceTop}
\newlength{\subsubsectionReduceBot}

\newlength{\eqnReduceTop}
\newlength{\eqnReduceBot}

\newlength{\horSkip}
\newlength{\verSkip}

\newlength{\figureHeight}
\usepackage{mysymbols}

\usepackage{url}
\usepackage{xspace}
\usepackage{comment}
\usepackage{color}
\usepackage{afterpage}
\usepackage{pdfpages}
\usepackage{framed}
\usepackage{fancybox}
\usepackage[normalem]{ulem}
\usepackage[author={DB},open=false]{pdfcomment}

\newcommand{\com}[1]{}

\iccvfinalcopy 
\def\iccvPaperID{861} 

\ificcvfinal\fi

\begin{document}

\title{Object-Proposal Evaluation Protocol is `Gameable'}

\author{Neelima Chavali\qquad Harsh Agrawal\qquad  Aroma Mahendru\qquad  Dhruv Batra\\
    Virginia Tech\\
    {\tt\small \{gneelima, harsh92, aroma.mahendru, dbatra\}@vt.edu}
}

\maketitle

\vspace{\abstractReduceTop}
\begin{abstract} 
	\vspace{\abstractReduceBot}
	Object proposals have quickly become the de-facto pre-processing step in a number of vision pipelines (for object detection, object discovery, and other tasks). Their performance is usually evaluated on partially annotated datasets. In this paper, we argue that the 
	choice of using a partially annotated dataset for evaluation of object proposals is problematic -- as we demonstrate via a thought experiment, the evaluation protocol is `gameable', in the sense that progress under this protocol does not necessarily correspond to a ``better'' category independent object proposal algorithm.
	
	To alleviate this problem, we: (1) Introduce a nearly-fully annotated version of PASCAL VOC dataset, which serves as a test-bed to check if object proposal techniques are overfitting to a particular list of categories. (2) Perform an exhaustive evaluation of object proposal methods on our introduced nearly-fully annotated PASCAL dataset and perform cross-dataset generalization experiments; and (3) Introduce a diagnostic experiment to detect the \emph{bias capacity} in an object proposal algorithm. This tool circumvents the need to collect a densely annotated dataset, which can be expensive and cumbersome to collect. Finally, we plan to release an easy-to-use toolbox which combines various publicly 
	available implementations of object proposal algorithms which standardizes the proposal generation and evaluation so that new methods can be added and evaluated on different datasets.
	We hope that the results presented in the paper will motivate the community to test the category independence of various object proposal methods by carefully choosing the evaluation protocol.
	
\end{abstract} 


\vspace{-0.5cm}
\vspace{\sectionReduceTop}
\section{Introduction}
\label{sec:intro}
\vspace{\sectionReduceBot}
\begin{figure*}[t]
	\centering
	\begin{subfigure}[b]{0.32\textwidth}
		\includegraphics[width=1\columnwidth]{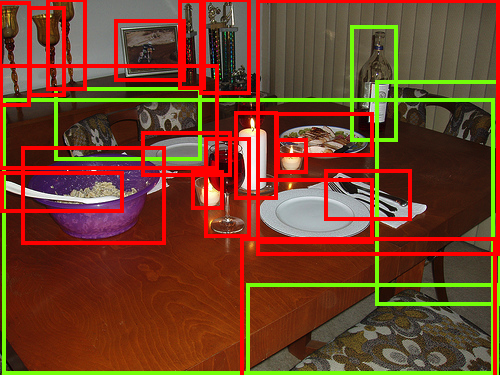}
		\subcaption{(Green) Annotated, (Red) Unannotated
			\label{fig:1}}
	\end{subfigure} 
	\,
	\begin{subfigure}[b]{0.32\textwidth}
		\includegraphics[width=1\columnwidth]{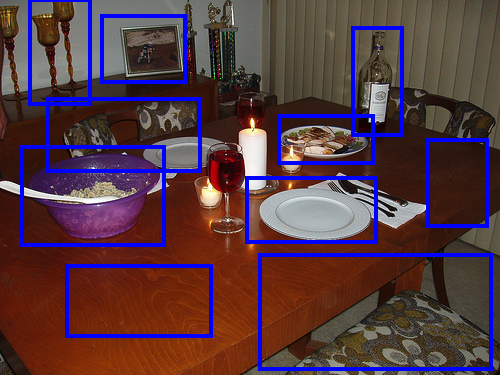}
		\subcaption{Method 1 with recall 0.6
			\label{fig:2}}
	\end{subfigure} 
	\,
	\begin{subfigure}[b]{0.32\textwidth}
		\includegraphics[width=1\columnwidth]{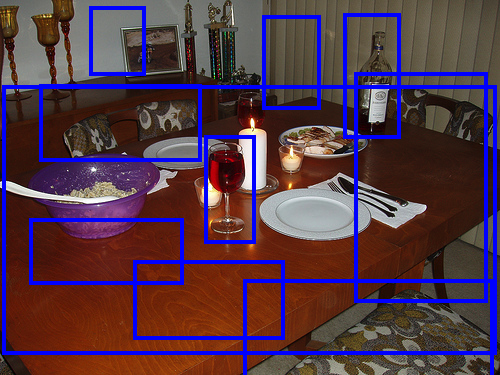}
		\subcaption{Method 2 with recall 1
			\label{fig:3}}
	\end{subfigure}
	\caption{(a) shows PASCAL annotations natively present in the dataset in green. Other objects that are not annotated but present in the image are shown in red; (b) shows Method 1 and (c) shows Method 2. Method 1 visually seems to recall more categories such as plates, glasses, \etc that Method 2 missed. Despite that, 
		the computed recall for Method 2 is higher because it recalled all instances of PASCAL 
		categories that were present in the ground truth.  Note that the number of proposals 
		generated by both methods is equal in this figure. 
		\label{fig:qualfig1}}
\end{figure*}

\begin{figure*}[t]
	\centering
	\begin{subfigure}[b]{0.32\textwidth}
		\includegraphics[width=1\columnwidth]{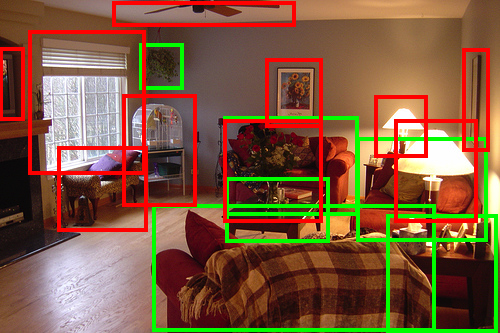}
		\subcaption{(Green) Annotated, (Red) Unannotated}
		\label{fig:1}
	\end{subfigure} 
	\,
	\begin{subfigure}[b]{0.32\textwidth}
		\includegraphics[width=1\columnwidth]{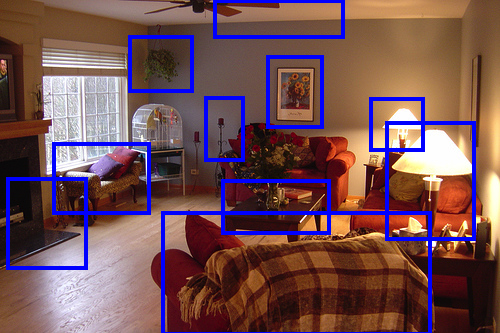}
		\subcaption{Method 1 with recall 0.5}
		\label{fig:2}
	\end{subfigure} 
	\,
	\begin{subfigure}[b]{0.32\textwidth}
		\includegraphics[width=1\columnwidth]{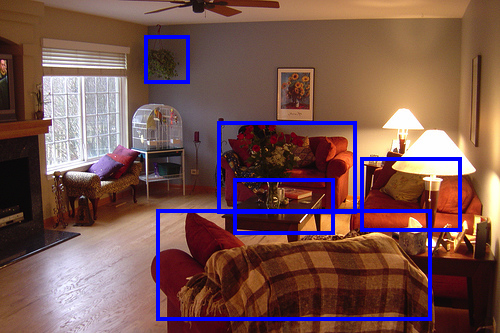}
		\subcaption{Method 2 with recall 0.83}
		\label{fig:3}
	\end{subfigure}
	\caption{(a) shows PASCAL annotations natively present in the dataset in green. Other objects that are not annotated but present in the image are shown in red; (b) shows Method 1 and (c) shows Method 2. 
		Method 1 visually seems to recall more categories such as lamps, picture, \etc that Method 2 
		missed. Clearly the recall for Method 1 \emph{should} be higher. 
		However, the calculated recall for Method 2 is significantly higher, 
		which is counter-intuitive. This is because Method 2 recalls more PASCAL category objects. 
		\label{fig:qualfig2}}
\end{figure*}
In the last few years, the Computer Vision community has witnessed the emergence of a new class 
of techniques called \textit{Object Proposal} algorithms \cite{ZitnickECCV14,EndresPAMI14,Arbelaez_CVPR14,AlexePAMI12,RahtuICCV11,ManenICCV13,RantalankilaCVPR14,UijlingsIJCV13,HumayunCVPR14,cheng2014bing,DBLP:conf/eccv/KrahenbuhlK14}. 

Object proposals 
are a set of candidate regions or bounding boxes in an image that may potentially contain an object.  

Object proposal algorithms have quickly become the de-facto pre-processing step 
in a number of vision pipelines --  
object detection\cite{DBLP:journals/corr/GirshickDDM13,DBLP:journals/corr/HeZR014,DBLP:journals/corr/SzegedyREA14,Wang_2013_ICCV,cinbis:hal-00873134,DBLP:journals/corr/ErhanSTA13,Kuznetsova_2015_CVPR,Tsai_2015_CVPR,Zhu_2015_CVPR,OOP_ICCV2015},
segmentation\cite{Carreira2012,conf/cvpr/ArbelaezHGGBM12,conf/eccv/CarreiraCBS12,conf/cvpr/DaiH015,Zhang_2015_CVPR},
object discovery\cite{deselaers2012weakly,DBLP:journals/corr/ChoKSP15, Kading_2015_CVPR,conf/cvpr/RubinsteinJKL13},
weakly supervised learning of object-object interactions\cite{cinbis2012contextual, prest2012weakly},
content aware media re-targeting\cite{sun2011scale}, action recognition in still images\cite{sener2012recognizing} and visual tracking\cite{DBLP:journals/corr/WangLGY15,DBLP:journals/corr/KwakCLPS15}.
Of all these tasks, object proposals have been particularly successful in object detection systems. 
For example, \emph{nearly all top-performing entries}\cite{szegedy2014going, lin2013network, ouyang2014deepid, DBLP:journals/corr/HeZR014} in the ImageNet 
Detection Challenge 2014 {\cite{ILSVRCarxiv14} used object proposals.
They are preferred over the formerly used sliding window paradigm due to their computational efficiency. Objects present in an image may vary in location, size, and aspect ratio. Performing an exhaustive search over such a high dimensional space is difficult. 
By using object proposals, computational effort can be focused on a small number of candidate windows. 

The focus of this paper is the protocol used for evaluating object proposals. Let us begin by asking -- \emph{what is the purpose of an object proposal algorithm?} 

In early works \cite{AlexePAMI12,EndresPAMI14,ManenICCV13}, the emphasis was on \emph{category independent object proposals}, where the goal is to identify instances of \emph{all} objects in the image irrespective of their category. While it can be tricky to precisely define what an ``object'' is\footnote{Most category independent object proposal methods define an object as ``stand-alone thing with a well-defined closed-boundary''. For ``thing" \vs ``stuff'' discussion, see \cite{HeitzKoller:ECCV08}.}, these early works presented cross-category evaluations to establish and measure category independence.

More recently, object proposals are increasingly viewed as \emph{detection proposals} {\cite{DBLP:conf/eccv/KrahenbuhlK14,UijlingsIJCV13,ZitnickECCV14,Krahenbuhl_2015_CVPR} where the goal is to improve the object detection pipeline, focusing on a chosen set of object classes (\eg \textasciitilde20 PASCAL categories). In fact, many modern proposal methods are learning-based\cite{cheng2014bing,DBLP:conf/eccv/KrahenbuhlK14,HumayunCVPR14,Krahenbuhl_2015_CVPR,DBLP:journals/pami/KangHEK15,DBLP:journals/corr/KuoHM15, deepprop,DBLP:journals/corr/PinheiroCD15} where the definition of an ``object'' is the set of annotated classes in the dataset. This increasingly blurs the boundary between a proposal algorithm and a detector.

Notice that the former definition has an emphasis on object discovery\cite{deselaers2012weakly,DBLP:journals/corr/ChoKSP15,conf/cvpr/RubinsteinJKL13}, while the latter definition emphasises on the ultimate performance of a detection pipeline. Surprisingly, despite the two different goals of `object proposal,' there exists only a single evaluation protocol\com{, which is the following}: 
\begin{compactenum}
	\item Generate proposals on a dataset: The most commonly used dataset for evaluation today is 
	the PASCAL VOC \cite{pascal-voc-2007} detection set. 
	Note that this is a \textit{partially annotated} dataset where 
	only the 20 PASCAL category instances are annotated. 
	
	\item Measure the performance of the generated proposals: typically in terms of `recall' 
	of the annotated instances. 
	Commonly used metrics are described in \secref{sec:eval}. 
\end{compactenum}

The central thesis of this paper is that the current evaluation protocol for object proposal methods is suitable for object detection pipeline but is a \emph{`gameable' and misleading protocol} for category independent tasks. By  evaluating only on a specific set of object categories, we fail to capture 
the performance of the proposal algorithms on \textit{all the remaining object categories that are present in the test set, but not annotated in the ground truth}. 

Figs.~\ref{fig:qualfig1}, \ref{fig:qualfig2} illustrate this idea on images from PASCAL VOC 2010. 
Column (a) shows the ground-truth object annotations (in green, the annotations natively present 
in the dataset for the 20 PASCAL categories --`chairs', `tables', `bottles', \etc; in red, the annotations that we added to the dataset by 
marking object such as `ceiling fan', `table lamp', `window', \etc originally annotated `background' in the dataset). 
Columns (b) and (c) show the outputs of two object proposal methods. Top row shows the case when both methods produce the same number of proposals; 
bottom row shows unequal number of proposals. 
We can see that proposal method in Column (b) seems to be more ``complete'', in the sense that it recalls or discovers a large number of instances. 
For instance, in the top row it detects a number of non-PASCAL  categories (`plate', `bowl', `picture frame', \etc) but misses out on finding the PASCAL category `table'. 
In both rows, the method 
in Column (c) is reported as achieving a higher recall, \emph{even in the bottom row, when it recalls strictly fewer objects, not just different ones}. The reason is that Column (c) recalls/discovers instances of the 20 PASCAL 
categories, which are the only ones annotated in the dataset. Thus, Method 2 appears to be a \emph{better} object proposal generator simply because it focuses on the annotated categories in the dataset.

While intuitive (and somewhat obvious) in hindsight, 
we believe this is \com{an important}a crucial finding because it makes the current protocol \textit{`gameable'} or susceptible  
to manipulation (both intentional and unintentional)  and misleading for measuring improvement in category independent object proposals.

Some might argue that if the end task is to detect a certain set of  categories (20 PASCAL or 80 COCO categories) then it is enough to evaluate on them and there is no need to care about other categories which are not annotated in the dataset. We agree, but it is important to keep in mind that object detection is not the only application of object proposals. There are other tasks for which it is important for proposal methods to generate category independent proposals. For example, in semi/unsupervised object localization\cite{deselaers2012weakly,DBLP:journals/corr/ChoKSP15, Kading_2015_CVPR,conf/cvpr/RubinsteinJKL13}  the goal is to identify all the objects in a given image that contains many object classes without any specific target classes. In this problem, there are no image-level annotations, an assumption of a single dominant class, or even a known number of object classes\cite{DBLP:journals/corr/ChoKSP15}. Thus, in such a setting, using a proposal method that has tuned itself to 20 PASCAL objects would not be ideal -- in the worst case, we may not discover any new objects. As mentioned earlier, there are many such scenarios including learning object-object interactions\cite{cinbis2012contextual, prest2012weakly}, content aware media re-targeting\cite{sun2011scale}, visual tracking\cite{DBLP:journals/corr/KwakCLPS15}, \etc


To summarize, the contributions of this paper are: 
\begin{compactitem}

	\item We report the `gameability' of the current object proposal evaluation protocol.
	
	\item 
	We demonstrate this `gameability' via a simple thought experiment 
	where we propose a `fraudulent' object proposal method that 
	\emph{significantly outperforms all existing object proposal techniques} on current metrics, but would under any no circumstances be considered a category independent proposal technique. As a side contribution of our work, we present a simple technique for producing state-of-art object proposals.
	
	\item After establishing the problem, we propose three ways of improving the current evaluation protocol 
	to measure the category independence of object proposals: 
	\begin{compactenum}
		\item evaluation on \emph{fully} annotated datasets,
		\item cross-dataset evaluation on \emph{densely} annotated datasets.
		\item a new evaluation metric that quantifies the \emph{bias capacity}
		of proposal generators.
	\end{compactenum} 
		For the first test, we introduce a nearly-fully annotated PASCAL VOC 2010 where we 
		annotated \emph{all instances of all object categories} occurring in the images. 
		
	\item We thoroughly evaluate existing proposal methods on this nearly-fully and two densely annotated datasets. 
	
	
	\item We will release all code and data for experiments, and an object proposals library that allows 
	for easy comparison of all popular object proposal techniques. 
\end{compactitem}

\vspace{\sectionReduceTop}
\section{Related Work}
\label{sec:related}
\vspace{\sectionReduceBot}


\textbf{Types of Object Proposals:}
Object proposals can be broadly categorized into two categories: 
\begin{compactitem}
\item \textbf{Window scoring}: In these methods, the space of all possible windows in an 
image is sampled to get a subset of the windows (\eg, via sliding window). 
These windows are then scored for the 
presence of an object based on the image features from the windows. The algorithms that fall 
under this category are \cite{AlexePAMI12, RahtuICCV11, ZitnickECCV14, cheng2014bing,Chen_2015_CVPR,deepprop}. 

\item \textbf{Segment based}: These algorithms involve over-segmenting an image and 
merging the segments using some strategy. These methods include 
\cite{UijlingsIJCV13, RantalankilaCVPR14, Arbelaez_CVPR14, DBLP:conf/eccv/KrahenbuhlK14, 
	HumayunCVPR14, EndresPAMI14, ManenICCV13,Wang_2015_CVPR,DBLP:journals/corr/KuoHM15,DBLP:journals/corr/PinheiroCD15}. The generated region proposals can be converted to bounding boxes if needed.
\end{compactitem}

\textbf{Beyond RGB proposals:} 
Beyond the ones listed above, a wide variety of algorithms fall under the umbrella of `object proposals'. 
For instance, \cite{oneata2014spatio, fragkiadaki2014spatio,Yu_2015_CVPR,Misra_2015_CVPR,Wu_2015_CVPR} used 
spatio-temporal object proposals for action recognition, segmentation and tracking in videos. 
 Another direction of work\cite{Gupta2014,banica,Banica_2015_CVPR} explores use of RGB-D cuboid proposals in an object detection 
and semantic segmentation in RGB-D images. 
While the scope of this paper is limited to proposals in RGB images, the central 
thesis of the paper (\ie, gameability of the evaluation protocol) is broadly applicable to other settings. 

\textbf{Evaluating Proposals:} 
\com{While a variety of approaches have been proposed for generating object proposals, 
t}There has been a relatively limited analysis and evaluation of proposal methods or the proposal evaluation protocol.
Hosang~\etal~\cite{Hosang2014} 
focus on evaluation of object proposal algorithms, 
in particular the stability of such algorithms on parameter changes and image perturbations. 
Their works shows that a large number of category independent proposal algorithms indeed generalize well to non-PASCAL categories, 
for instance in the ImageNet 200 category detection dataset~\cite{ILSVRCarxiv14}. Although these findings are important (and consistent with our experiments), they are unrelated to the `gameability' of the evaluation protocol, which is our focus.
In ~\cite{Hosang2015}, authors present an analysis of various proposal methods regarding proposal repeatability, ground truth annotation recall, and their impact on detection performance. They also introduced a new evaluation metric (Average Recall). 
Their argument for a new metric is the need for a better localization between generated proposals 
and ground truth. While this is a valid and significant concern, it is orthogonal 
to the`gameability' of the evaluation protocol, which to the best of our knowledge 
has not been previously addressed. Another recent related work perhaps is \cite{PTVG2015}, which analyzes the state-of-the-art  methods in segment-based
object proposals, focusing on the challenges faced
when going from PASCAL VOC to MS COCO. They also
analyze how aligned the proposal methods are with the bias
observed in MS COCO towards small objects and the center
of the image and propose a method to boost their performance. Although there is a discussion about biases in datasets  but it is unlike our theme, which is `gameability' due to these biases.
As stated earlier, while  early papers \cite{AlexePAMI12,EndresPAMI14,ManenICCV13} reported cross-dataset or cross-category generalization experiments similar to ones reported in this paper, with the trend of learning-based proposal methods, these experiments  and concerns seem to have fallen out of standard practice, which we show is problematic.
\vspace{-0.25cm}
\vspace{\sectionReduceTop}
\section{Evaluating Object Proposals}
\label{sec:eval}
\vspace{\sectionReduceBot}
%

Before we describe our evaluation and analysis, let us first look at the object proposal evaluation 
protocol that is widely used today. 
The following two factors are involved: 
\begin{compactenum}
	\item \textbf{Evaluation Metric}: 
	The metrics used for evaluating object proposals are all typically functions of 
	intersection over union (IOU) (or Jaccard Index) between generated proposals and ground-truth annotations.  For two boxes/regions $b_i$ and $b_j$, IOU is defined as:
	\begin{equation}
	\text{IOU}(b_i, b_j)=\frac{area(b_i \cap b_j)}{area(b_i \cup b_j)}
	\end{equation}
	The following metrics are commonly used: 
	%
	\begin{asparaitem}
		\item \textbf{Recall $@$ IOU Threshold $t$}: 
		For each ground-truth instance, this metric checks whether the `best' proposal from list $L$
		has IOU greater than a threshold $t$. If so, this ground truth instance is considered `detected' 
		or `recalled'. 
		Then average recall is measured over all the ground truth instances:
		%
		%
		\begin{equation}
		\text{Recall }@ \, t=\frac{1}{|G|} \sum\limits_{g_i \in G}\text{I }[{\max_{l_j \in L} \, \text{IOU}}(g_i,l_j) > t],
		\end{equation}
		where $I[\cdot]$ is an indicator function for the logical preposition in the argument.
		Object proposals are evaluated using this metric in two ways: 
		\begin{compactitem}
		\item plotting Recall-\vs-\#proposals by fixing $t$
		\item plotting Recall-\vs-$t$ by fixing the \#proposals in $L$.
		\end{compactitem} 
		
		\vspace{3pt}
		\item \textbf {Area Under the recall Curve (AUC)}: 
		 AUC summarizes the area under the Recall-\vs-\#proposals plot  for different values of $t$ in a single plot. This metric  measures AUC-\vs-\#proposals. It is also plotted by varying \#proposals in $L$ and plotting AUC-vs-$t$.
		\item \textbf{Volume Under Surface (VUS)}: This measures the average recall by linearly varying $t$  and varying the \#proposals in $L$ on either linear or log scale. Thus it merges both kinds of AUC plots into one.
		\item \textbf {Average Best Overlap (ABO)}: 
		This metric eliminates the need for a threshold. 
		We first calculate the overlap between each ground truth annotation $g_i$ $\in$ $G$, and the `best' object hypotheses in $L$. ABO is calculated as the average:
		\begin{equation}
		\text{ABO}=\frac{1}{|G|} \sum\limits_{g_i \in G}{\max_{l_j \in L} \, \text{IOU}}(g_i,l_j)
		\end{equation}
		ABO is typically is calculated on a per class basis. 
		Mean Average Best Overlap (MABO) is  defined as the mean ABO over all classes.
		\item \textbf{Average Recall (AR)}: This metric was recently introduced in \cite{Hosang2015}. 
		Here, average recall (for IOU between 0.5 to 1)-\vs-\#proposals in $L$ is plotted. 
		AR also summarizes proposal performance across different values of $t$. 
		AR was shown to correlate with ultimate detection performance better than other metrics. 
	\end{asparaitem}
	\item \textbf{Dataset}:  
	The most commonly used datasets are the  
	the PASCAL VOC \cite{pascal-voc-2007} detection datasets. 
	Note that these are \textit{partially annotated} datasets where 
	only the 20 PASCAL category instances are annotated. 
	Recently analyses have been shown on ImageNet \cite{Hosang2014Bmvc}, 
	which has more categories annotated than PASCAL, but is still a partially annotated dataset. 
\end{compactenum}
\label{sec:Eval_protocol}
\vspace{-3mm}
\vspace{\sectionReduceTop}
\section{ A Thought Experiment: \\ How to Game the Evaluation Protocol}
\label{sec:thought}
\vspace{\sectionReduceBot}

Let us conduct a thought experiment to demonstrate that the object proposal evaluation 
protocol can be `gamed'. 

Imagine yourself reviewing a paper claiming to introduce a new object proposal method -- 
called DMP. 

Before we divulge the details of DMP, consider the performance of DMP shown in 
\figref{pascal2010auc_tp} on the PASCAL VOC 2010 dataset, under the AUC-\vs-\#proposals metric. 

\begin{figure}[H]
	\centering
	\includegraphics[width=0.8\columnwidth]{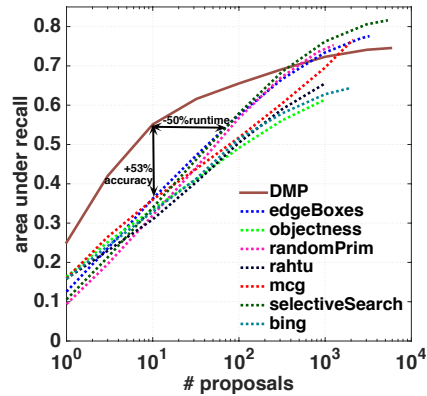}
	\caption{Performance of different object proposal methods (dashed lines) and our proposed `fraudulent' 
		method (DMP) on the PASCAL VOC 2010 dataset. 
		We can see that DMP \emph{significantly} outperforms all other proposal generators.
		See text for details. 
	\label{pascal2010auc_tp}}
\end{figure}
%
As we can clearly see, the proposed method DMP \emph{significantly} exceeds all 
existing proposal methods \cite{ZitnickECCV14,EndresPAMI14,Arbelaez_CVPR14,AlexePAMI12,RahtuICCV11,ManenICCV13,UijlingsIJCV13,cheng2014bing,DBLP:conf/eccv/KrahenbuhlK14} (which seem to have little variation over one another). The improvement\com{ in AUC} at some points in the curve (\eg, at M=10) seems to be \emph{an order of magnitude} larger than all previous incremental improvements reported in the literature!
In addition to the \com{improvement}gain in AUC at a fixed M, DMPs also achieves the same AUC (0.55) at an \emph{order of magnitude fewer} number of proposals (M=10 \vs M=~50 for edgeBoxes\cite{ZitnickECCV14}). Thus, fewer proposals need to be processed by the \com{downstream}ensuing detection system, resulting in an equivalent run-time speedup. This seems to indicate that a significant \com{advancement}progress has been made in the field of generating object proposals.

So what is our proposed state-of-art technique DMP? \\
It is a mixture-of-experts model, 
consisting of 20 experts, where each expert is a deep feature (fc7)-based~\cite{donahue_arxiv13} 
objectness detector. At this point, you, the savvy reader, are probably already 
beginning to guess what we did.

DMP stands for `Detector Masquerading as Proposal generator'. We trained object detectors 
for the 20 PASCAL categories (in this case with RCNN\cite{DBLP:journals/corr/GirshickDDM13}), 
and then used these 20 detectors to produce the top-M most confident detections (after NMS), and declared 
them to be `object proposals'. 

The point of this experiment is to demonstrate the following fact -- 
clearly, no one would consider a collection of 20 object detectors to be a category independent object proposal method. However, our existing evaluation protocol declared the union of these top-M detections to be state-of-the-art.

Why did this happen? Because the protocol today involves evaluating a proposal generator on 
\emph{a partially annotated} dataset such as PASCAL. The protocol does not reward recall of 
non-PASCAL categories; in fact, early recall (near the top of the list of candidates) of non-PASCAL 
objects results in a penalty for the proposal generator! 
As a result, a proposal generator that tunes 
itself to these 20 PASCAL categories (either explicitly via training or implicitly via design choices 
or hyper-parameters) 
will be declared a better proposal generator when it may not be (as illustrated by DMP). 
Notice that as learning-based object proposal methods improve on this metric, ``in the limit'' \emph{the best object proposal technique is a detector for the annotated categories}, similar to our DMP. Thus, we should be cautious of methods proposing incremental improvements on this protocol -- improvements on this protocol do not necessarily lead to a better category independent object proposal method. 

This thought experiment exposes the inability of the existing protocol to evaluate 
category independence.
\label{sec:exp}
\vspace{\sectionReduceTop}
\section{Evaluation on Fully and Densely Annotated Datasets}
\vspace{\sectionReduceBot}
\begin{figure*}[ht]
	\vspace{-0.4cm}
	\centering
	\begin{minipage}[b]{0.23\textwidth}
		\includegraphics[width=1\columnwidth]{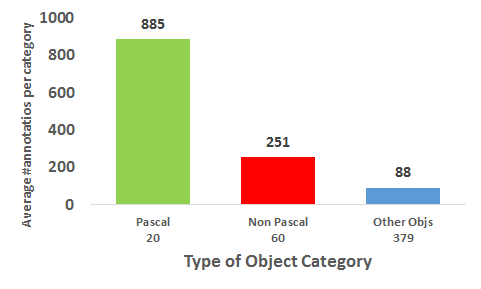}
		\subcaption{Average \#annotations for different categories.}
		\label{barplot_freq}
	\end{minipage}%
	\,\,\
	\begin{minipage}[b]{0.23\textwidth}
		\includegraphics[width=1\columnwidth]{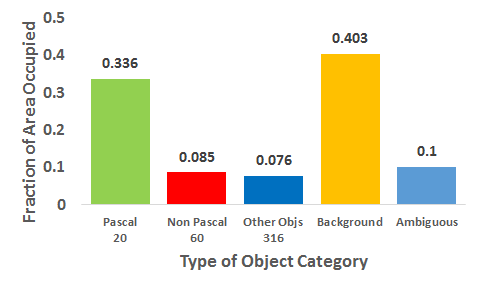}
		\subcaption{Fraction of image-area covered by different categories.}
		\label{barplot_area}
	\end{minipage}
	\,\,\
	\begin{minipage}[b]{0.23\textwidth}
		\includegraphics[width=1\columnwidth]{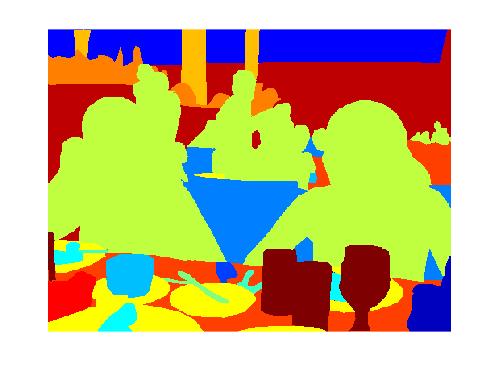}\vspace{-7pt}
		\subcaption{PASCAL Context annotations~\cite{mottaghi_cvpr14}.}
		\label{context}
	\end{minipage}%
	\,\,\
	\begin{minipage}[b]{0.23\textwidth}
		\includegraphics[width=1\columnwidth]{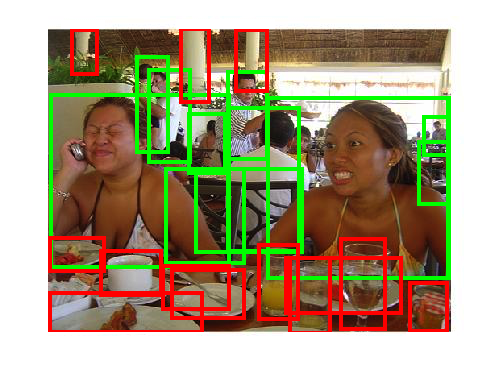}\vspace{-7pt}
		\subcaption{Our augmented annotations.\\}
		\label{mod_context}
	\end{minipage}
	\caption{(a),(b) Distribution of object classes in PASCAL Context with respect to different attributes.
		(c),(d) Augmenting PASCAL Context with instance-level annotations. 
		(Green = PASCAL 20 categories; Red = new objects)}
	\label{context_anno_fig}
\end{figure*}
As described in the previous section, the problem of `gameability' is occuring due to the evaluation of proposal methods on partially annotated datasets. An intuitive solution  would be evaluating on a \emph{fully} annotated dataset. \\
In the next two subsections, we evaluate the performance of 7 popular object proposal methods\cite{AlexePAMI12,UijlingsIJCV13,cheng2014bing,ZitnickECCV14,Arbelaez_CVPR14,RahtuICCV11,ManenICCV13} and two DMPs 
(RCNN\cite{DBLP:journals/corr/GirshickDDM13} and DPM\cite{lsvm-pami}) on one nearly-fully and two densely annotated datasets containing
many more object categories. 
This is to quantify how much the performance of our `fraudulent' proposal generators (DMPs) drops  
once the bias towards the 20 PASCAL categories is diminished (or completely removed). \\
We begin by \emph{creating} a nearly-fully annotated dataset 
by building on the effort of PASCAL Context \cite{mottaghi_cvpr14} and evaluate on this nearly-fully annotated modified instance level PASCAL Context; 
followed by cross-dataset evaluation on other partial-but-densely annotated datasets MS COCO \cite{LinECCV14coco} 
and NYU-Depth V2 \cite{Silberman:ECCV12}. 

\textbf{Experimental Setup:} On MS COCO and PASCAL Context datasets we conducted experiments as follows:
\begin{compactitem}
	\item Use the existing evaluation protocol for evaluation, \ie, evaluate only on the 20 PASCAL categories.
	\item Evaluate on all the annotated classes.
	\item For the sake of completeness, we also report results on all the classes except the PASCAL 20 classes.\footnote{On NYU-Depth V2 performance is only evaluated on all categories. This is because only 8 PASCAL categories are present in this dataset.}
\end{compactitem}

\textbf{Training of DMPs:} The two DMPs we use are based on two popular object detectors - DPM\cite{lsvm-pami} and RCNN\cite{DBLP:journals/corr/GirshickDDM13}.
We train DPM 
on 20 PASCAL categories and use it as an object proposal method. To generate large number of proposals, 
we chose a low value of threshold in Non-Maximum Suppression (NMS). Proposals are generated for each category and a score is assigned to them by the corresponding DPM for that category. 
These proposals are then merge-sorted on the basis of this score. 
Top M proposals are selected from this sorted list where M is the number of proposals to be generated.\\
Another (stronger) DMP is RCNN which is a detection pipeline that uses 20 SVMs 
(each for one PASCAL category) trained on deep features (fc7)~\cite{donahue_arxiv13} extracted on selective search boxes. 
Since RCNN itself uses selective search proposals, it should be viewed as a trained \emph{reranker} 
of selective search boxes. As a consequence, it ultimately equals selective search performance once 
the number of candidates become large. We used the pretrained SVM models released 
with the RCNN code, which were trained on the 20 classes of PASCAL VOC 2007 trainval set. 
For every test image, we generate the Selective Search proposals using the `FAST' mode and 
calculate the 20 SVM scores for each proposal. 
The `objectness' score of a proposal is then the maximum of the 20 SVM scores. All the proposals are then sorted by this score and top M proposals are selected.\footnote{It was observed that merge-sorting calibrated/rescaled  SVM scores led to inferior performance as compared to merge-sorting without rescaling.}

\textbf{Object Proposals Library:} 
To ease the process of carrying out the experiments, we created an open source, 
easy-to-use object proposals library. This can be used to seamlessly generate object proposals using 
all the existing algorithms\cite{ZitnickECCV14,EndresPAMI14,Arbelaez_CVPR14,AlexePAMI12,RahtuICCV11,ManenICCV13,RantalankilaCVPR14,UijlingsIJCV13,HumayunCVPR14} (for which the Matlab code has been released by the respective authors)\com{. 
Our library can also be used to} and evaluate these proposals on any dataset using the commonly used metrics\com{: 
Recall $@$ Threshold/Detection Rate, ABO, AUC}. This library will be made publicly available.
\vspace{-0.1cm}
\vspace{\subsectionReduceTop}
\subsection{Fully Annotated Dataset}
\vspace{\subsectionReduceBot}
\paragraph{PASCAL Context:}This dataset was introduced by Mottaghi~\etal~\cite{mottaghi_cvpr14}. 
It contains additional annotations for all images of PASCAL VOC 2010 dataset \cite{pascal-voc-2010}. 
The annotations are semantic segmentation maps, where \emph{every single pixel} 
previously annotated `background' in PASCAL was assigned a category label. 
\begin{figure*}[htp]
\centering
\begin{subfigure}[b]{0.3\textwidth}
 \includegraphics[width=1.1\columnwidth]{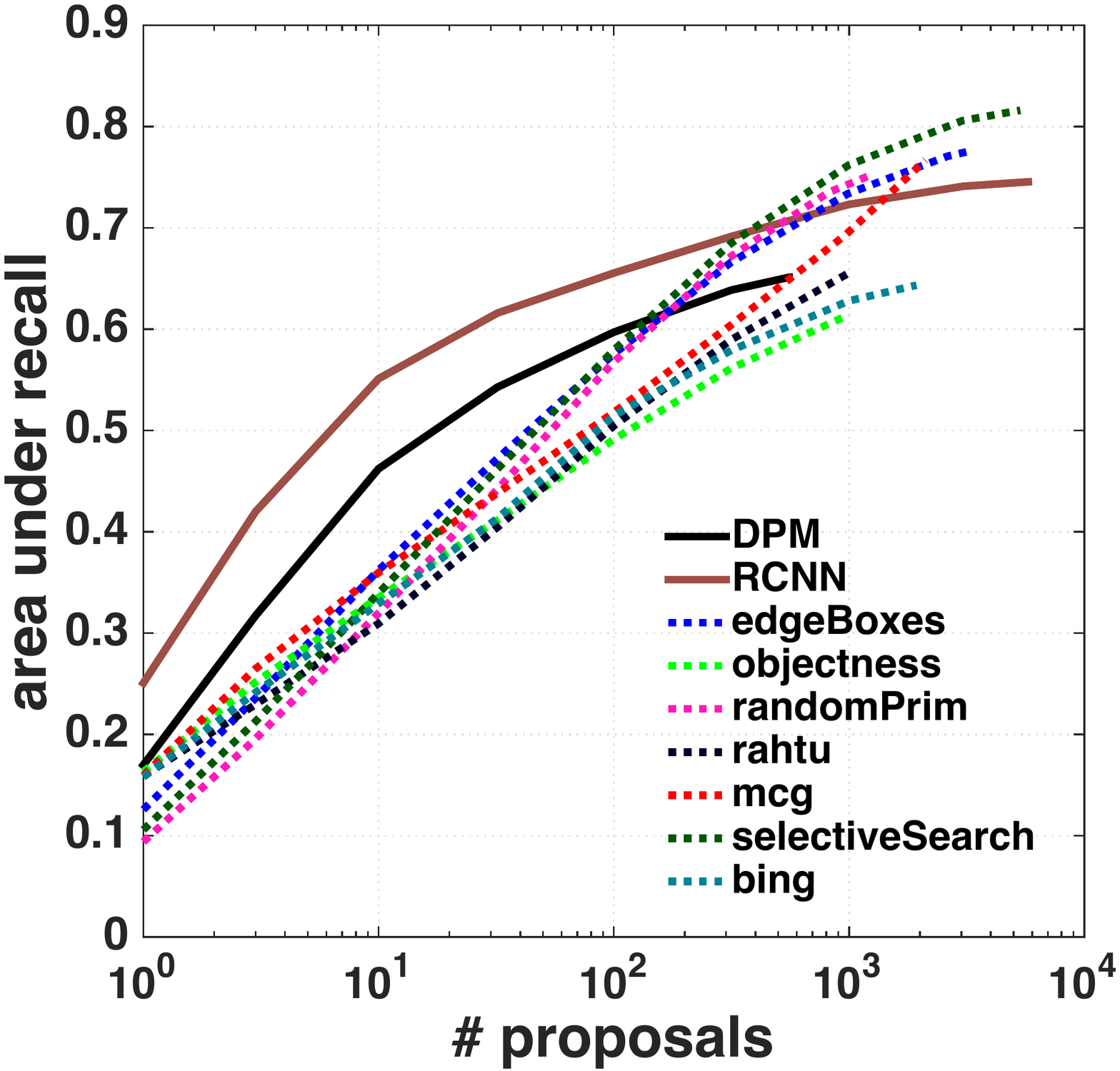}
\subcaption{Performance on PASCAL Context, only 20 PASCAL classes annotated.}
\label{fig:pascal2010aucAllCat}
\end{subfigure}
\,\,\,
\begin{subfigure}[b]{0.3\textwidth}
 \includegraphics[width=1.1\columnwidth]{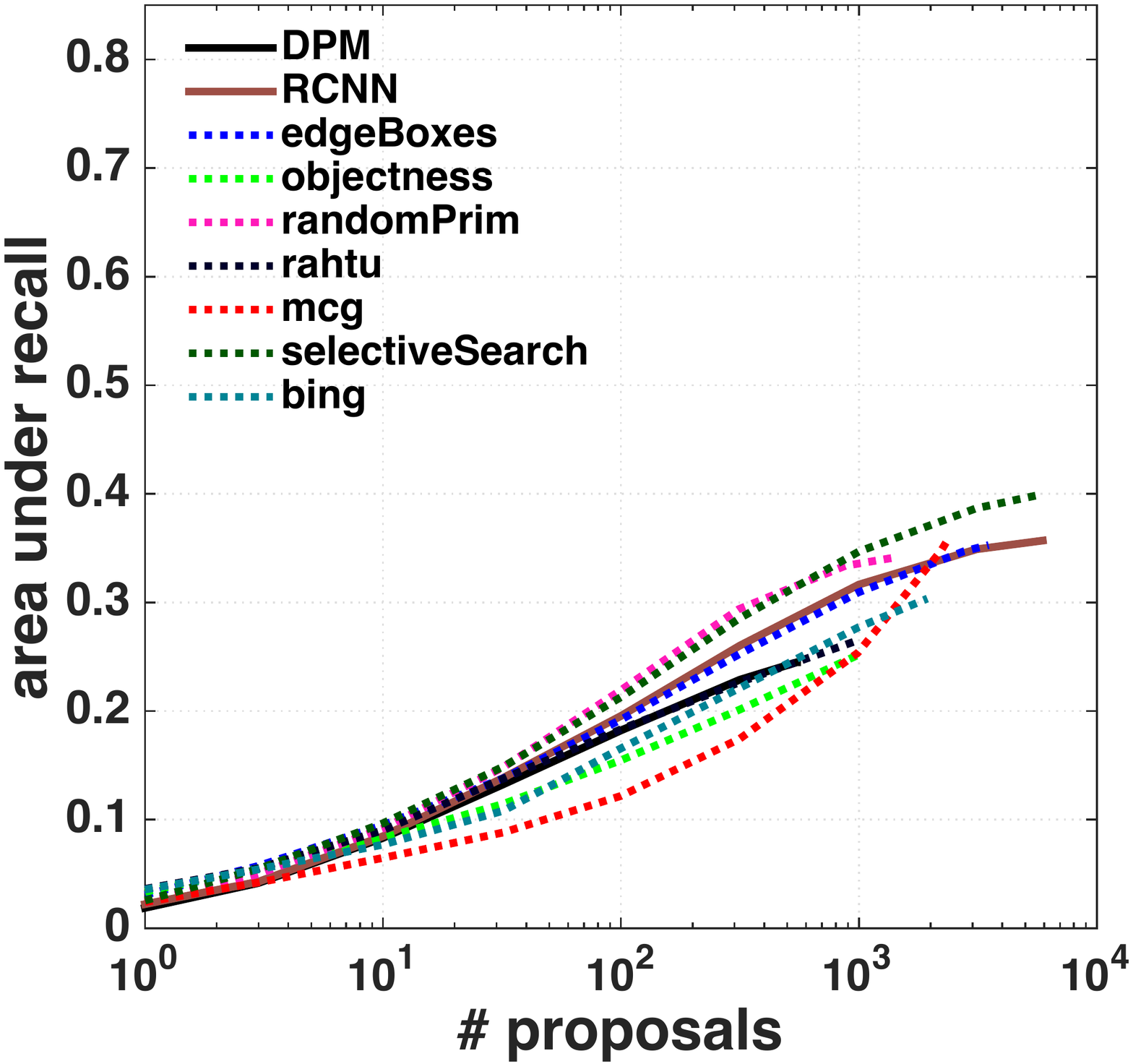}
\subcaption{Performance on PASCAL Context, only 60 non-PASCAL classes annotated.} 
	\label{fig:pascal2010aucNonPasCat}
\end{subfigure}
\,\,\,
\begin{subfigure}[b]{0.3\textwidth}
	\includegraphics[width=1.1\columnwidth]{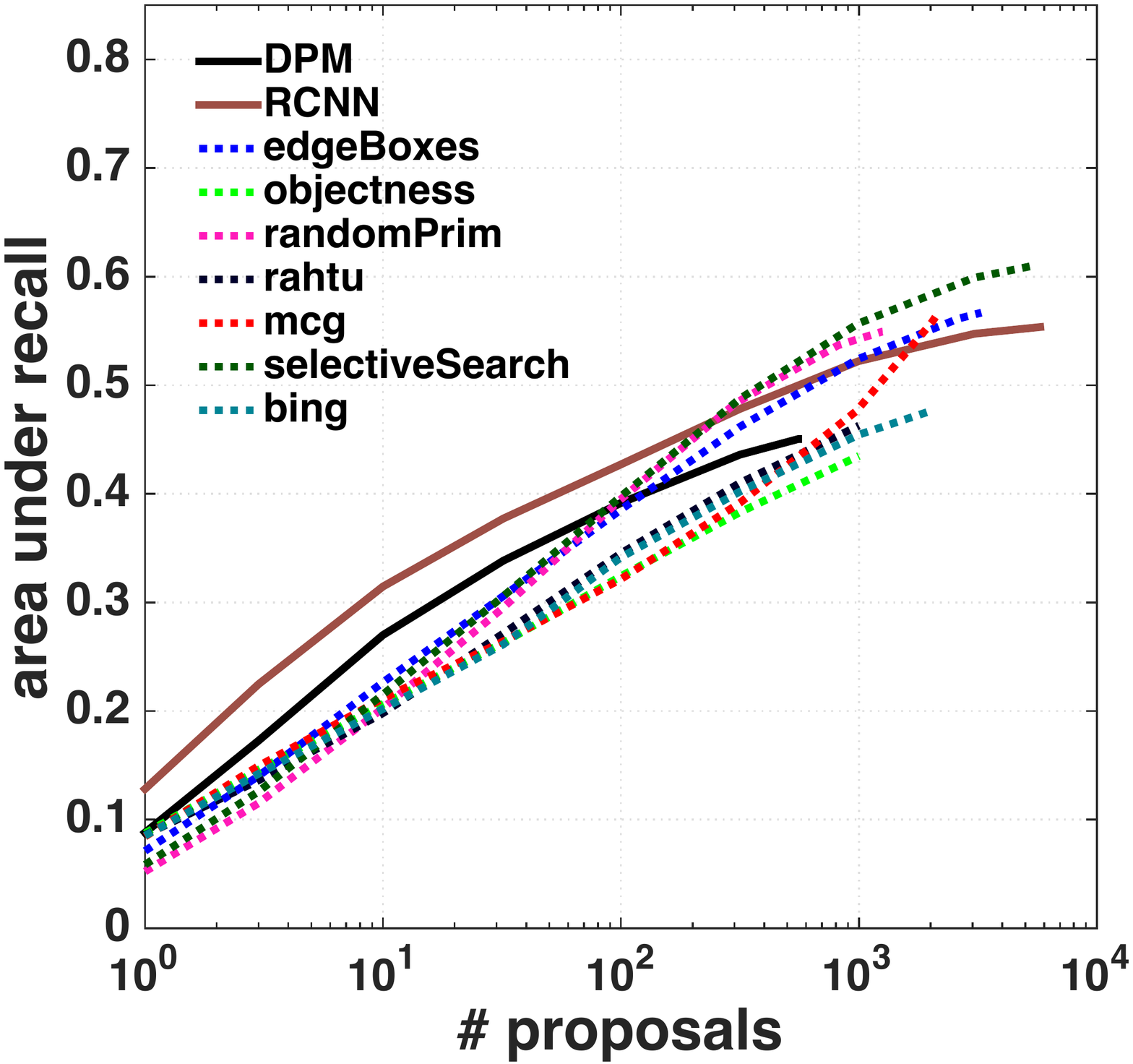}
	\subcaption{Performance on PASCAL Context, all classes annotated.} 
		\label{fig:pascal2010AllCat}
\end{subfigure}
\centering
\begin{subfigure}[b]{0.3\textwidth}
	\includegraphics[width=1.1\columnwidth]{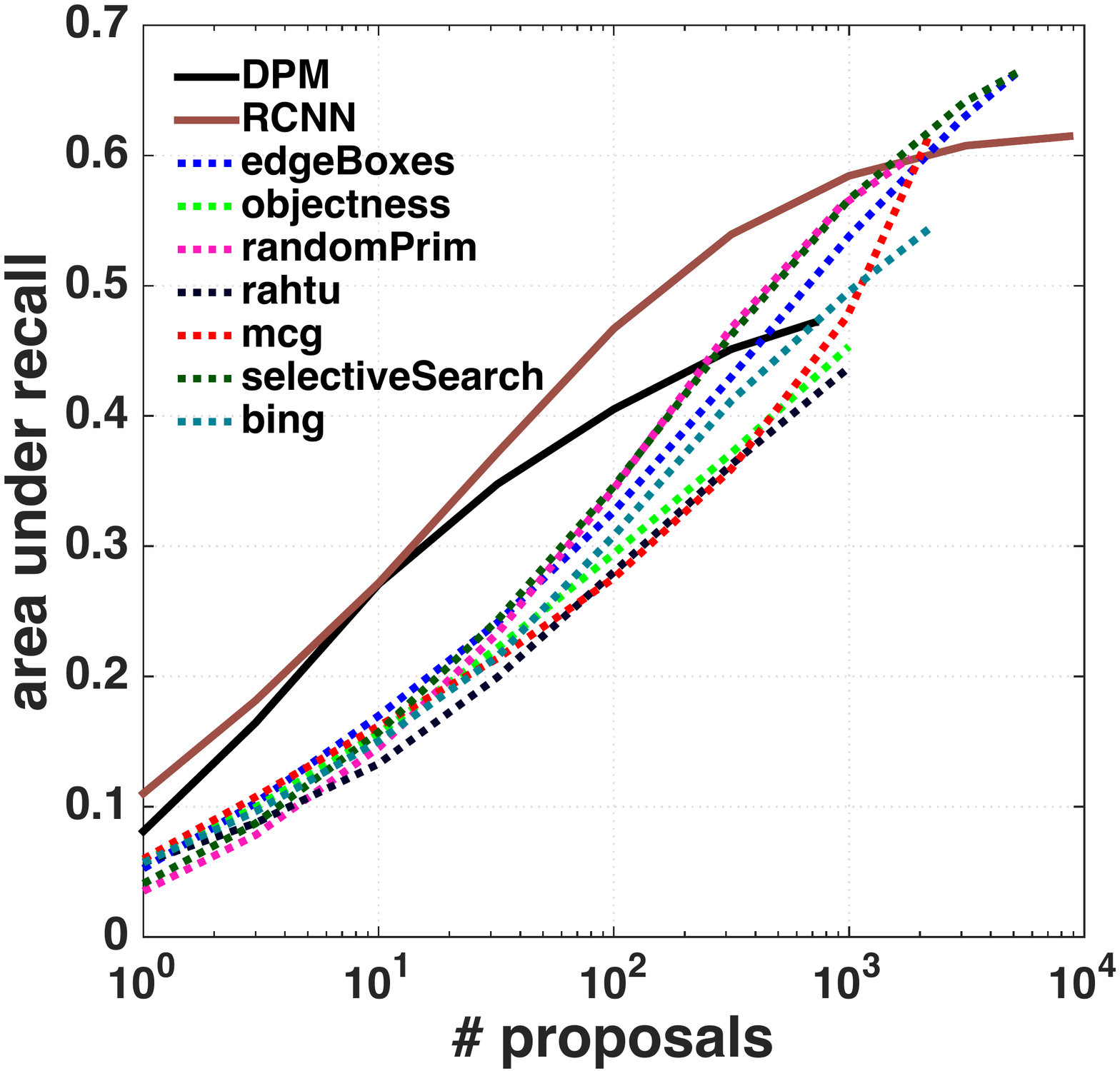}
	\subcaption{Performance on MS COCO, only 20 PASCAL classes annotated.}
		\label{fig:cocoAuc20Cat} 
\end{subfigure}
\,\,\,
\begin{subfigure}[b]{0.3\textwidth}
 \includegraphics[width=1.1\columnwidth]{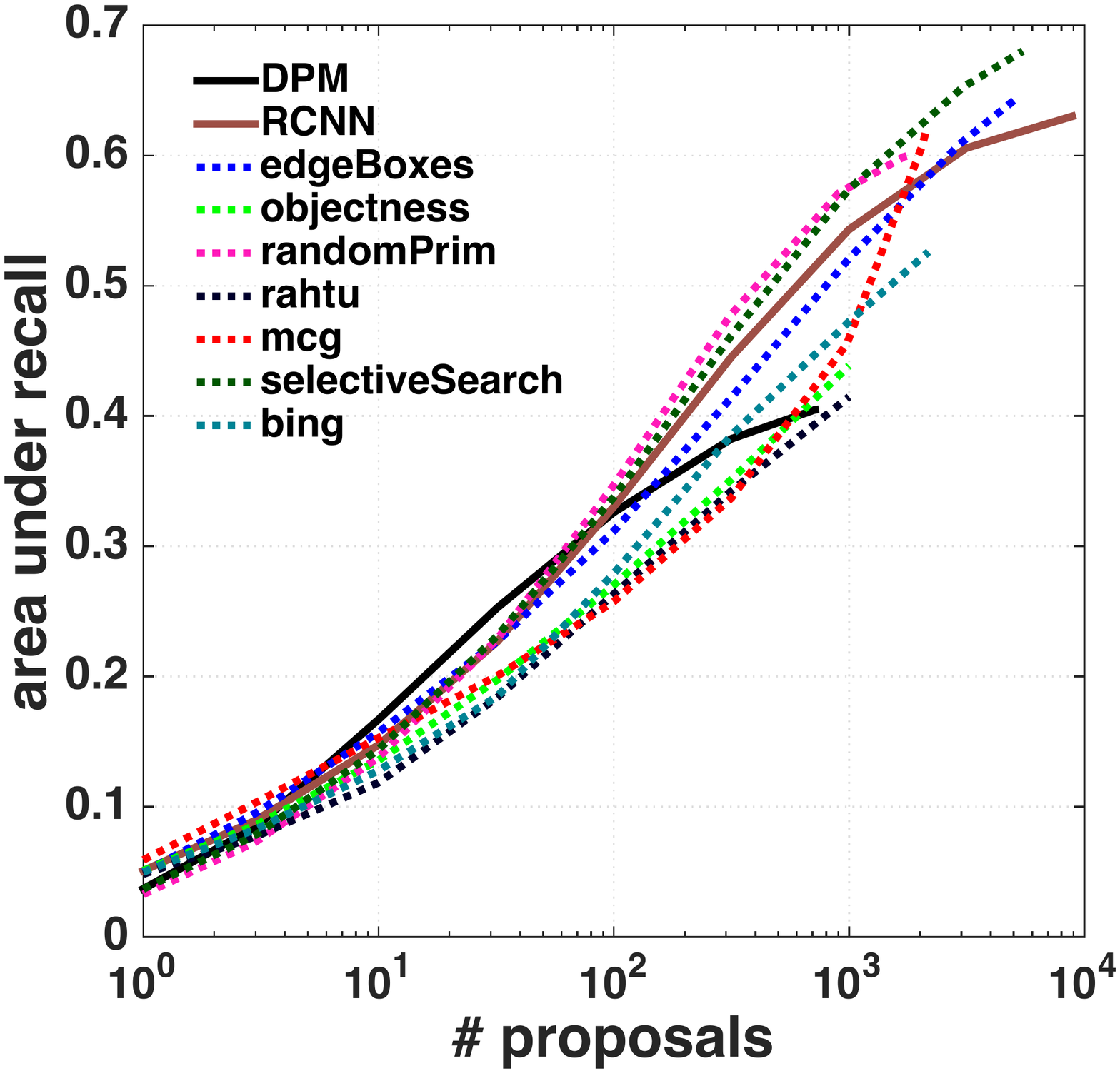}
\subcaption{Performance on MS COCO, only 60 non-PASCAL classes annotated.} 
	\label{fig:cocoAuc70Cat} 
\end{subfigure}
\,\,\,
\begin{subfigure}[b]{0.3\textwidth}
	\includegraphics[width=1.1\columnwidth]{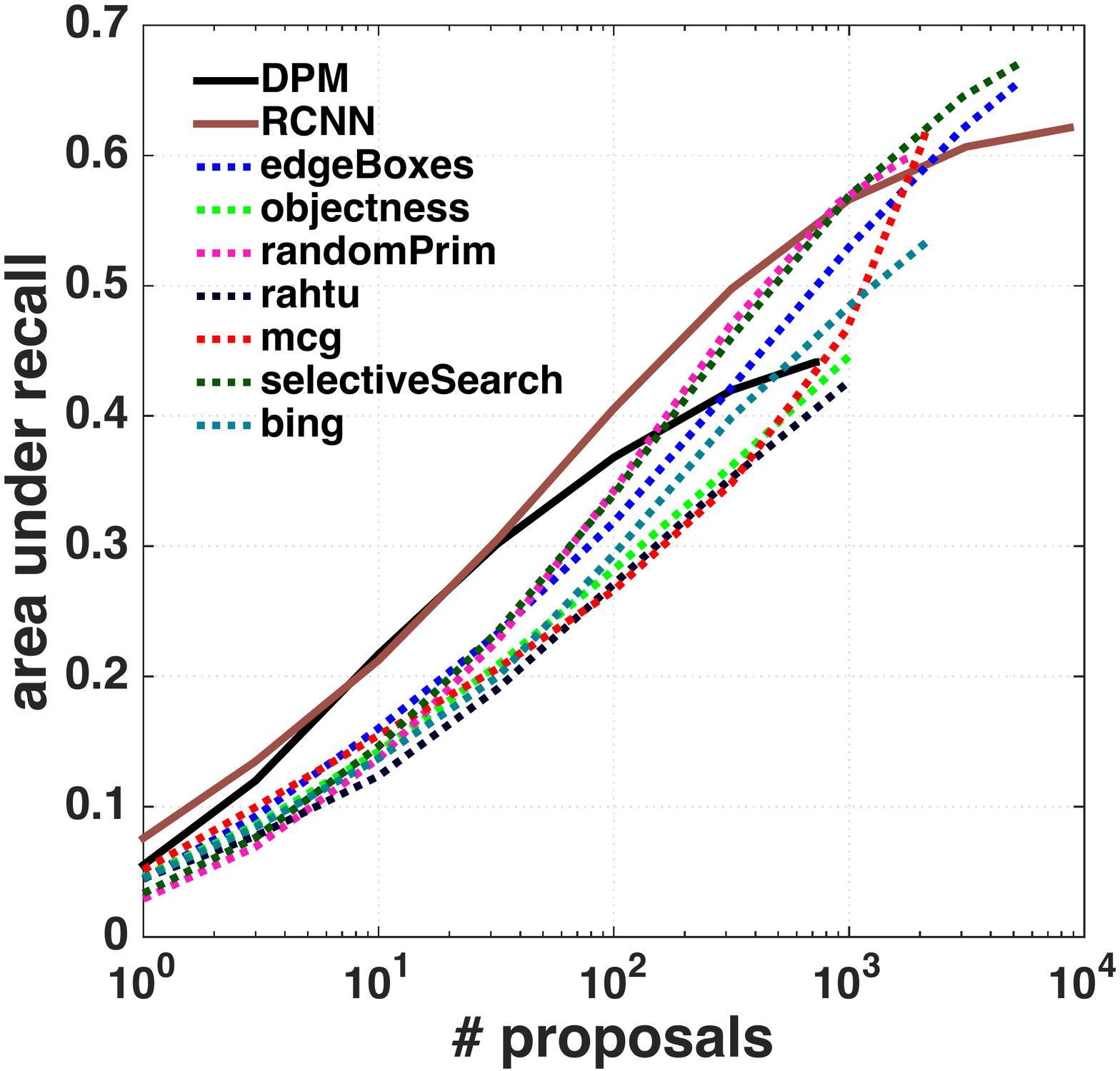}
	\subcaption{Performance on MS COCO, all classes annotated.\label{fig:cocoauc90Cat}}
\end{subfigure}
\centering
\begin{subfigure}[b]{0.3\textwidth}
	\includegraphics[width=1.1\columnwidth]{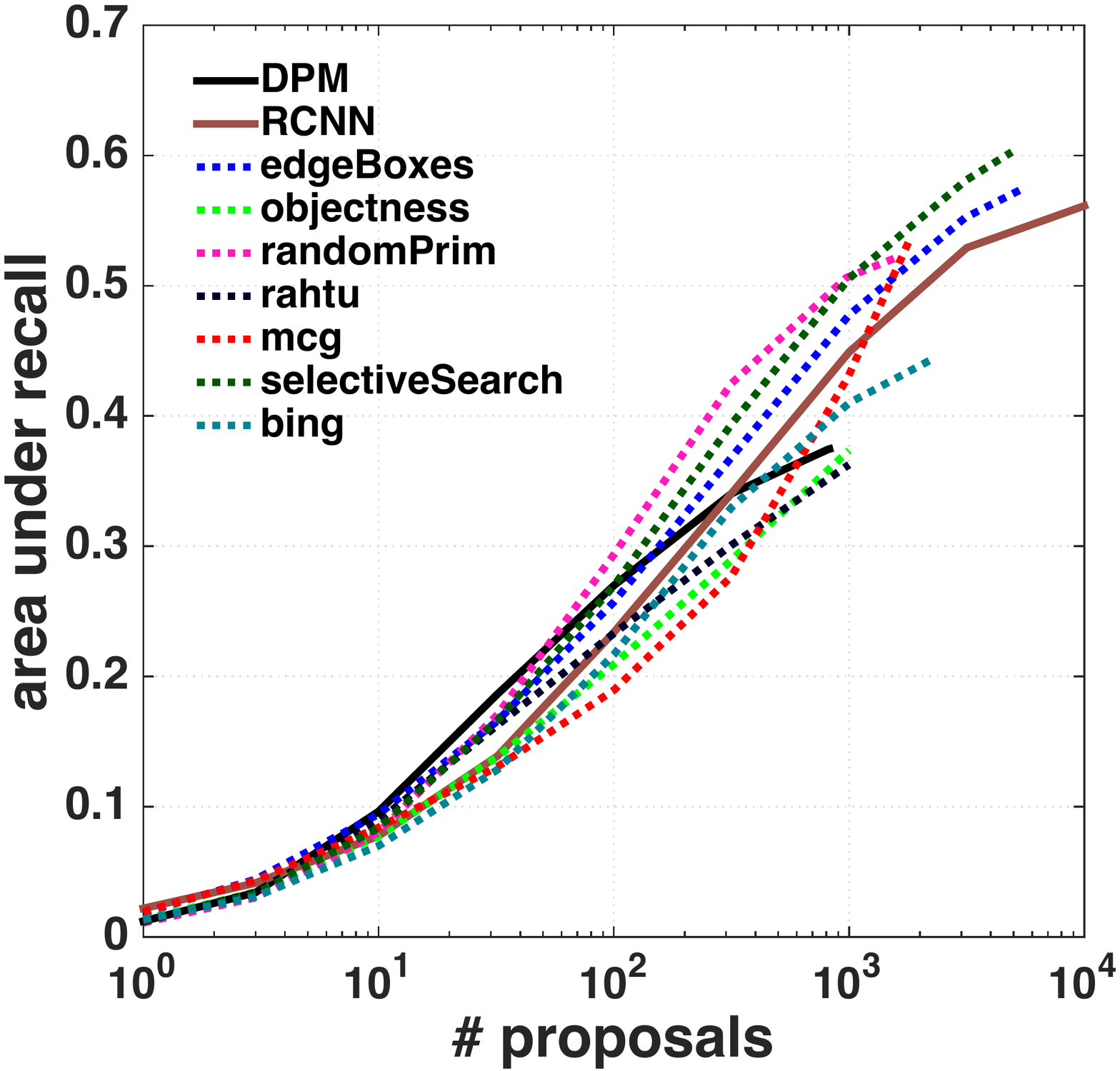}
	\subcaption{Performance on NYU-Depth V2, all classes annotated
		\label{fig:nyuv2_allcats}}
\end{subfigure}
\,\,\,
\begin{subfigure}[b]{0.3\textwidth}
	\includegraphics[width=1.1\columnwidth]{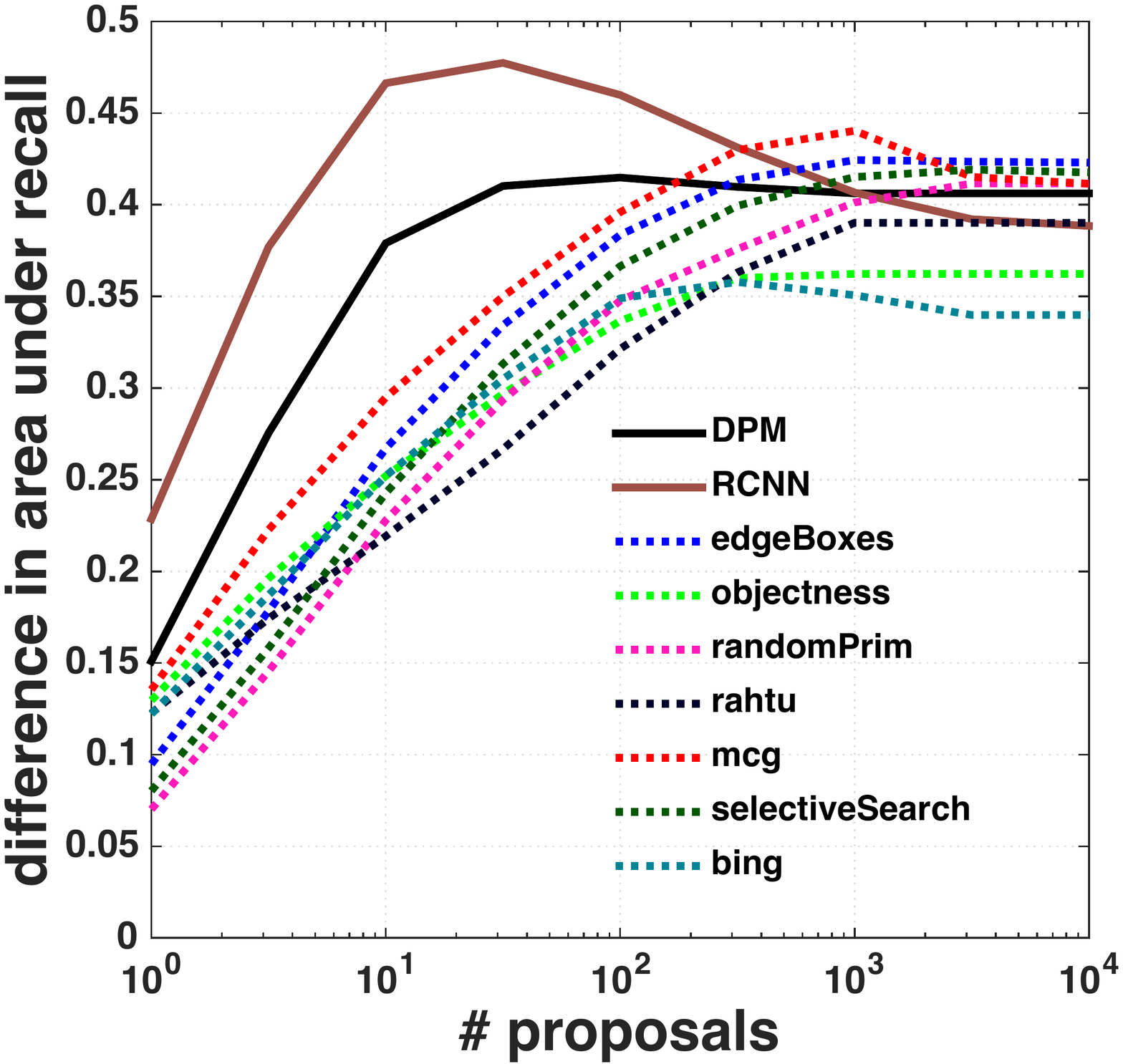}
	\subcaption{\label{fig:diff_pascal}AUC @ 20 categories - AUC @ 60 categories on PASCAL Context.}
\end{subfigure}
\,\,\,
\begin{subfigure}[b]{0.3\textwidth}
	\includegraphics[width=1.1\columnwidth]{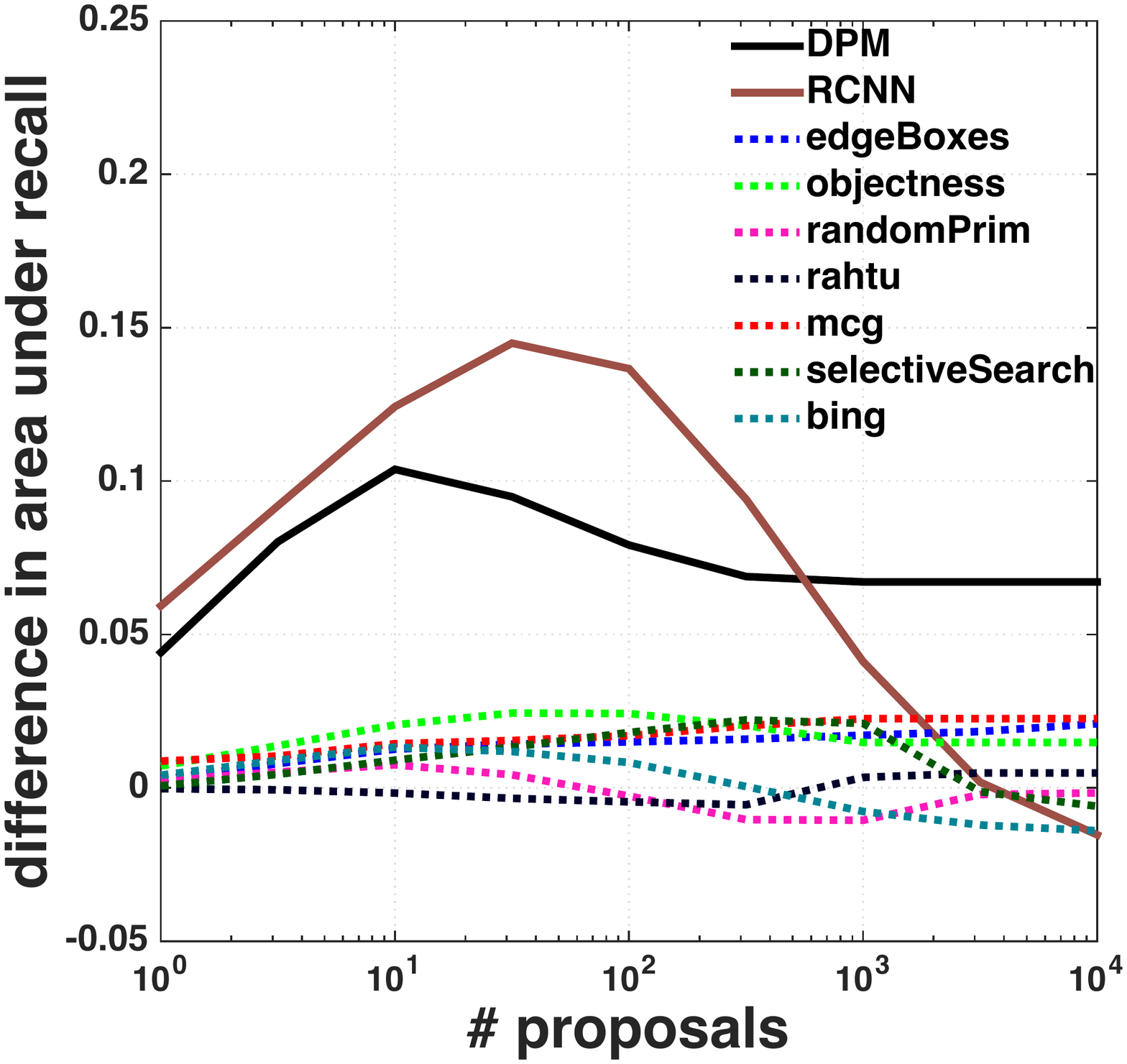}
	\subcaption{\label{fig:diff_coco}AUC @ 20 categories - AUC @ 60 categories on MS COCO.}
\end{subfigure}
\caption{Performance of different methods on PASCAL Context, MS COCO and NYu Depth-V2 with different sets of annotations.}
\end{figure*}
In total, annotations have been provided for 459 categories. 
This includes the original 20 PASCAL categories and new 
classes such as keyboard, fridge, picture, cabinet, plate, clock. \\
Unfortunately, the dataset contains only 
category-level semantic segmentations\com{, not \emph{instance-level} segmentations}. 
For our task, we needed instance-level bounding box annotations, which cannot be 
reliably extracted from category-level segmentation masks\com{ because the masks for several instances (of 
say chairs) may be merged together into a single `blob' in the category-level mask}. 

\textbf{Creating Instance-Level Annotations for PASCAL Context:}
Thus, we created instance-level bounding box annotations for all images in PASCAL Context dataset. 
First, out of the 459 category labels in PASCAL Context, we identified 396  categories to be `things', and ignored the remaining `stuff' 
or `ambiguous' categories\footnote{\eg, a `tree' may be a `thing' 
or `stuff' subject to camera viewpoint.} -- 
neither of these lend themselves to bounding-box-based object detection. See supplement for details.\com{A complete list is available in the supplementary material. }\\
We selected the 60 most frequent non-PASCAL categories from this list of `things' and manually annotated all their instances. Selecting only top 60 categories is a reasonable choice because the average per category frequency in the dataset for all the other categories (even after including background/ambiguous categories) was roughly one third as that of the chosen 60 categories (\figref{barplot_freq}). Moreover, the percentage of pixels in an image left unannotated (as `background') drops from 58\% in original PASCAL to 50\% in our nearly-fully annotated PASCAL Context. This manual annotation was 
performed with the aid of the semantic segmentation maps present in the PASCAL Context annotations. Examples annotations are shown in \figref{mod_context}. For detailed statistics, see supplement.

\textbf{Results and Observations:} 
We now explore how changes in the dataset and annotated categories affect the results of the thought experiment 
from \secref{sec:thought}. 
Figs.~\ref{fig:pascal2010aucAllCat}, ~\ref{fig:pascal2010aucNonPasCat},~\ref{fig:pascal2010AllCat}, ~\ref{fig:diff_pascal}
compare the performance of DMPs 
with a number of existing proposal methods \cite{ZitnickECCV14,EndresPAMI14,Arbelaez_CVPR14,AlexePAMI12,RahtuICCV11,ManenICCV13,UijlingsIJCV13,cheng2014bing,DBLP:conf/eccv/KrahenbuhlK14} 
on PASCAL Context. 

We can see in Column (a) that when evaluated on only 20 PASCAL categories 
DMPs trained on these categories appear to significantly outperform all proposal generators. 
However, we can see that they are not category independent because they suffer a big 
drop in performance when evaluated on 60 non-PASCAL categories in Column (b). 
Notice that on PASCAL context, \emph{all proposal generators} suffer a drop in performance 
between the 20 PASCAL categories and 60 non-PASCAL categories. We hypothesize that this due to the fact that the non-PASCAL categories tend to be generally smaller 
than the PASCAL categories (which were the main targets of the dataset curators) and hence difficult to detect.  But this could also be due to  the reason that authors of these methods made certain choices while designing these approaches which catered better to the 20 annotated categories.
However, the key observation here (as shown in ~\figref{fig:diff_pascal}) is that DMPs suffer the biggest drop. This drop is much greater than all the other approaches.
It is interesting to note that due to the ratio of instances of 20 PASCAL categories vs other 60 categories, 
DMPs continue to slightly outperform proposal generators when evaluated on all categories, as shown 
in Column (c). 
\vspace{-0.25cm}
\vspace{\subsectionReduceTop}
\subsection{Densely Annotated Datasets } 
\vspace{\subsectionReduceBot}
Besides being expensive, ``full'' annotation of images is somewhat ill-defined due to the hierarchical nature of object semantics (\eg are object-parts such as bicycle-wheel, windows in a building, eyes in a face, \etc also objects?). One way to side-step this issue is to use datasets with dense annotations (albeit at the same granularity) and conduct cross-dataset evaluation.

\textbf{MS COCO:} Microsoft Common Objects in Context (MS COCO) dataset \cite{LinECCV14coco} 
contains 91 common object categories with 82 of them having more than 5,000 labeled instances. 
It not only has significantly higher number of instances per category than the PASCAL, 
but also considerably more object instances per image (7.7) 
as compared to ImageNet (3.0) and PASCAL (2.3). 

\textbf{NYU-Depth V2:} NYU-Depth V2 dataset \cite{Silberman:ECCV12} is comprised of video sequences from a 
variety of indoor scenes as recorded by both the RGB and Depth cameras. 
It features 1449 densely labeled pairs of aligned RGB and depth imageswith instance-level annotations. We used these 1449 densely annotated 
RGB images for evaluating object proposal algorithms. To the best of our knowledge, 
this is the first paper to compare proposal methods on such a dataset. 

\textbf{Results and Observations:} Figs. \ref{fig:cocoAuc20Cat}, \ref{fig:cocoAuc70Cat}, \ref{fig:cocoauc90Cat}, \ref{fig:diff_coco} show a plot similar to PASCAL Context on MS COCO. Again, DMPs outperform all other methods on PASCAL categories but fail to do so for the Non-PASCAL categories.
\figref{fig:nyuv2_allcats} shows results for NYU-Depth V2. 
See that when many classes in the test dataset are 
not PASCAL classes, DMPs tend to perform poorly, although it is interesting that the performance is 
still not as poor as the worst proposal generators. Results on other evaluation criteria are in the supplement.\\
\vspace{-0.5cm}
\vspace{\sectionReduceTop}
\section{Bias Inspection}
\vspace{\sectionReduceBot}
So far, we have discussed two ways of detecting `gameability' -- evaluation on nearly-fully annotated dataset and cross-dataset evaluations on densely annotated datasets.
\begin{figure*}[ht]
	\vspace{-0.3cm}
	\centering
	\begin{minipage}[b]{0.32\textwidth}
		\centering
		\includegraphics[width=0.93\textwidth]{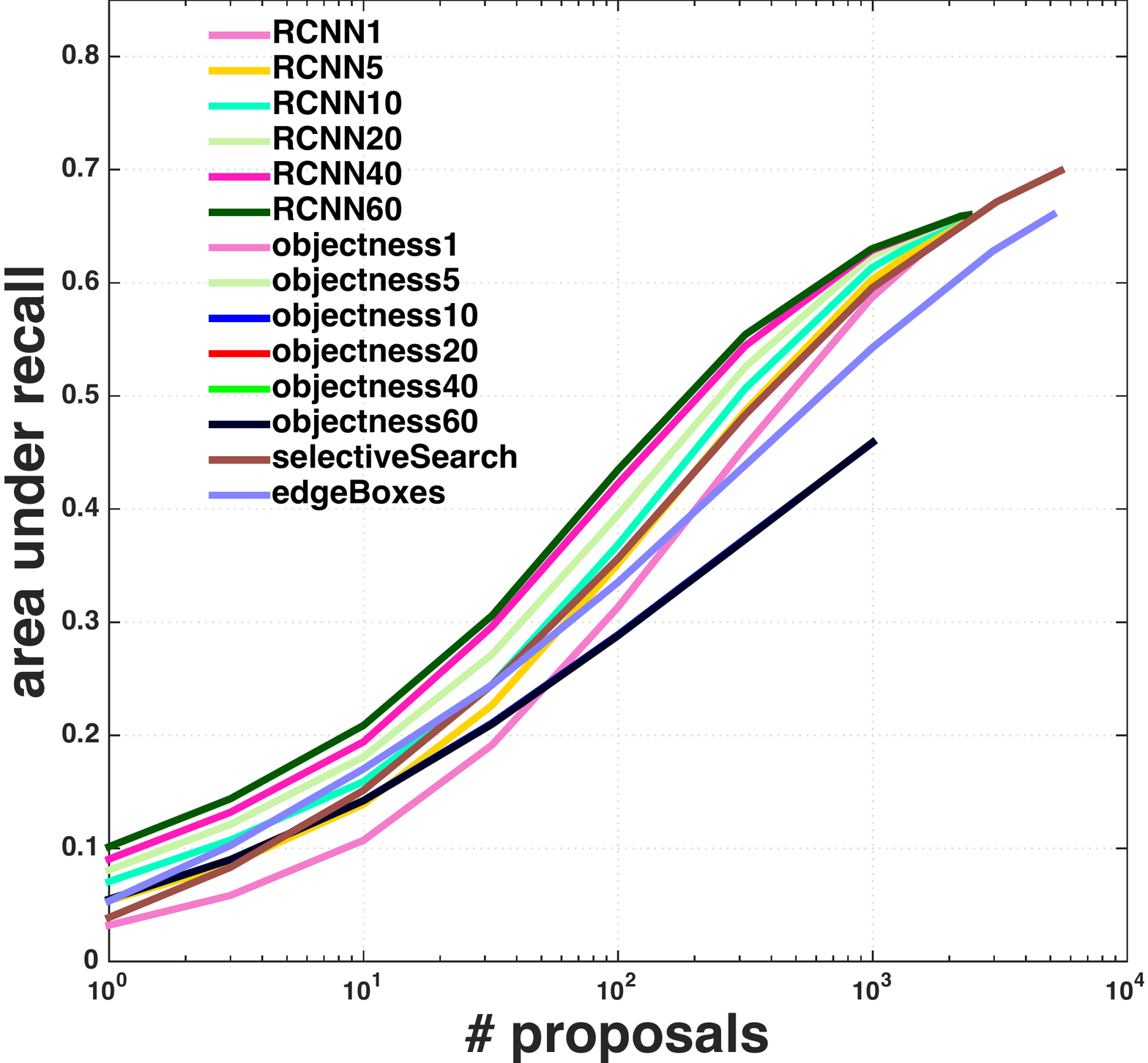}
		\subcaption{\label{trainExp} Area under recall \vs \# proposals for various \#seen categories}
	\end{minipage}
	\,\,\
	\begin{minipage}[b]{0.32\textwidth}
		\includegraphics[width=1.01\columnwidth]{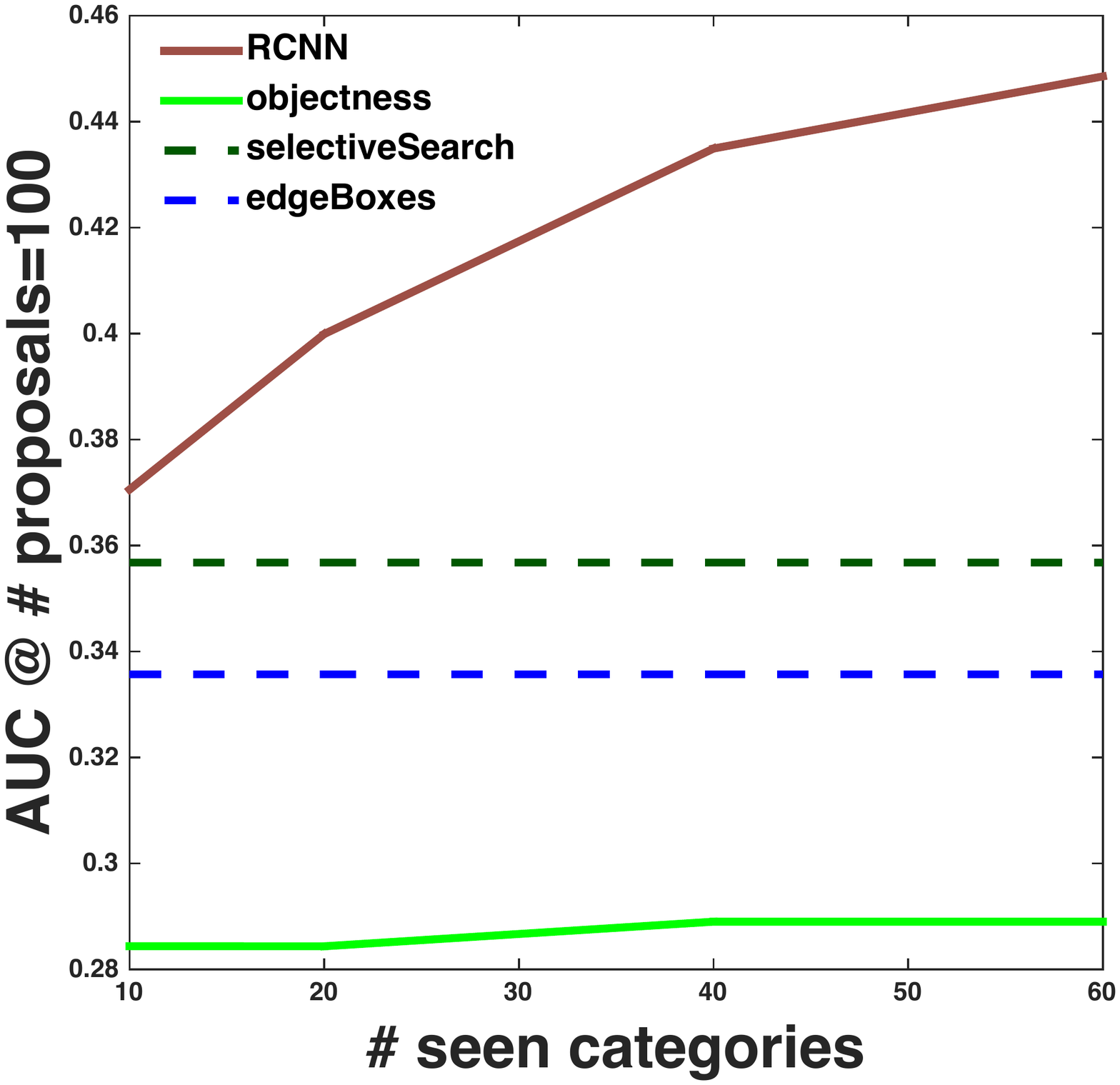}
				\vspace{-0.5cm}
		\subcaption{\label{trainExp2}Area under recall \vs \#training-categories.}
	\end{minipage}
	\,\,\
	\begin{minipage}[b]{0.32\textwidth}
		\includegraphics[width=1.01\columnwidth]{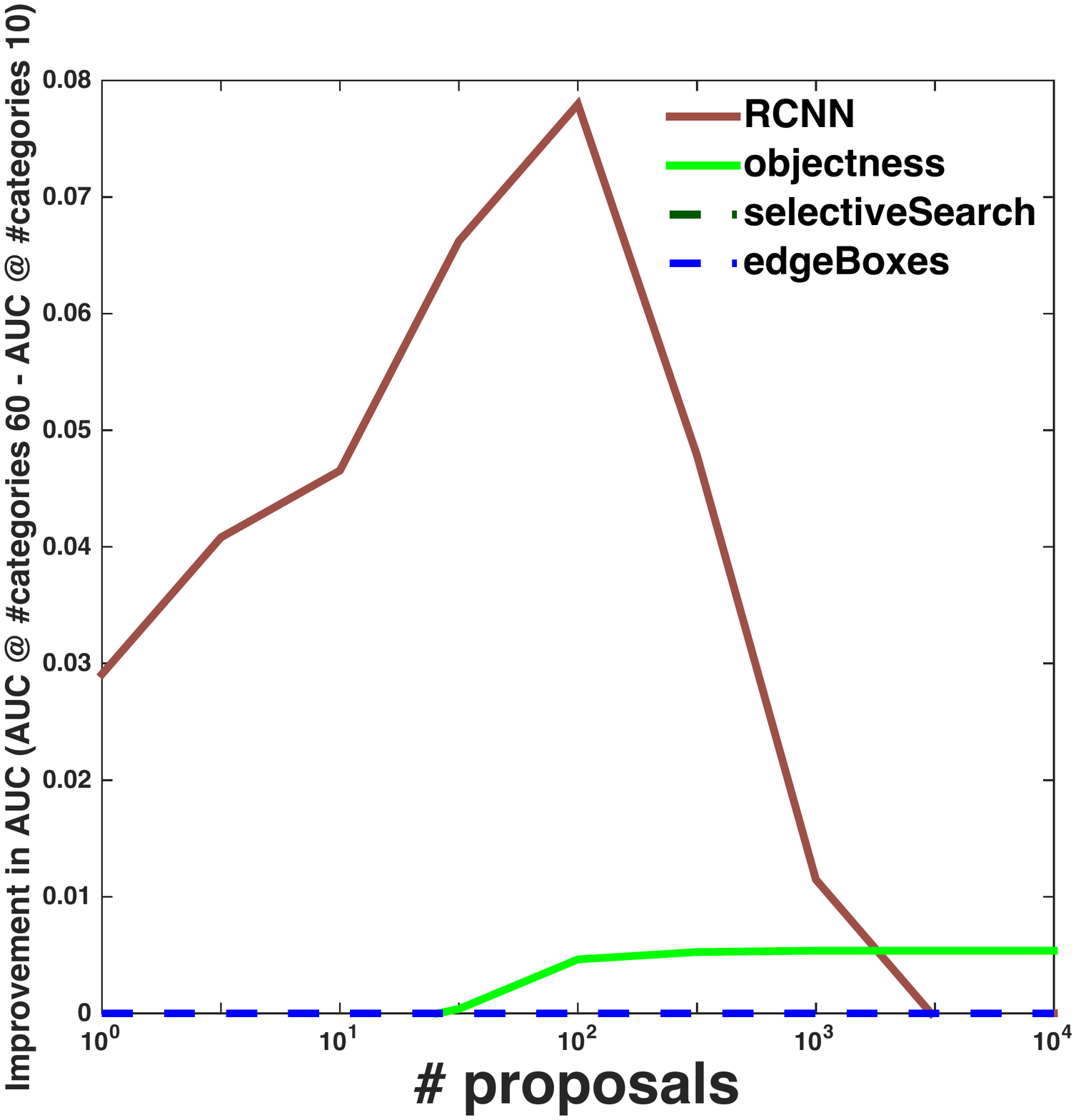}
				\vspace{-0.5cm}
		\subcaption{\label{trainExp3}Improvement in area under recall from \#seen categories =10 to 60 \vs \# proposals.}
	\end{minipage}
	\caption{\label{trainExpall}Performance of RCNN and other proposal generators vs number of object categories used for training.
		We can see that RCNN has the most `bias capacity' while the performance of other methods is nearly 
		(or absolutely) constant. }
\end{figure*}
Although these methods are fairly useful for bias detection,  they have certain limitations. Datasets can be unbalanced. Some categories can be more frequent than others while others can be hard to detect (due to choices made in dataset collection). These issues need to be resolved for perfectly unbiased evaluation. However, generating unbiased datasets is an expensive and time-consuming process. Hence, to detect the bias without getting unbiased datasets, we need a method  which can measure performance of proposal methods in a way that category specific biases can be accounted for and the extent or the \emph{capacity} of this bias can be measured.
We introduce such a method  in this section. 
\vspace{-0.15cm}
\vspace{\subsectionReduceTop}
\subsection{Assessing Bias Capacity}
\vspace{\subsectionReduceBot}

Many proposal methods\cite{cheng2014bing,DBLP:conf/eccv/KrahenbuhlK14,HumayunCVPR14,Krahenbuhl_2015_CVPR,DBLP:journals/pami/KangHEK15,DBLP:journals/corr/KuoHM15, deepprop,DBLP:journals/corr/PinheiroCD15} rely on explicit training to learn an ``objectness'' model, similar to DMPs. 
Depending upon which, how many categories they are trained on, these methods could have a biased view of ``objectness''.\\
One way of measuring the \emph{bias capacity} in a proposal method to plot the 
performance \vs the number of `seen' categories while evaluating on some held-out set. A method that involves little or no training will be a flat curve on this plot. Biased methods such as DMPs will get better and better as more categories are seen in training. 
Thus, this analysis can help us find biased or `gamebility-prone' methods like DMPs that are/can be tuned to specific classes. To the best of our knowledge, no previous work has attempted to measure bias capacity by varying the number of `object' categories seen at training time.
In this experiment, we compared the performance of 
one DMP method (RCNN), 
one learning-based proposal method (Objectness), and two non learning-based 
proposal methods (Selective Search\cite{UijlingsIJCV13}, EdgeBoxes\cite{ZitnickECCV14}) 
as a function of the number of `seen' categories (the categories trained on\footnote{The seen categories are picked in the order they are listed in MS COCO dataset (\ie, no specific criterion was used).}) on MS COCO\cite{LinECCV14coco} dataset. 
Method names `RCNNTrainN', `objectnessTrainN' indicate that they were trained on images that contain annotations for only N categories (50 instances per category). Total number of images for all 60 categories was \textasciitilde2400  (because some images contain >1 object). Once trained, these methods were evaluated on 
a randomly-chosen set of \textasciitilde500 images, which had annnotations for all 60 categories.\\
\figref{trainExp} shows Area under Recall \vs \#proposals curve for learning-based methods trained on different sets of categories. 
\figref{trainExp2} and \figref{trainExp3} show the variation of AUC \vs \# seen categories and improvement due to increase in training categories (from 10 to 60)  \vs \#proposals respectively, for RCNN and objectness when trained on different sets of categories.
The key observation to make here is that with even a modest increase in `seen' categories with the same amount of increased training data, performance improvement of RCNN is significantly more than objectness. Selective Search~\cite{UijlingsIJCV13} and edgeBoxes~\cite{ZitnickECCV14} are the dashed straight lines
since there is no training involved.

These results clearly indicate that as RCNN sees more categories, its performance improves. One might argue that the reason might be that the method is learning more `objectness' as it is seeing more data. However, as discussed above, the increase in the dataset size is marginal (\textasciitilde40 images per category) and hence it unlikely that such a significant improvement is observed due to that. Thus, it is reasonable to conclude that this improvement is because the method is learning class specific features.

Thus, this approach can be used to reason about `gameability-prone' and `gameability-immune' proposal methods without creating an expensive fully annotated dataset. We believe this simple but effective diagnostic experiment  would help to detect and thus contribute in managing the category specific bias in all learning-based methods.
\label{sec:newexp}
\vspace{\sectionReduceTop}
\section{Conclusion}
\vspace{\sectionReduceBot}

In this paper, we make an explicit distinction between the two mutually co-existing but different interpretations 
of object proposals.
The current evaluation protocol for object proposal methods 
is suitable only for detection proposals and is a biased `gameable' protocol for category-independent 
object proposals. By  evaluating only on a specific set of object categories, we fail to capture 
the performance of the proposal algorithm on all the remaining object categories that are present in the test set, but not annotated in the ground truth. 
	We demonstrate this gameability via a simple thought experiment 
	where we propose a `fraudulent' object proposal method that 
	outperforms all existing object proposal techniques on current metrics. 
We conduct a thorough evaluation of existing object proposal methods on three densely annotated datasets. 
We introduce a fully-annotated version of PASCAL VOC 2010 where we 
annotated all instances of all object categories occurring in all images. 
We hope this dataset will be broadly useful. 

Furthermore, since densely annotating the dataset is a tedious and costly task; 
we proposed a set of diagnostic tools to plug the vulnerability of the current protocol.  

Fortunately, we find that none of existing proposal methods seem to be biased, 
most of the existing algorithms 
and do generalize well to different datasets and in our experiments even on densely annotated datasets. 
In that sense, our findings are consistent with results in \cite{Hosang2015}. 
However, that should not prevent us from recognizing and safeguarding against 
the flaws in the protocol, lest we over-fit as a community to a specific set of object classes. 

\nocite{DBLP:journals/corr/ErhanSTA13,kk-lpo-15,DBLP:journals/corr/RenHG015,DBLP:journals/corr/PinheiroCD15}
\newpage
\section{Appendix}
The main paper demonstrated how the object proposal evaluation protocol is `gameable' and performed some experiments to detect this `gameability'. In this supplement, we present additional details and results which support the arguments presented in the main paper.\\
In section \ref{related_work}, we list and briefly describe the different object proposal algorithms which we used for our experiments. Following  this, details of instance-level PASCAL Context are discussed in section \ref{context}. Then we present the results on nearly-fully annotated  dataset, cross dataset evaluation on other evaluation metrics in section \ref{more_results}. We also show the per category performance of various methods on MS COCO and PASCAL Context in section \ref{percat}.

\vspace{\sectionReduceTop}
\subsection{Overview of Object Proposal Algorithms}
\vspace{\sectionReduceTop}
Table \ref{tab:RelatedWorksTable} provides an overview of some popular object proposal algorithms. The symbol $^{*}$ indicates methods we have evaluated in this paper. Note that a majority of the approaches are learning based. 
\begin{table*}[htpb]
	\vspace*{-0.5cm}
	\hspace*{-0.5cm}
	\centering
	\begin{tabular}{| p{2.5cm} |l |l| p{3cm} | p{2cm}| p{3cm}|}
		\hline
		Method & Code Source & Approach & Learning Involved & Metric & Datasets \\ \hline
		$objectness^{*}$ & {Source code from \cite{objectnesscode}}  & Window scoring & Yes supervised, train on 6 PASCAL classes and their own custom dataset of 50 images & Recall $@$ t \(\geq\) 0.5 vs \# proposals & PASCAL VOC 07 test set, test on unseen 16 PASCAL classes\\ \hline
		$selectiveSearch^{*}$ & {Source code from \cite{selectivesearchcode}}  & Segment based & No & Recall $@$ t \(\geq\) 0.5 vs \# proposals, MABO, per class ABO & PASCAL VOC 2007 test set, PASCAL VOC 2012 train val set\\ \hline
		$rahtu^{*}$ & {Source code from\cite{rahtucode}} & Window Scoring & Yes, two stages. Learning of generic bounding box prior on PASCAL VOC 2007 train set, weights for feature combination learnt on the dataset released with \cite{objectnesscode}  & Recall $@$ t > various IoU thresholds and \# proposals, AUC &  PASCAL VOC 2007 test set\\ \hline
		$randomPrim^{*}$ & {Source code from \cite{rpcode}} & Segment based & Yes supervised, train on 6 PASCAL categories  & Recall $@$ t > various IOU thresholds using 10k and 1k proposals & Pascal VOC 2007 test set/2012 trainval set on 14 categories not used in training  \\ \hline
		$mcg^{*}$ & {Source code from \cite{mcgcode}}& Segment based & Yes & NA, only segments were evaluated & NA (tested on segmentation dataset) \\ \hline
		$edgeBoxes^{*}$ & {Source code from \cite{ebcode}} & Window scoring & No & AUC, Recall $@$ t > various IOU thresholds and \# proposals, Recall vs IoU & PASCAL VOC 2007 testset\\ \hline 
		$bing^{*}$ & {Source code from \cite{bingcode}} & Window scoring & Yes supervised, on PASCAL VOC 2007 train set, 20 object classes/6 object classes & Recall $@$ t> 0.5 vs \# proposals & PASCAL VOC 2007 detection complete test set$/$14 unseen object categories \\ \hline
		$rantalankila$ & {Source code from \cite{rantalankilacode}} & Segment based & Yes & NA, only segments are evaluated & NA (tested on segmentation dataset)\\ \hline
		$Geodesic$ & {Source code from \cite{gopcode}} & Segment based & Yes, for seed placement and mask construction  on PASCAL VOC 2012 Segmentation training set&VUS at 10k and 2k windows, Recall vs IoU threshold, Recall vs proposals  & PASCAL 2012 detection validation set \\ \hline
		$Rigor$ & {Source code from \cite{rigorcode}} & Segment based & Yes, pairwise potentials between super pixels learned on BSDS-500 boundary detection dataset & NA, only segments were evaluated & NA (tested on segmentation dataset)\\ \hline
		$endres$ & {Source code from \cite{endrescode}} & Segment based & Yes & NA, only segments are evaluated & NA  (tested on segmentation dataset) \\ \hline
	\end{tabular}
	\captionof{table}{Properties of existing bounding box approaches. * indicates the methods which have studied in this paper.} \label{tab:RelatedWorksTable} 
\end{table*}
\label{related_work}
\vspace{\sectionReduceTop}
\subsection{Details of PASCAL Context Annotation}
\vspace{\sectionReduceTop}
As explained in section 5.1 of the main paper, PASCAL Context provides full annotations for PASCAL VOC 2010 dataset in the form of semantic segmentations. A total of 459 classes have labeled in this dataset. We split these into three categories namely Objects/Things, Background/Stuff and Ambiguous as shown in Tables \ref{contextobjects}, \ref{contextnonobjects} and \ref{contextamb}.  Most classes (396) were put in the `Objects' category. 20 of these are PASCAL categories. Of the remaining 376, we selected the most frequently occurring 60 categories and manually created instance level annotations for the same.

\textbf{Statistics of New Annotations:} 
 We made the following observations on our new annotations: 
\begin{compactitem}
	\item The number of instances we annotated for the extra 60 categories were about the same as the number of instances for annotated for 20 PASCAL categories in the original PASCAL VOC. This shows that about half the annotations were missing and thus a lot of genuine proposal candidates are not being rewarded.
	\item Most non-PASCAL categories
	occupy a small percentage of the image. This is understandable given that the dataset was curated with these categories. The other categories just happened to be in the pictures.
\end{compactitem}

\label{context}
\begin{table*}[!htbp]
\resizebox{\textwidth}{!}{%
\begin{tabular}{@{}|llllllll|l@{}}
\cmidrule(r){1-8}
\multicolumn{8}{|c|}{\textbf{Object/Thing Classes in PASCAL Context Dataset}}                                                                            & \multicolumn{1}{c}{} \\ \cmidrule(r){1-8}
accordion           & candleholder    & drainer            & funnel              & lightbulb         & pillar               & sheep              & tire             &                      \\
aeroplane           & cap             & dray               & furnace             & lighter           & pillow               & shell              & toaster          &                      \\
airconditioner      & car             & drinkdispenser     & gamecontroller      & line              & pipe                 & shoe               & toilet           &                      \\
antenna             & card            & drinkingmachine    & gamemachine         & lion              & pitcher              & shoppingcart       & tong             &                      \\
ashtray             & cart            & drop               & gascylinder         & lobster           & plant                & shovel             & tool             &                      \\
babycarriage        & case            & drug               & gashood             & lock              & plate                & sidecar            & toothbrush       &                      \\
bag                 & casetterecorder & drum               & gasstove            & machine           & player               & sign               & towel            &                      \\
ball                & cashregister    & drumkit            & giftbox             & mailbox           & pliers               & signallight        & toy              &                      \\
balloon             & cat             & duck               & glass               & mannequin         & plume                & sink               & toycar           &                      \\
barrel              & cd              & dumbbell           & glassmarble         & map               & poker                & skateboard         & train            &                      \\
baseballbat         & cdplayer        & earphone           & globe               & mask              & pokerchip            & ski                & trampoline       &                      \\
basket              & cellphone       & earrings           & glove               & mat               & pole                 & sled               & trashbin         &                      \\
basketballbackboard & cello           & egg                & gravestone          & matchbook         & pooltable            & slippers           & tray             &                      \\
bathtub             & chain           & electricfan        & guitar              & mattress          & postcard             & snail              & tricycle         &                      \\
bed                 & chair           & electriciron       & gun                 & menu              & poster               & snake              & tripod           &                      \\
beer                & chessboard      & electricpot        & hammer              & meterbox          & pot                  & snowmobiles        & trophy           &                      \\
bell                & chicken         & electricsaw        & handcart            & microphone        & pottedplant          & sofa               & truck            &                      \\
bench               & chopstick       & electronickeyboard & handle              & microwave         & printer              & spanner            & tube             &                      \\
bicycle             & clip            & engine             & hanger              & mirror            & projector            & spatula            & turtle           &                      \\
binoculars          & clippers        & envelope           & harddiskdrive       & missile           & pumpkin              & speaker            & tvmonitor        &                      \\
bird                & clock           & equipment          & hat                 & model             & rabbit               & spicecontainer     & tweezers         &                      \\
birdcage            & closet          & extinguisher       & headphone           & money             & racket               & spoon              & typewriter       &                      \\
birdfeeder          & cloth           & eyeglass           & heater              & monkey            & radiator             & sprayer            & umbrella         &                      \\
birdnest            & coffee          & fan                & helicopter          & mop               & radio                & squirrel           & vacuumcleaner    &                      \\
blackboard          & coffeemachine   & faucet             & helmet              & motorbike         & rake                 & stapler            & vendingmachine   &                      \\
board               & comb            & faxmachine         & holder              & mouse             & ramp                 & stick              & videocamera      &                      \\
boat                & computer        & ferriswheel        & hook                & mousepad          & rangehood            & stickynote         & videogameconsole &                      \\
bone                & cone            & fireextinguisher   & horse               & musicalinstrument & receiver             & stone              & videoplayer      &                      \\
book                & container       & firehydrant        & horse-drawncarriage & napkin            & recorder             & stool              & videotape        &                      \\
bottle              & controller      & fireplace          & hot-airballoon      & net               & recreationalmachines & stove              & violin           &                      \\
bottleopener        & cooker          & fish               & hydrovalve          & newspaper         & remotecontrol        & straw              & wakeboard        &                      \\
bowl                & copyingmachine  & fishtank           & inflatorpump        & oar               & robot                & stretcher          & wallet           &                      \\
box                 & cork            & fishbowl           & ipod                & ornament          & rock                 & sun                & wardrobe         &                      \\
bracelet            & corkscrew       & fishingnet         & iron                & oven              & rocket               & sunglass           & washingmachine   &                      \\
brick               & cow             & fishingpole        & ironingboard        & oxygenbottle      & rockinghorse         & sunshade           & watch            &                      \\
broom               & crabstick       & flag               & jar                 & pack              & rope                 & surveillancecamera & waterdispenser   &                      \\
brush               & crane           & flagstaff          & kart                & pan               & rug                  & swan               & waterpipe        &                      \\
bucket              & crate           & flashlight         & kettle              & paper             & ruler                & sweeper            & waterskateboard  &                      \\
bus                 & cross           & flower             & key                 & paperbox          & saddle               & swimring           & watermelon       &                      \\
cabinet             & crutch          & fly                & keyboard            & papercutter       & saw                  & swing              & whale            &                      \\
cabinetdoor         & cup             & food               & kite                & parachute         & scale                & switch             & wheel            &                      \\
cage                & curtain         & forceps            & knife               & parasol           & scanner              & table              & wheelchair       &                      \\
cake                & cushion         & fork               & knifeblock          & pen               & scissors             & tableware          & window           &                      \\
calculator          & cuttingboard    & forklift           & ladder              & pencontainer      & scoop                & tank               & windowblinds     &                      \\
calendar            & disc            & fountain           & laddertruck         & pencil            & screen               & tap                & wineglass        &                      \\
camel               & disccase        & fox                & ladle               & person            & screwdriver          & tape               & wire             &                      \\
camera              & dishwasher      & frame              & laptop              & photo             & sculpture            & tarp               &                  &                      \\
cameralens          & dog             & fridge             & lid                 & piano             & scythe               & telephone          &                  &                      \\
can                 & dolphin         & frog               & lifebuoy            & picture           & sewer                & telephonebooth     &                  &                      \\
candle              & door            & fruit              & light               & pig               & sewingmachine        & tent               &                  &                      \\ \bottomrule
\end{tabular}
}
\captionof{table}{Object/Thing Classes in PASCAL Context
\label{contextobjects}}
\end{table*}
\begin{table}[!htbp]
\begin{tabular}{|llll|}
\hline
\multicolumn{4}{|c|}{\textbf{Ambiguous Classes in PASCAL Context Dataset}} \\ \hline
artillery        & escalator           & ice         & speedbump     \\
bedclothes       & exhibitionbooth     & leaves      & stair         \\
clothestree      & flame               & outlet      & tree          \\
coral            & guardrail           & rail        & unknown       \\
dais             & handrail            & shelves     &               \\ \hline
\end{tabular}
\captionof{table}{Ambiguous Classes in PASCAL Context}
\label{contextamb}
\end{table}
\begin{table}[!htbp]
\resizebox{\columnwidth}{!}{%
\begin{tabular}{|llll|}
\hline
\multicolumn{4}{|c|}{\textbf{Background/Stuff Classes in PASCAL Context Dataset}} \\ \hline
atrium              & floor              & parterre        & sky               \\
bambooweaving       & foam               & patio           & smoke             \\
bridge              & footbridge         & pelage          & snow              \\
building            & goal               & plastic         & stage             \\
ceiling             & grandstand         & platform        & swimmingpool      \\
concrete            & grass              & playground      & track             \\
controlbooth        & ground             & road            & wall              \\
counter             & hay                & runway          & water             \\
court               & kitchenrange       & sand            & wharf             \\
dock                & metal              & shed            & wood              \\
fence               & mountain           & sidewalk        & wool              \\ \hline
\end{tabular}
}
\captionof{table}{Background/Stuff Classes in PASCAL Context}
\label{contextnonobjects}
\end{table}
\vspace{\sectionReduceTop}
\subsection{Evaluation of Proposals on Other Metrics}
\vspace{\sectionReduceTop}
In this section, we show the performance of different proposal methods and DMPs on MS COCO dataset on various metrics. \figref{20coco1000} shows performance on Recall-vs-IOU metric at 1000 \#proposals on PASCAL 20 categories. \figref{20coco05}, \figref{20coco07} show performance on Recall-\vs-\#proposals metric at  0.5 and 0.7 IOU respectively.
Similarly in Figs. \ref{70coco1000},\ref{70coco05}, \ref{70coco07} and Figs. \ref{90coco1000},\ref{90coco05}, \ref{90coco07}, we can see the performance of all proposal methods and DMPs on these three metrics where 60 non-PASCAL and all categories respectively are annotated in the MS COCO dataset. \\
These metrics also demonstrate the same trend as shown by the AUC-\vs-\#proposals in the main paper. When only PASCAL categories are annotated (Figs. \ref{20coco1000},\ref{20coco05}, \ref{20coco07} ), DMPs outperform all proposal methods. However, when other categories are also annotated (Figs. \ref{90coco1000},\ref{90coco05}, \ref{90coco07}) or the performance is evaluated specifically on the other categories (Figs. \ref{70coco1000},\ref{70coco05}, \ref{70coco07}), DMPs cease to be the top performers.\\
Finally, we also report results on different metrics PASCAL Context (\figref{allcat_context}) and NYU-Depth v2 (\figref{nyu2}). They also show similar trends, supporting the claims made in the paper.
\begin{figure*}[htb]
	\centering
	\begin{subfigure}[b]{0.3\textwidth}
		\includegraphics[width=1.0\columnwidth]{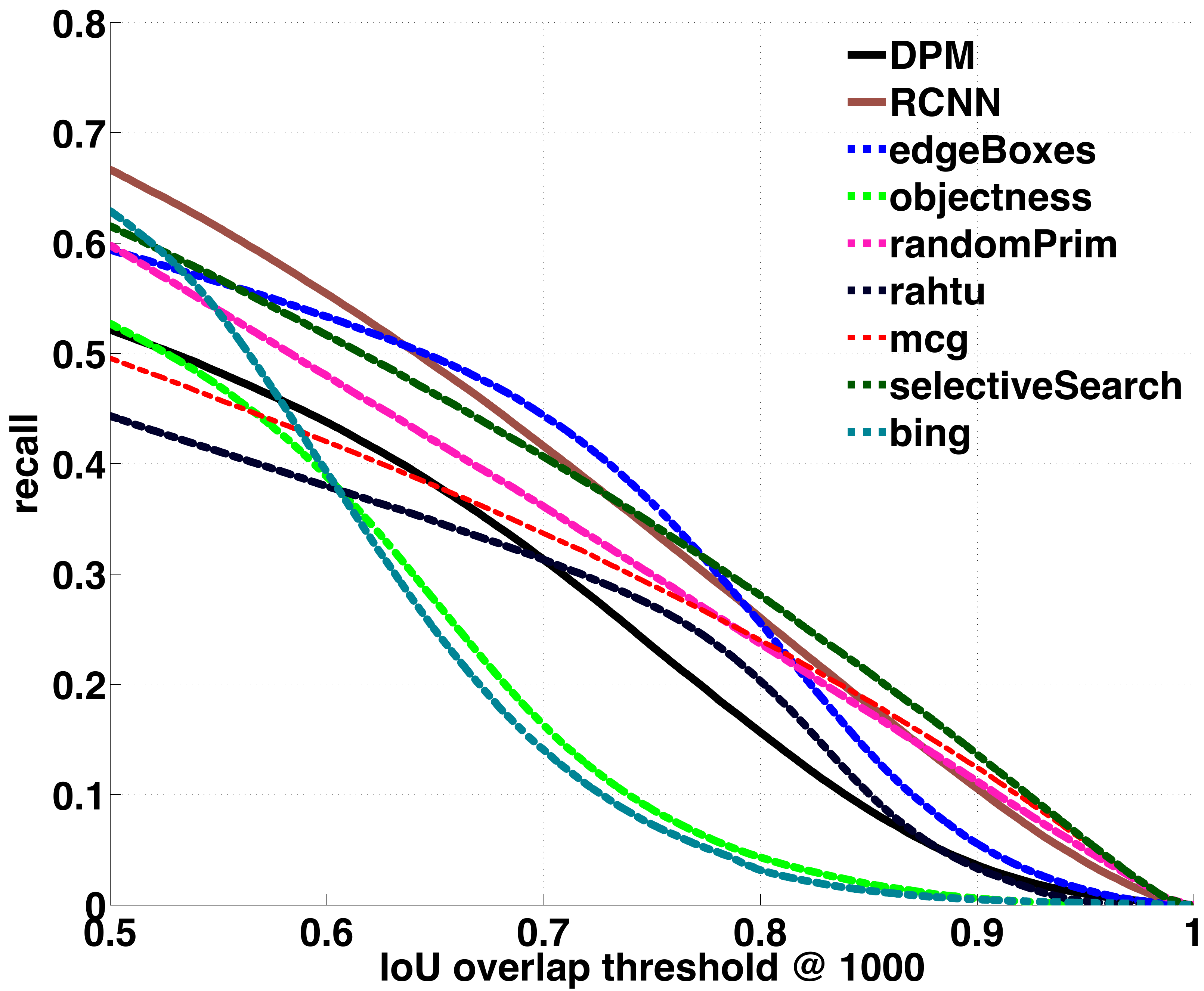}
		\subcaption{\label{20coco1000}Recall vs IOU at 1000 proposals for 20 PASCAL categories annotated in MS COCO validation dataset}
	\end{subfigure}
	\,\,\,
	\begin{subfigure}[b]{0.3\textwidth}
		\includegraphics[width=1.0\columnwidth]{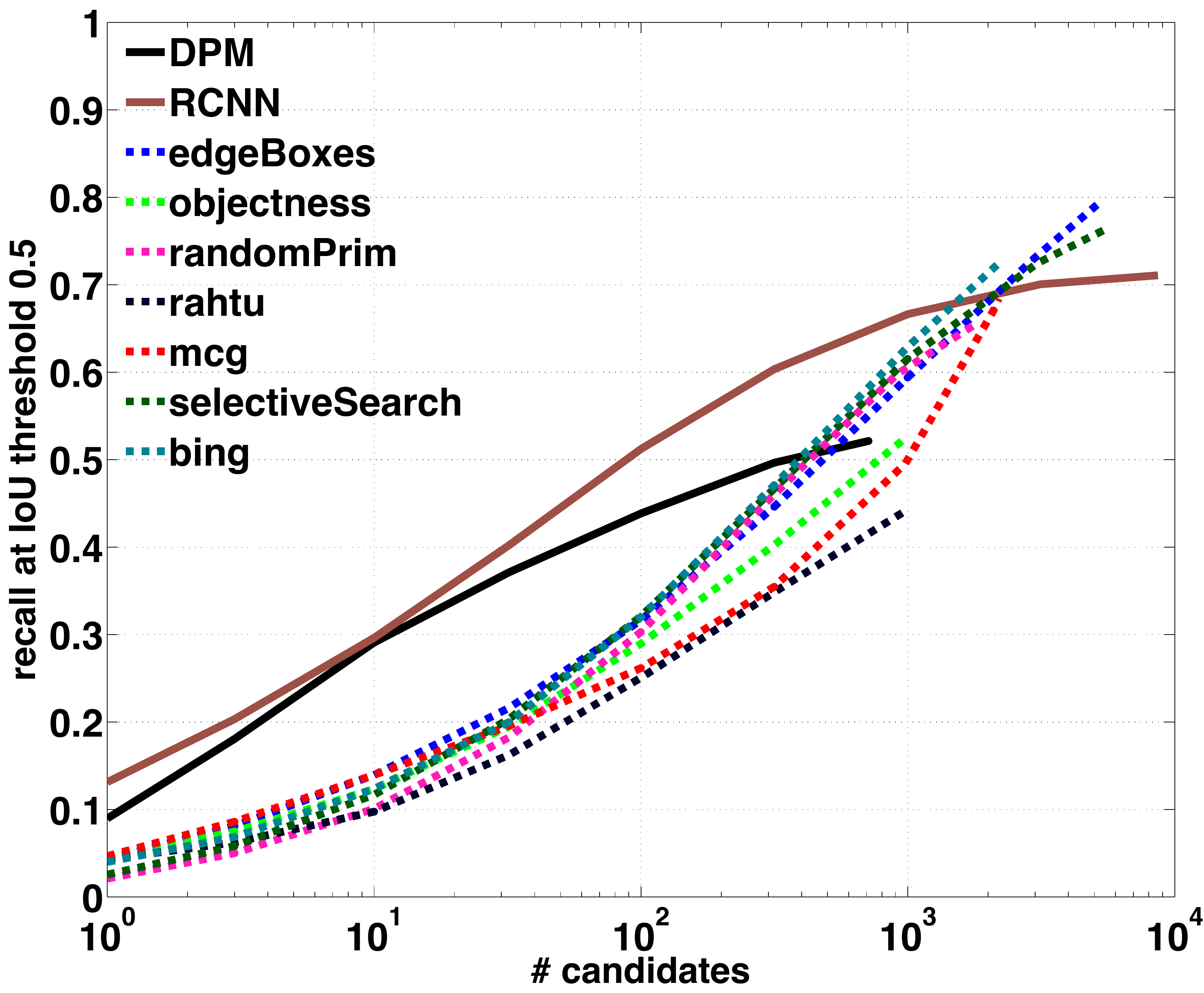}
		\subcaption{\label{20coco05} Recall \vs number of proposals at 0.5 IOU for 20 PASCAL categories annotated in MS COCO validation dataset}  
	\end{subfigure}
	\,\,\,
	\begin{subfigure}[b]{0.3\textwidth}
		\includegraphics[width=1.0\columnwidth]{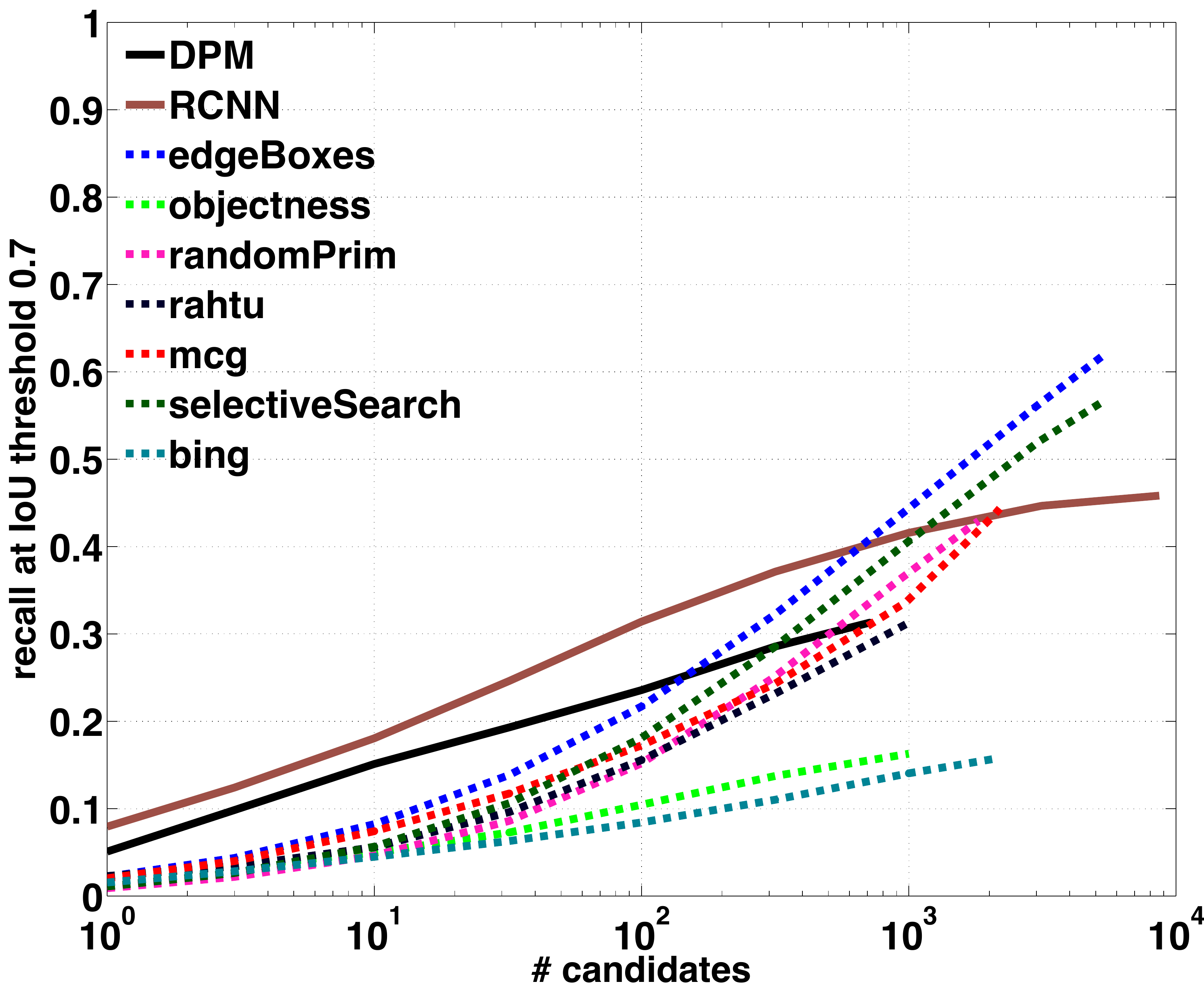}
		\subcaption{\label{20coco07}  Recall \vs number of proposals at 0.7 IOU for 20 PASCAL categories annotated in MS COCO validation dataset} 
	\end{subfigure}
	\label{p_coco}
	\centering
	\begin{subfigure}[b]{0.3\textwidth}
		\includegraphics[width=1.0\columnwidth]{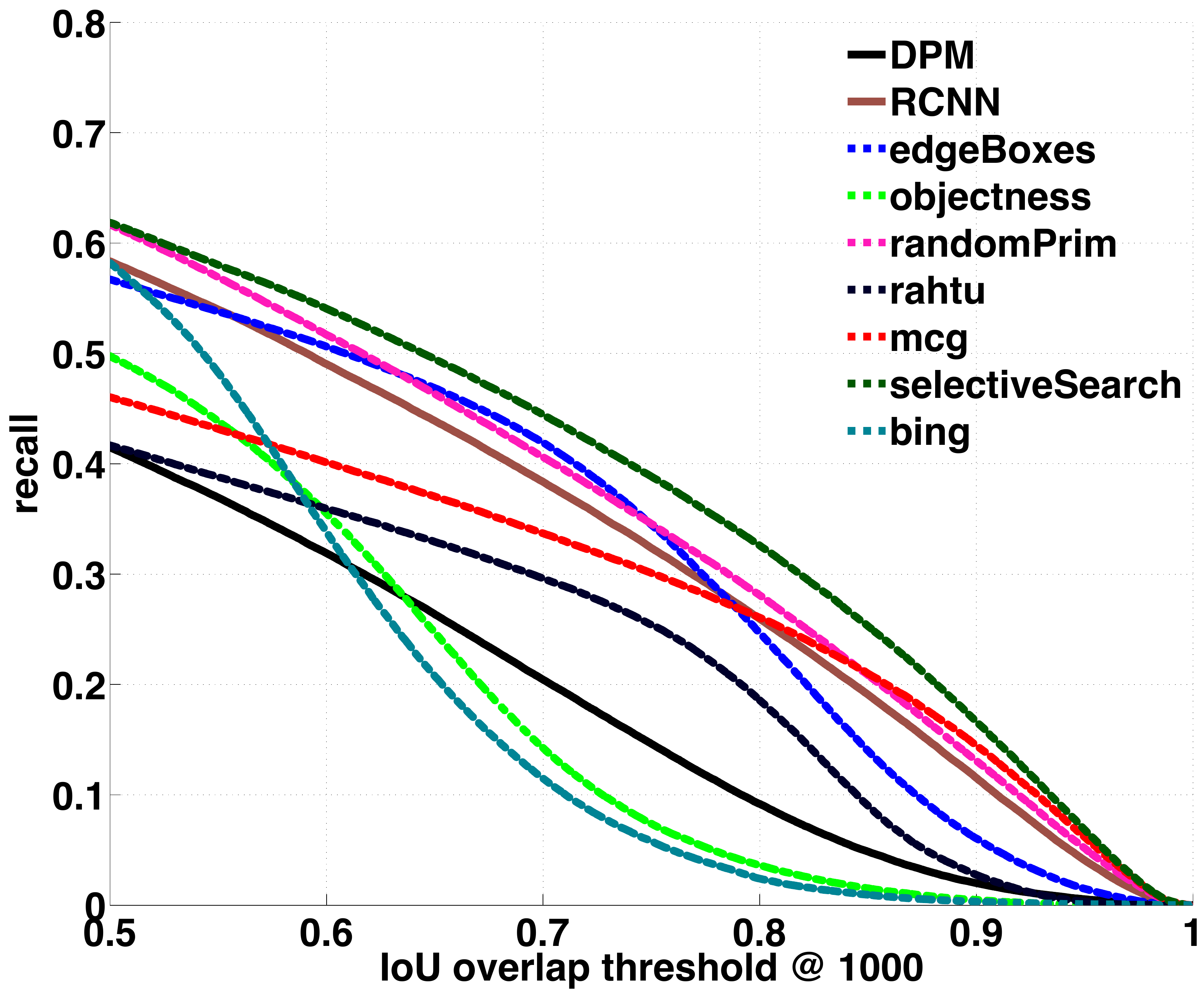}
		\subcaption{\label{70coco1000}Recall vs IOU at 1000 proposals for 60 non-PASCAL categories annotated in MS COCO validation dataset}
	\end{subfigure}
	\,\,\,
	\begin{subfigure}[b]{0.3\textwidth}
		\includegraphics[width=1.0\columnwidth]{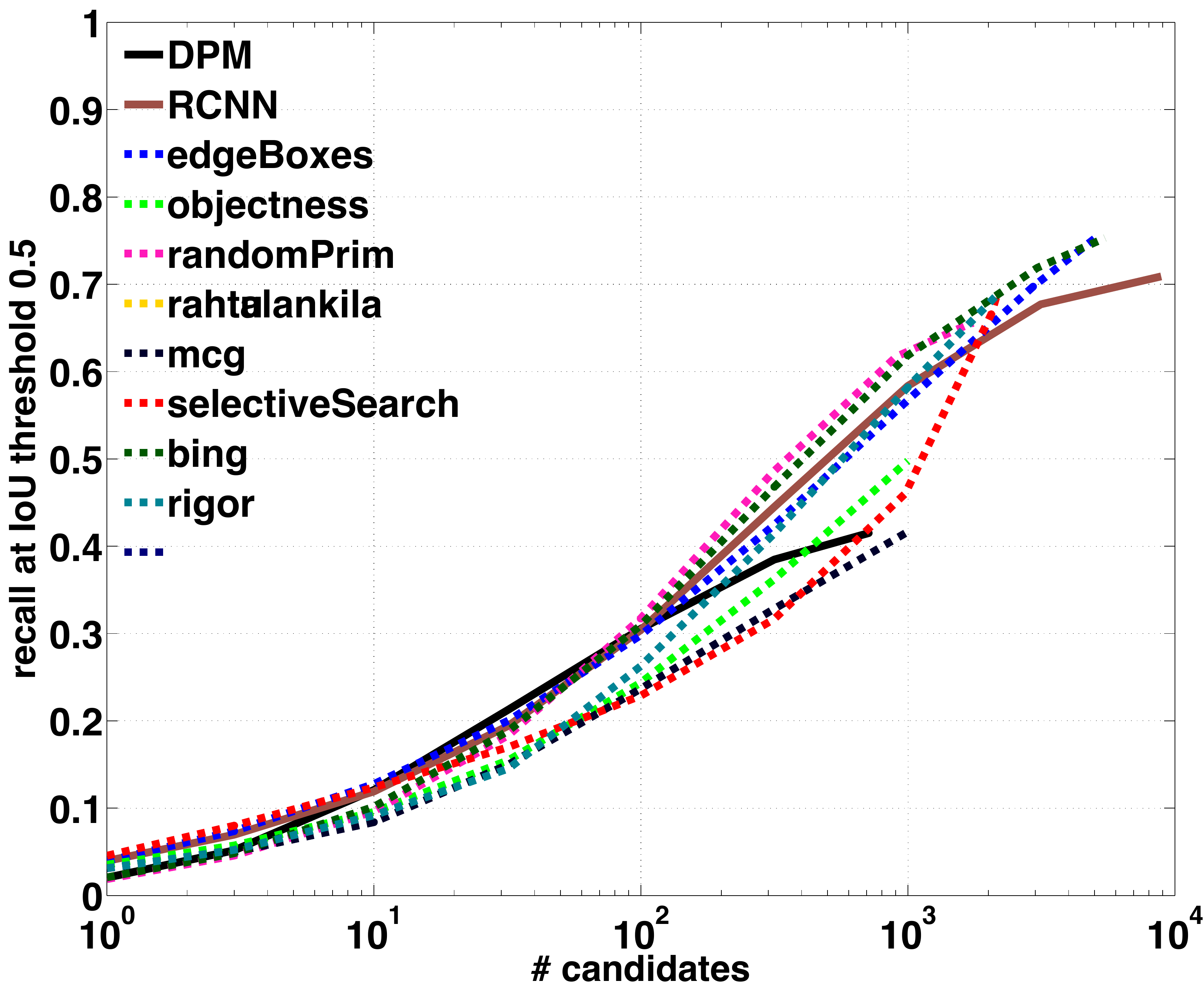}
		\subcaption{\label{70coco05}Recall \vs number of proposals at 0.5 IOU for 60 non-PASCAL categories annotated in MS COCO validation dataset} 
	\end{subfigure}
	\,\,\,
	\begin{subfigure}[b]{0.3\textwidth}
		\includegraphics[width=1.0\columnwidth]{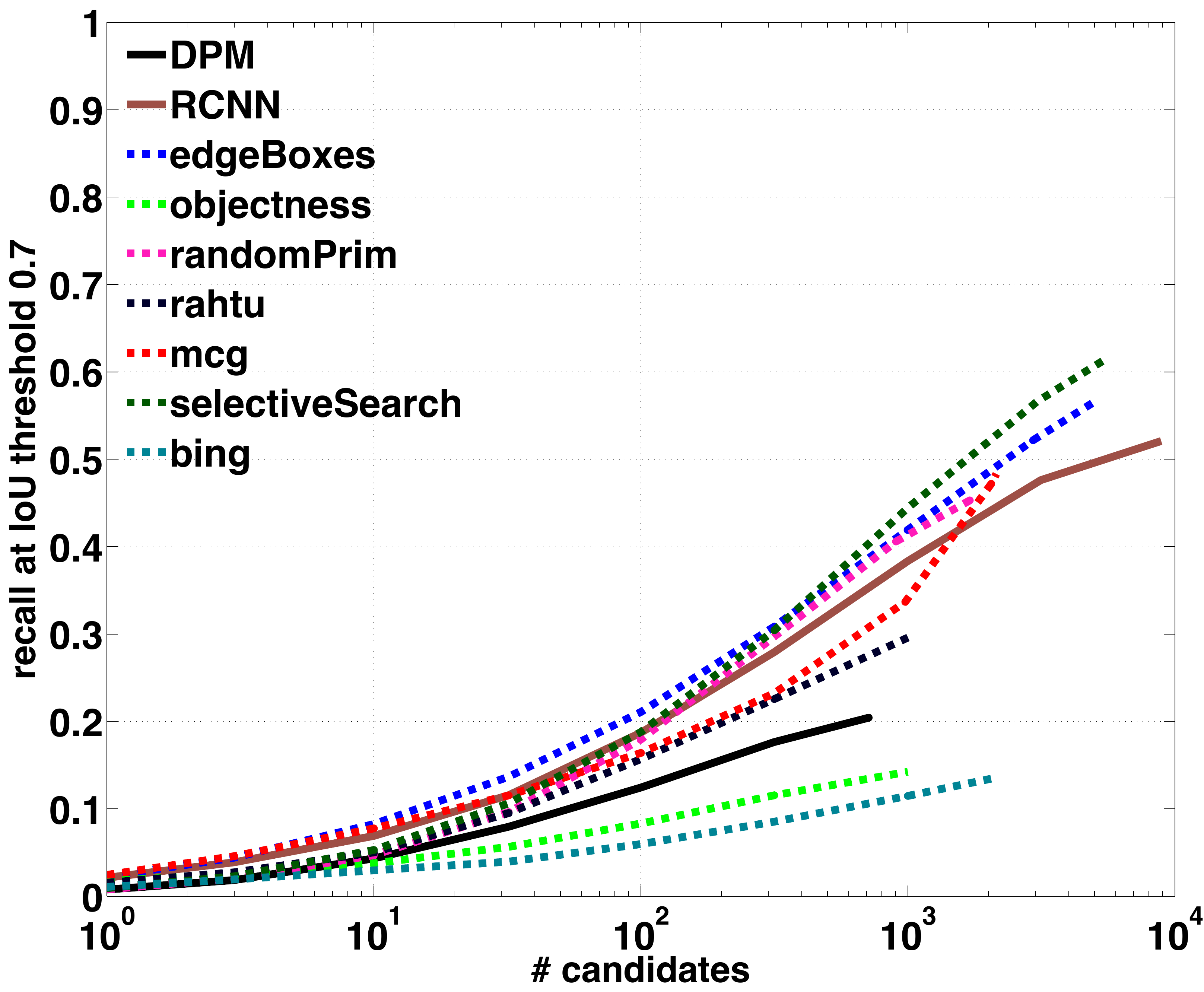}
		\subcaption{\label{70coco07}Recall \vs number of proposals at 0.7 IOU for 60 non-PASCAL categories annotated in MS COCO validation dataset} 
	\end{subfigure}
	\label{np_coco}
	\centering
	\begin{subfigure}[b]{0.3\textwidth}
		\includegraphics[width=1.0\columnwidth]{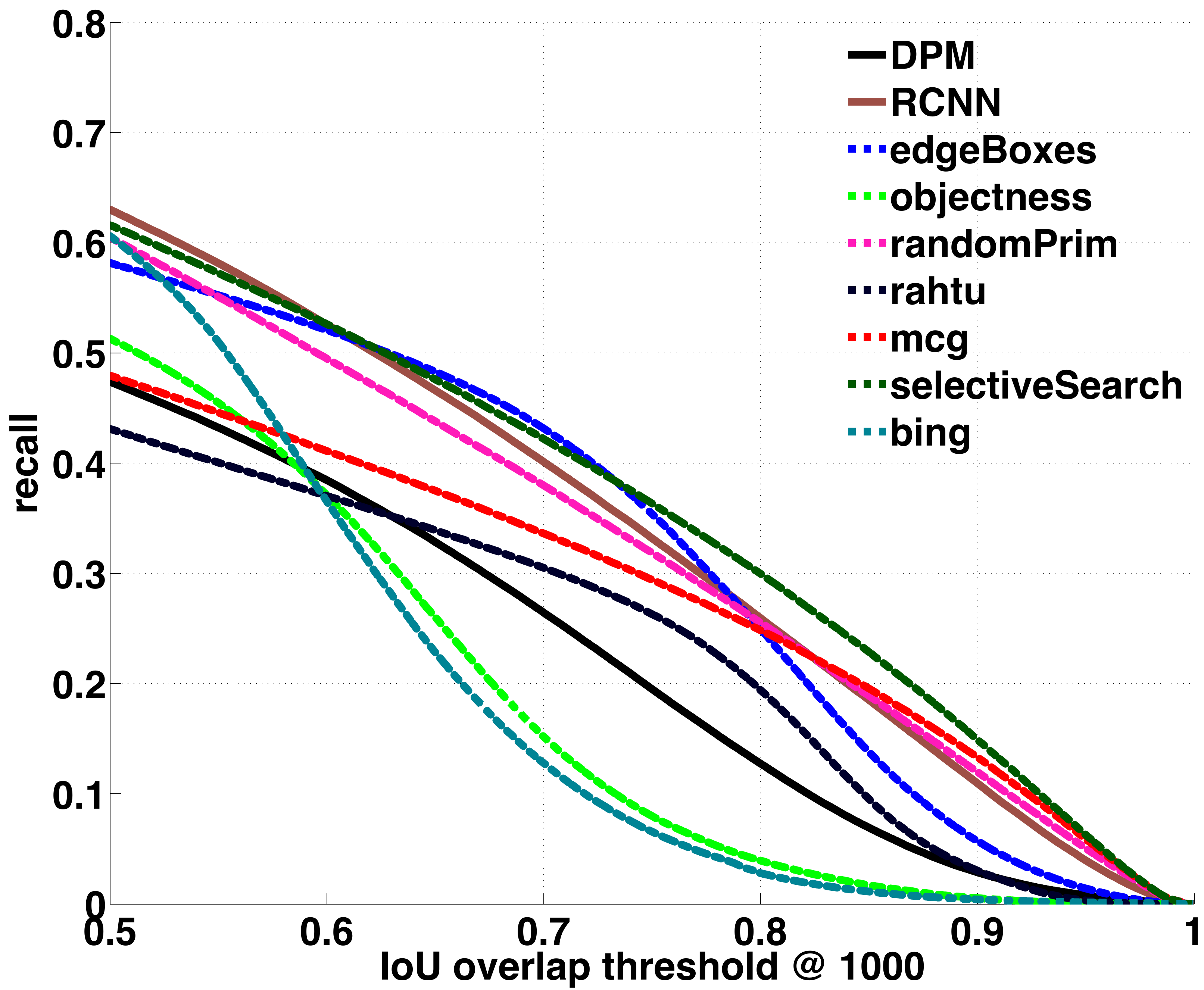}
		\subcaption{\label{90coco1000}Recall vs IOU at 1000 proposals for all categories annotated in MS COCO validation dataset}
	\end{subfigure}
	\,\,\,
	\begin{subfigure}[b]{0.3\textwidth}
		\includegraphics[width=1.0\columnwidth]{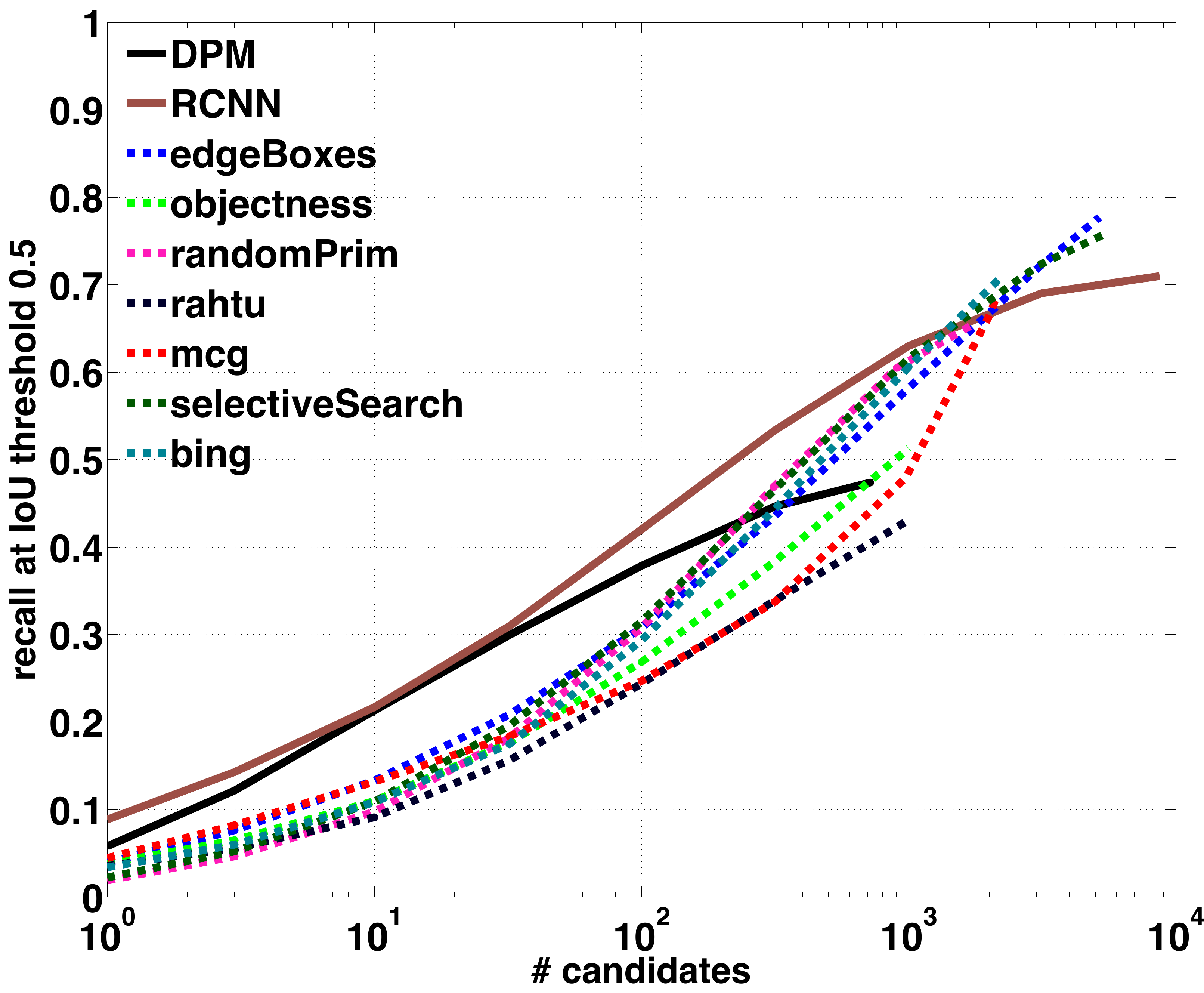}
		\subcaption{\label{90coco05}Recall \vs number of proposals at 0.5 IOU for all categories annotated in MS COCO validation dataset} 
	\end{subfigure}
	\,\,\,
	\begin{subfigure}[b]{0.3\textwidth}
		\includegraphics[width=1.0\columnwidth]{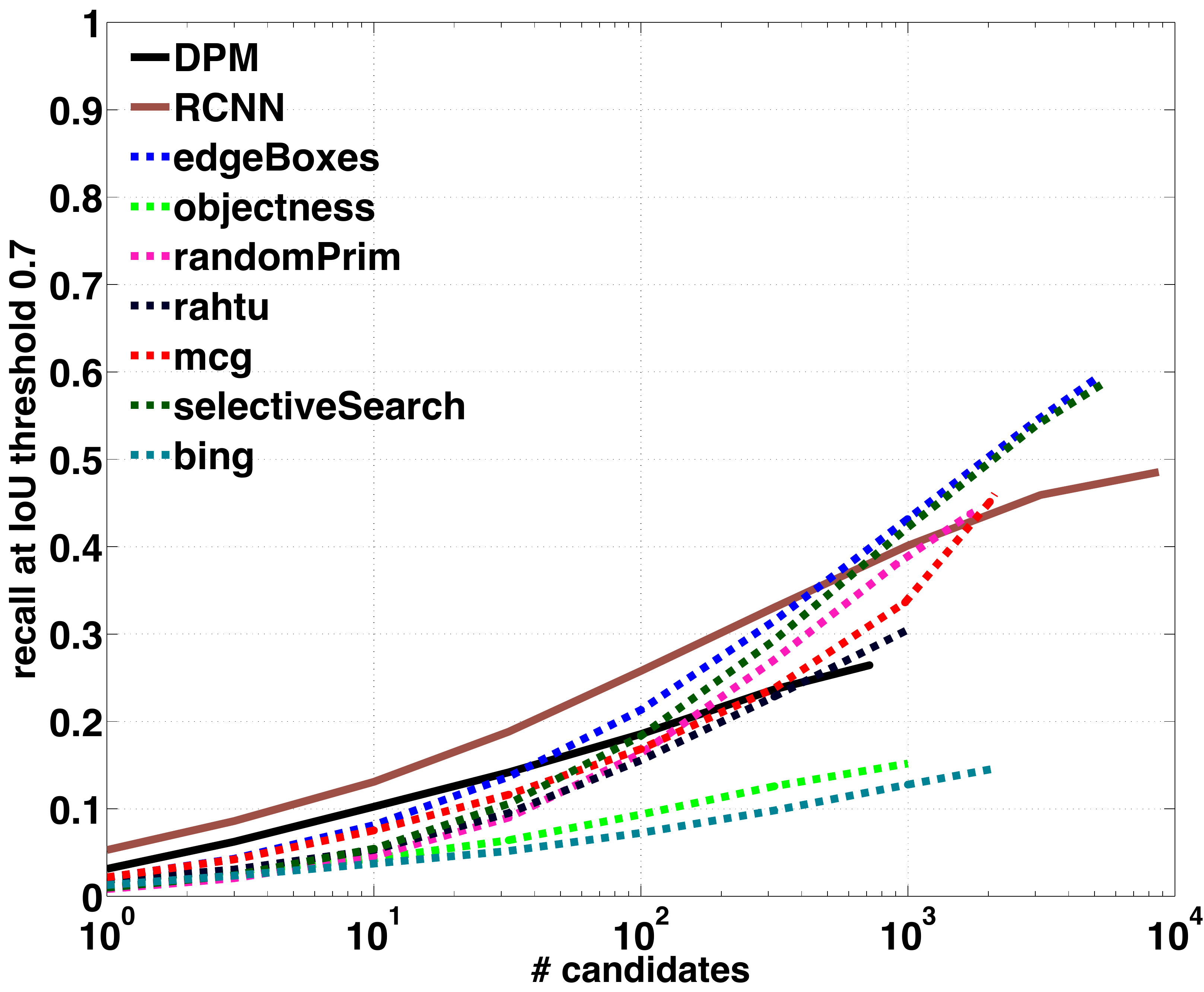}
		\subcaption{\label{90coco07}Recall \vs number of proposals at 0.7 IOU for all categories annotated in MS COCO validation dataset} 
	\end{subfigure}
	\caption{\label{allcat_coco}Performance of various object proposal methods on different evaluation metrics when evaluated on MS COCO dataset.}
\end{figure*}
\begin{figure*}
	\centering
	\begin{subfigure}[b]{0.3\textwidth}
		\includegraphics[width=1.0\columnwidth]{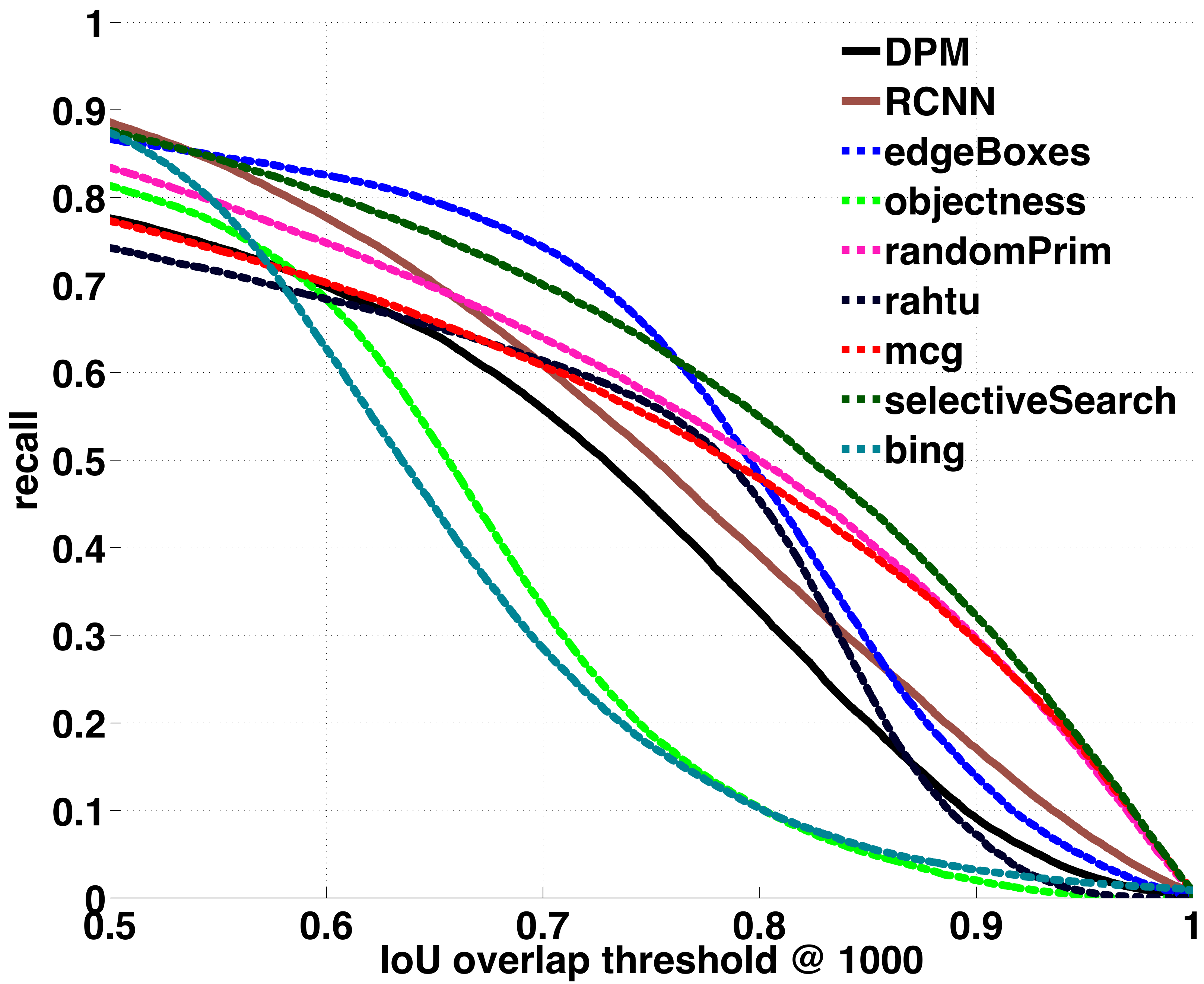}
		\subcaption{\label{20context1000}Recall vs IOU at 1000 proposals for 20 PASCAL categories annotated in PASCAL Context dataset}
	\end{subfigure}
	\,\,\,
	\begin{subfigure}[b]{0.3\textwidth}
		\includegraphics[width=1.0\columnwidth]{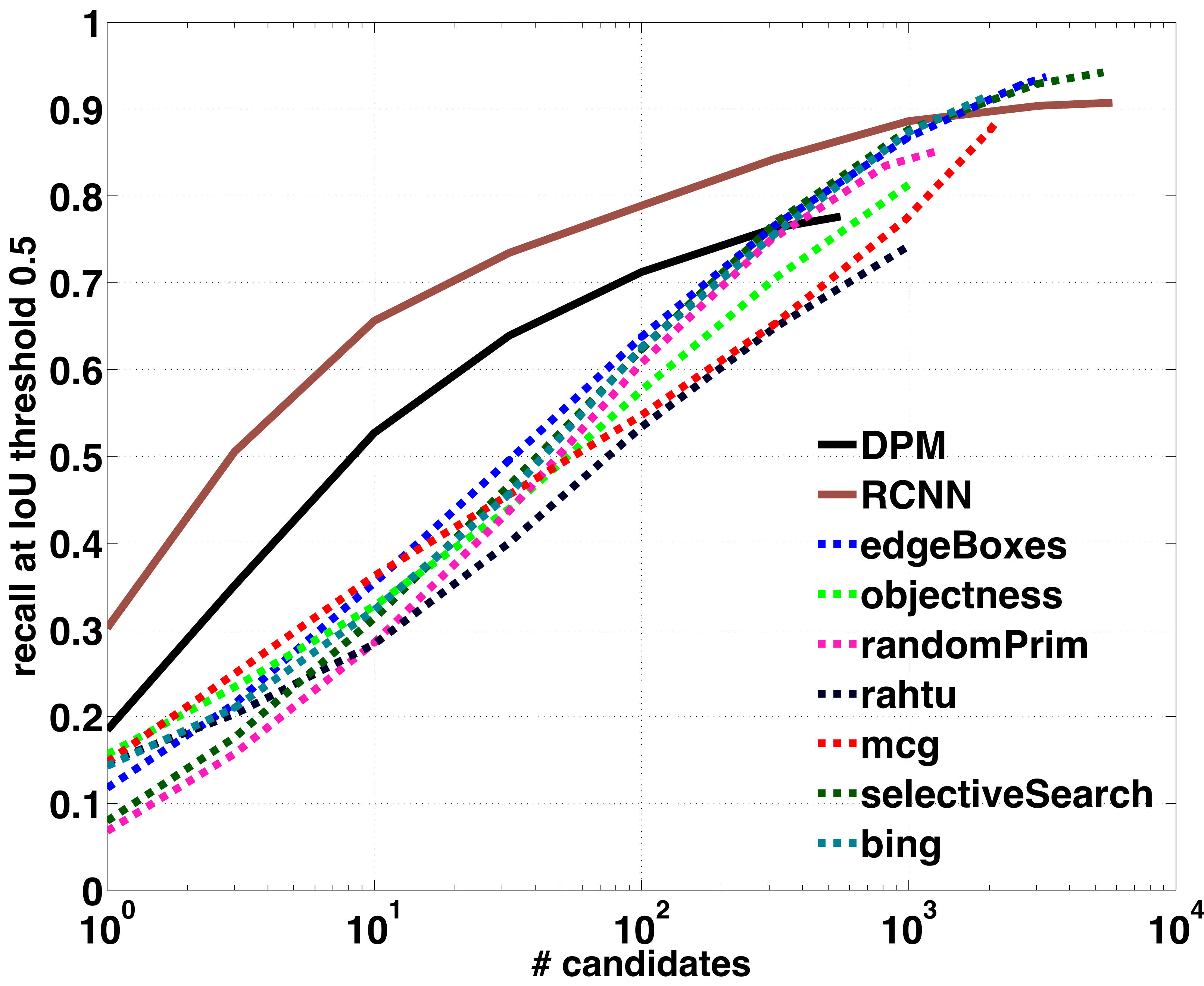}
		\subcaption{\label{20context05}Recall \vs number of proposals at 0.5 IOU for 20 PASCAL annotated in PASCAL Context dataset} 
	\end{subfigure}
	\,\,\,
	\begin{subfigure}[b]{0.3\textwidth}
		\includegraphics[width=1.0\columnwidth]{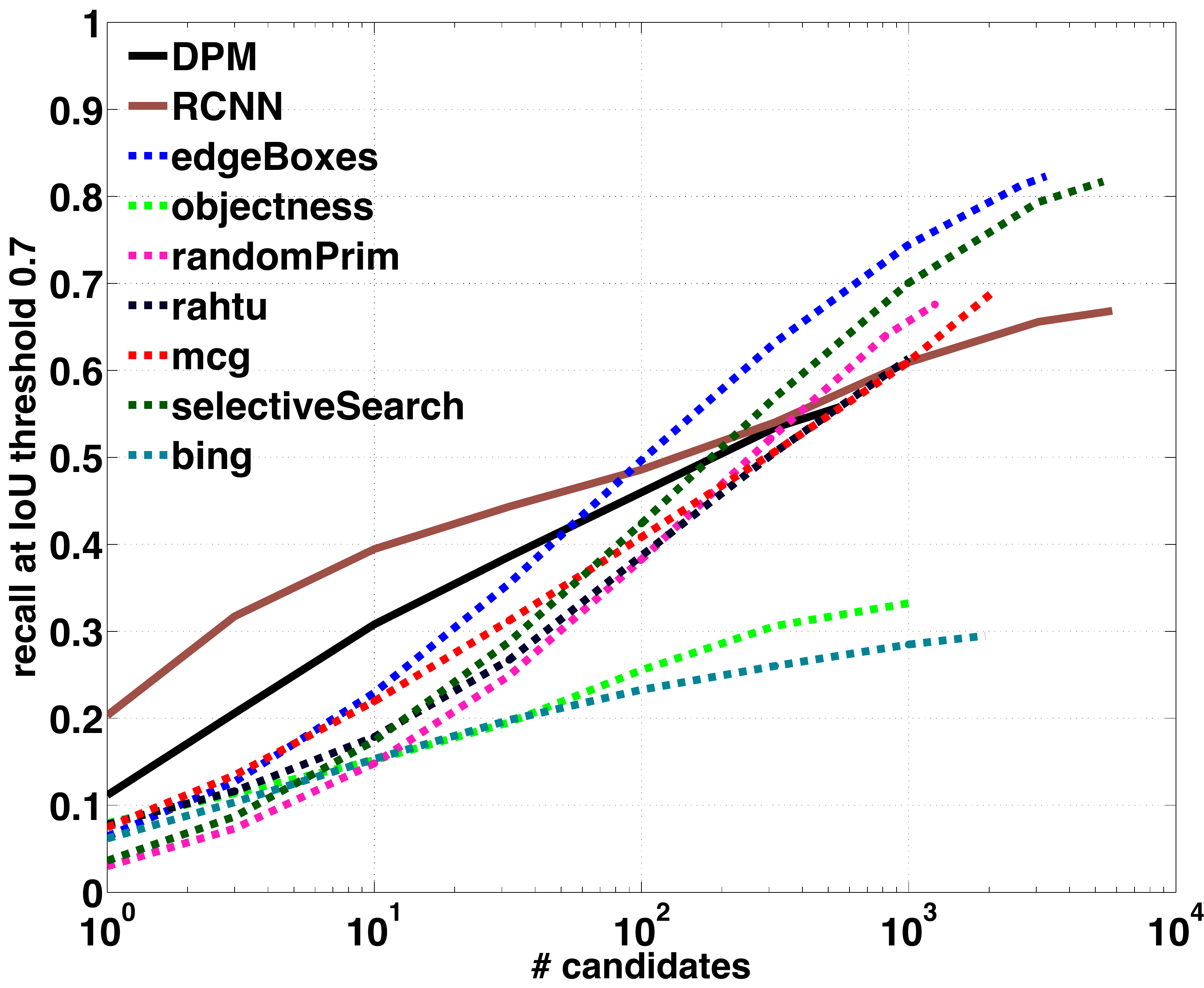}
		\subcaption{\label{20context07}Recall \vs number of proposals at 0.7 IOU for 20 PASCAL categories annotated in PASCAL Context dataset} 
	\end{subfigure}
	\label{20cat_context}
	\centering
	\begin{subfigure}[b]{0.3\textwidth}
		\includegraphics[width=1.0\columnwidth]{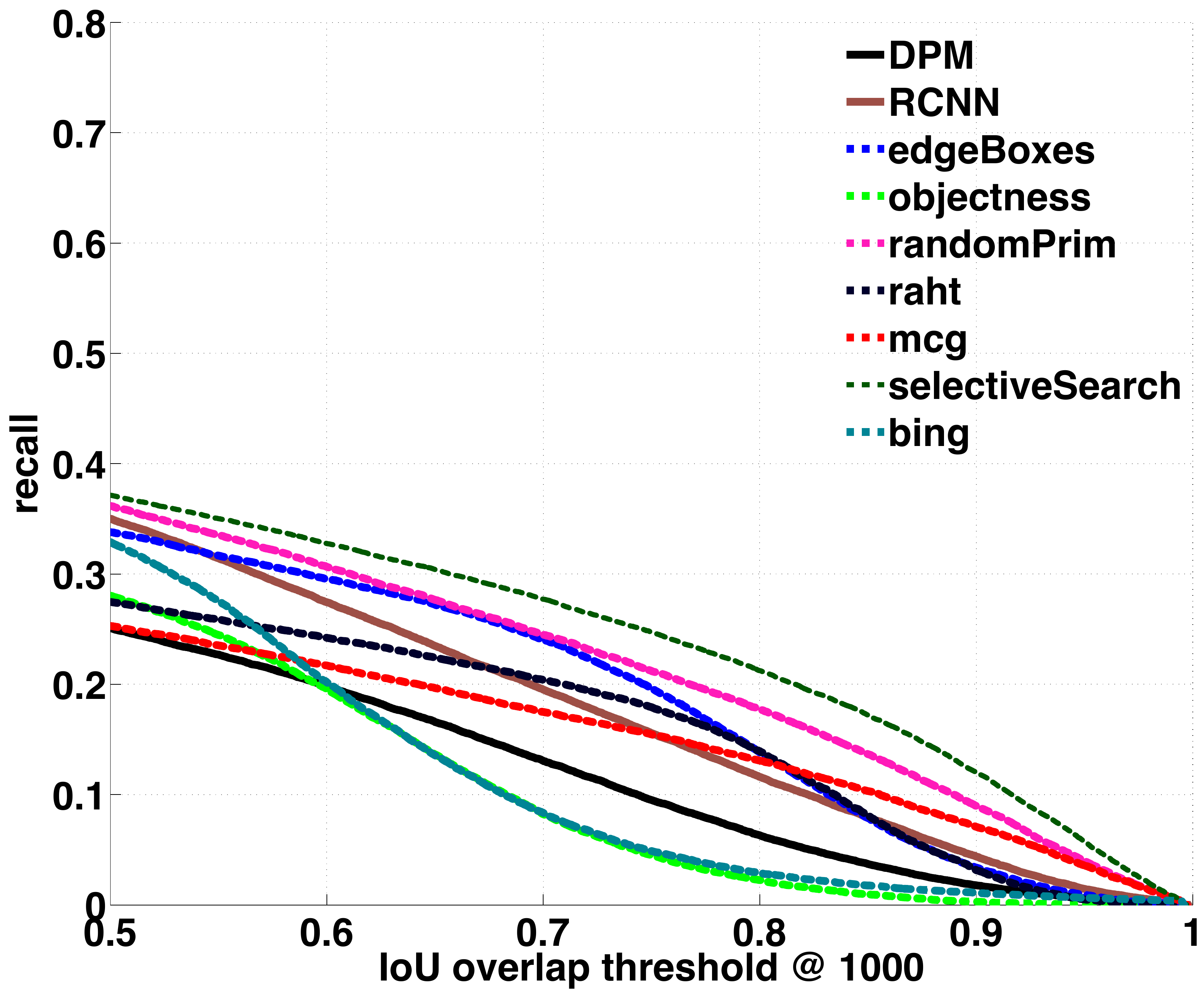}
		\subcaption{\label{60context1000}Recall vs IOU at 1000 proposals for non-PASCAL categories annotated in PASCAL Context dataset}
	\end{subfigure}
	\,\,\,
	\begin{subfigure}[b]{0.3\textwidth}
		\includegraphics[width=1.0\columnwidth]{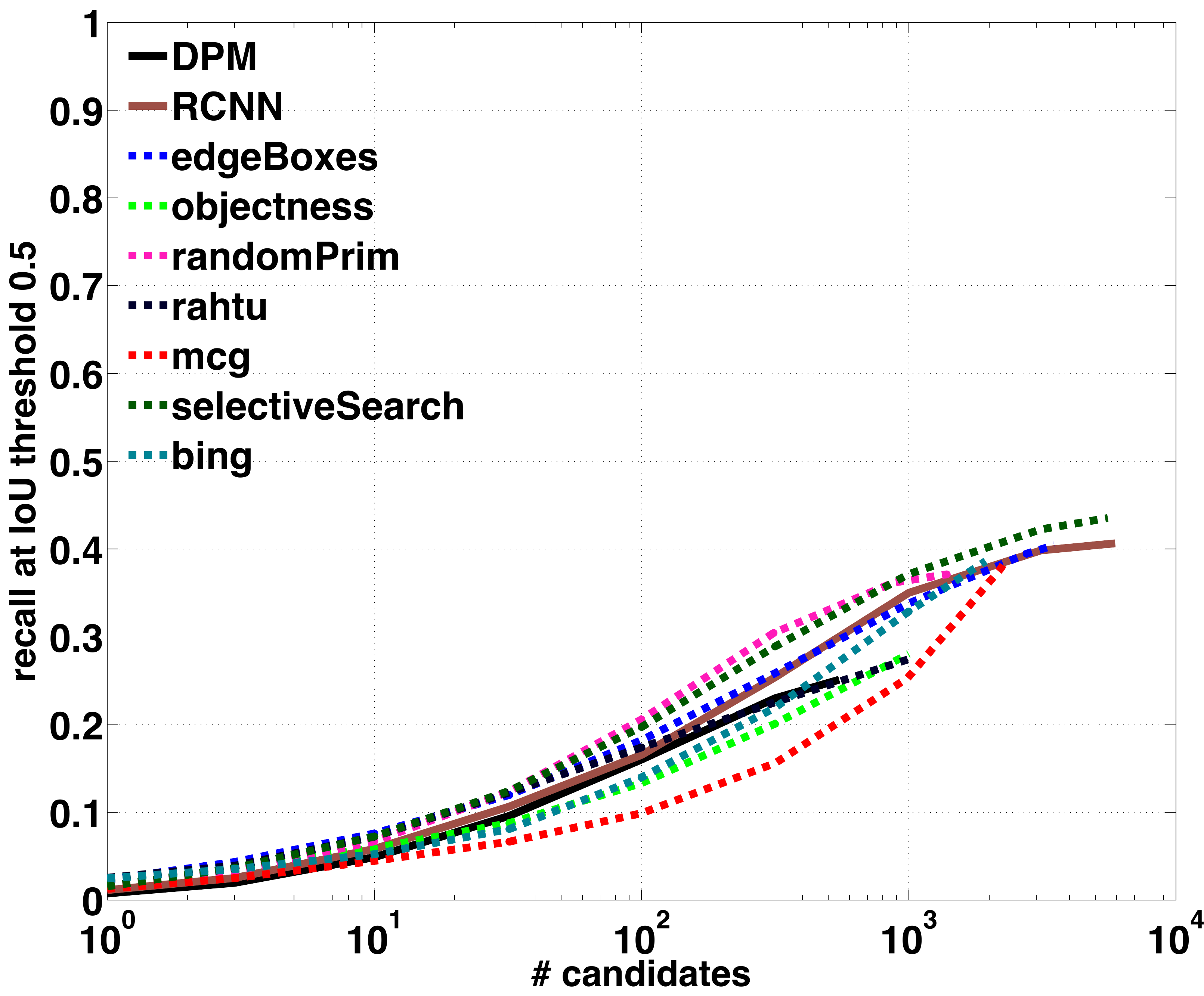}
		\subcaption{\label{60context05}Recall \vs number of proposals at 0.5 IOU for non-PASCAL annotated in PASCAL Context dataset} 
	\end{subfigure}
	\,\,\,
	\begin{subfigure}[b]{0.3\textwidth}
		\includegraphics[width=1.0\columnwidth]{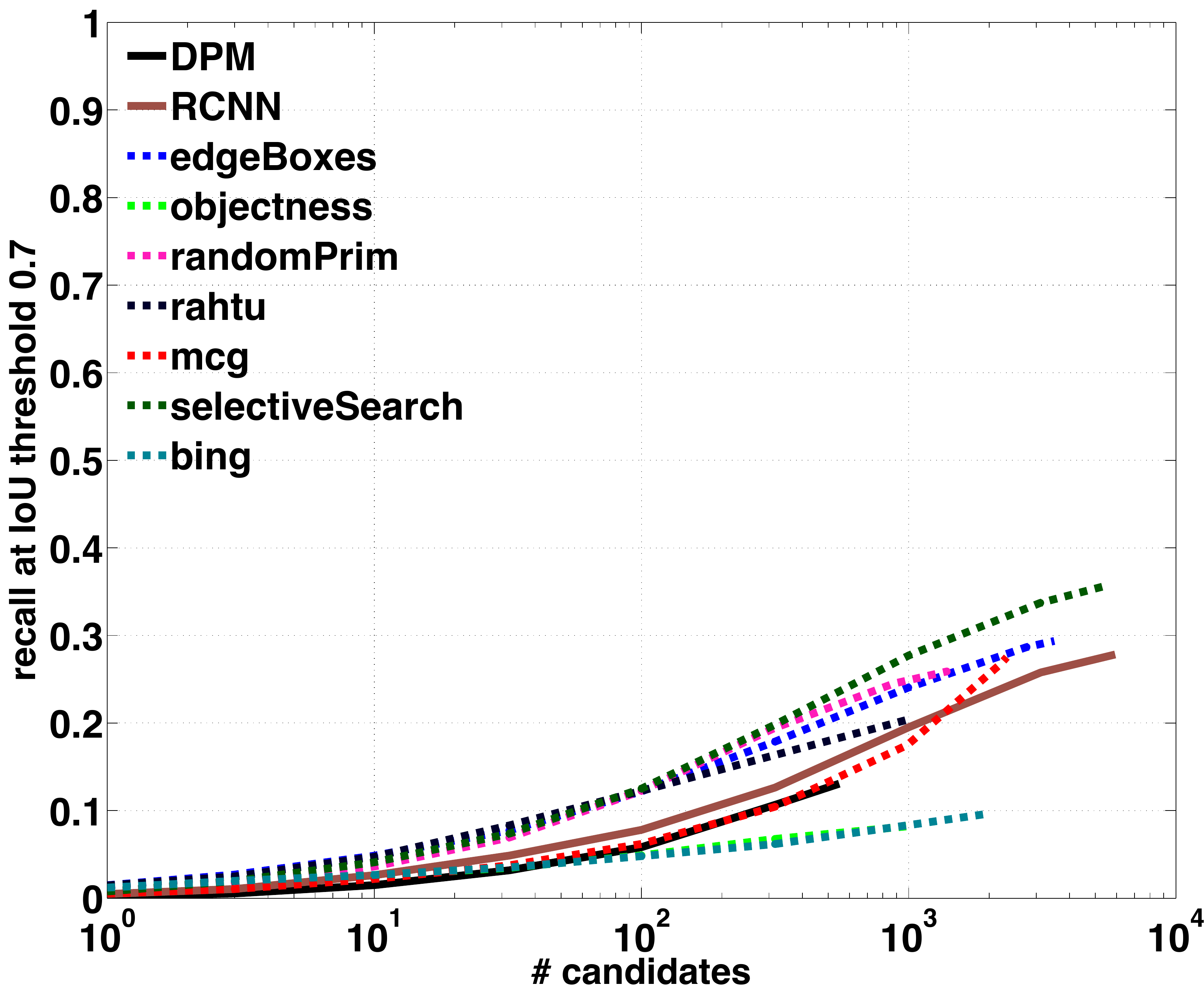}
		\subcaption{\label{60context07}Recall \vs number of proposals at 0.7 IOU for non-PASCAL categories annotated in PASCAL Context dataset} 
	\end{subfigure}
	\label{60cat_context}
	\centering
	\begin{subfigure}[b]{0.3\textwidth}
		\includegraphics[width=1.0\columnwidth]{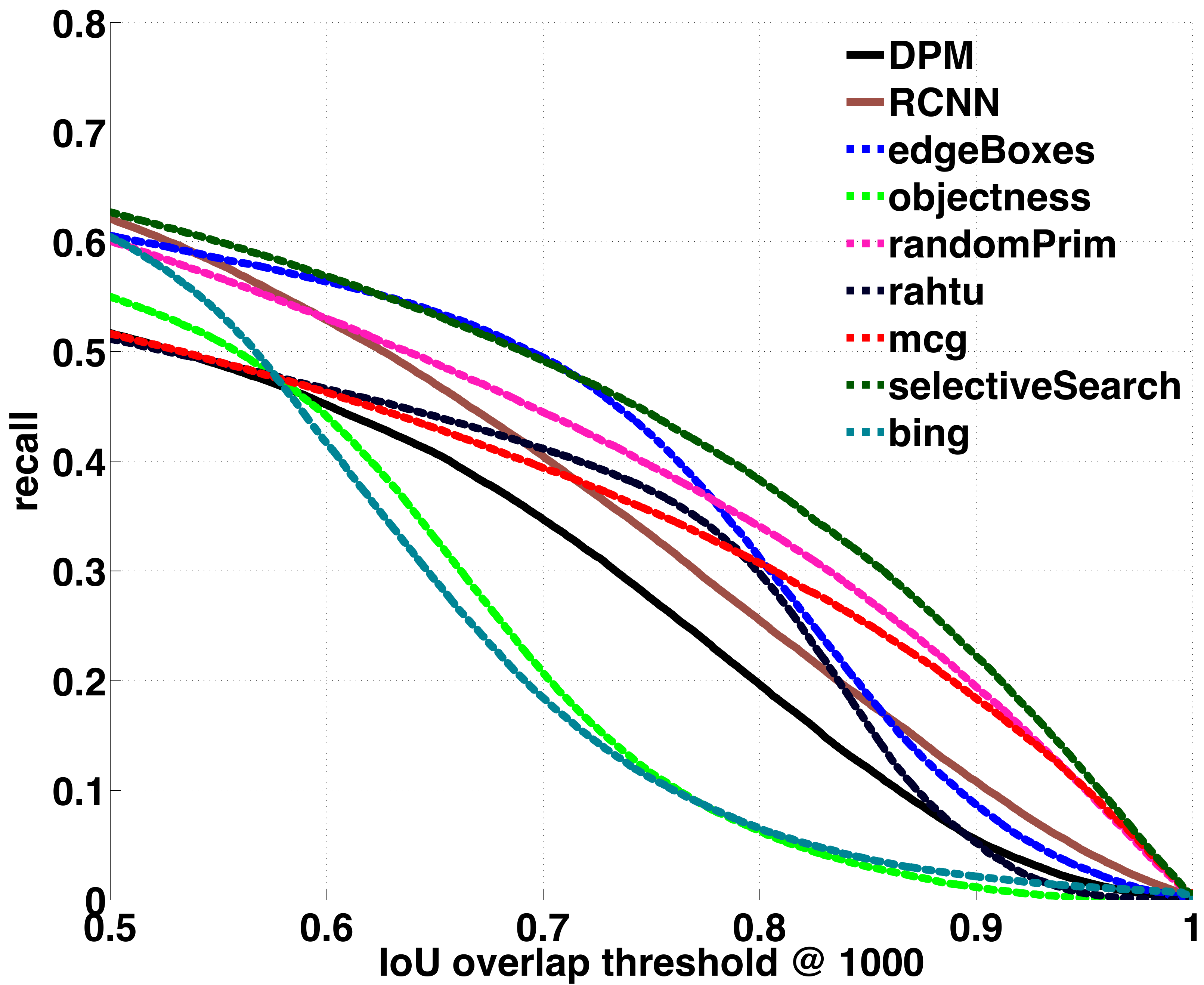}
		\subcaption{\label{80context1000}Recall vs IOU at 1000 proposals for all categories annotated in PASCAL Context dataset}
	\end{subfigure}
	\,\,\,
	\begin{subfigure}[b]{0.3\textwidth}
		\includegraphics[width=1.0\columnwidth]{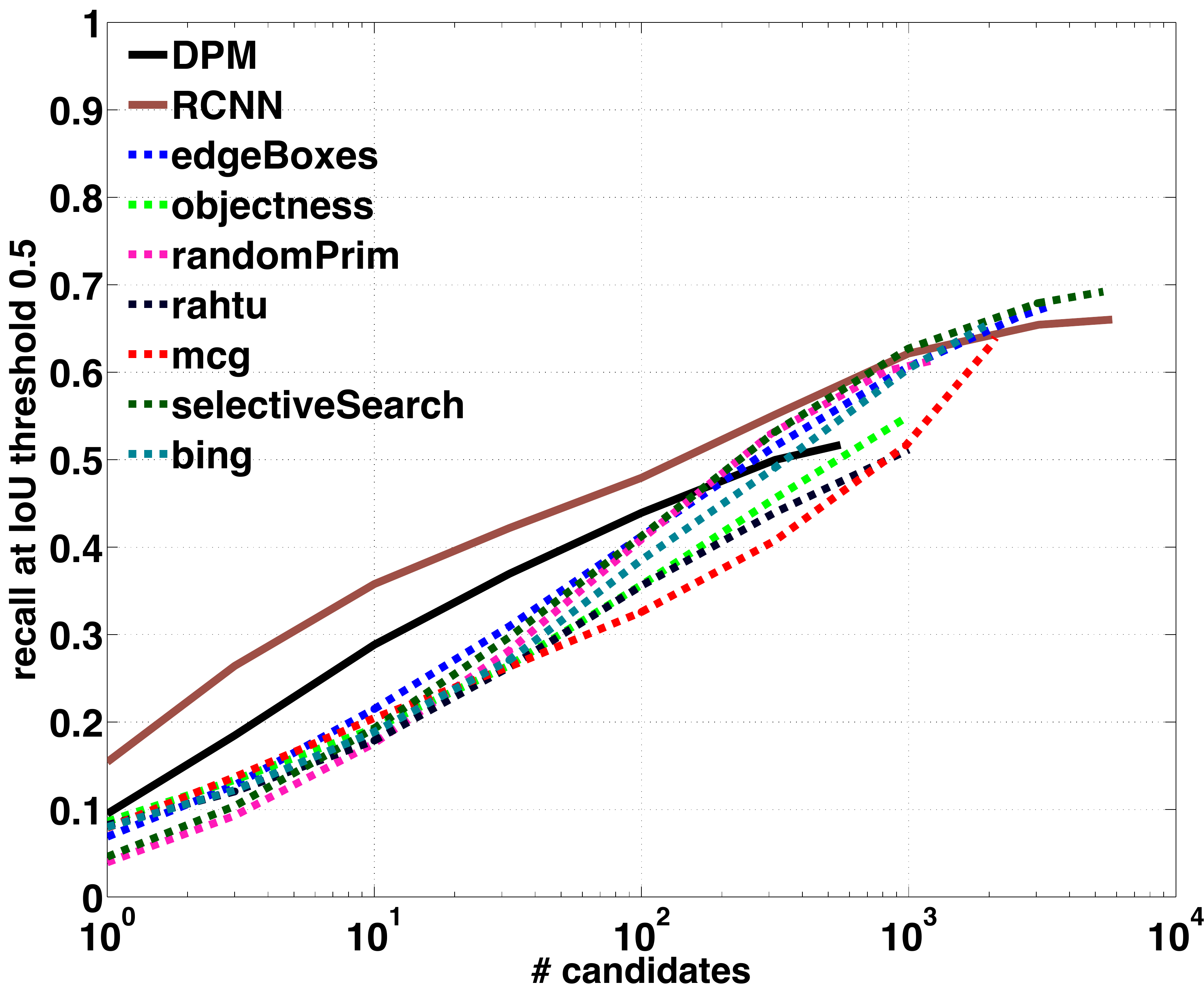}
		\subcaption{\label{80context05}Recall \vs number of proposals at 0.5 IOU for all categories annotated in PASCAL Context dataset} 
	\end{subfigure}
	\,\,\,
	\begin{subfigure}[b]{0.3\textwidth}
		\includegraphics[width=1.0\columnwidth]{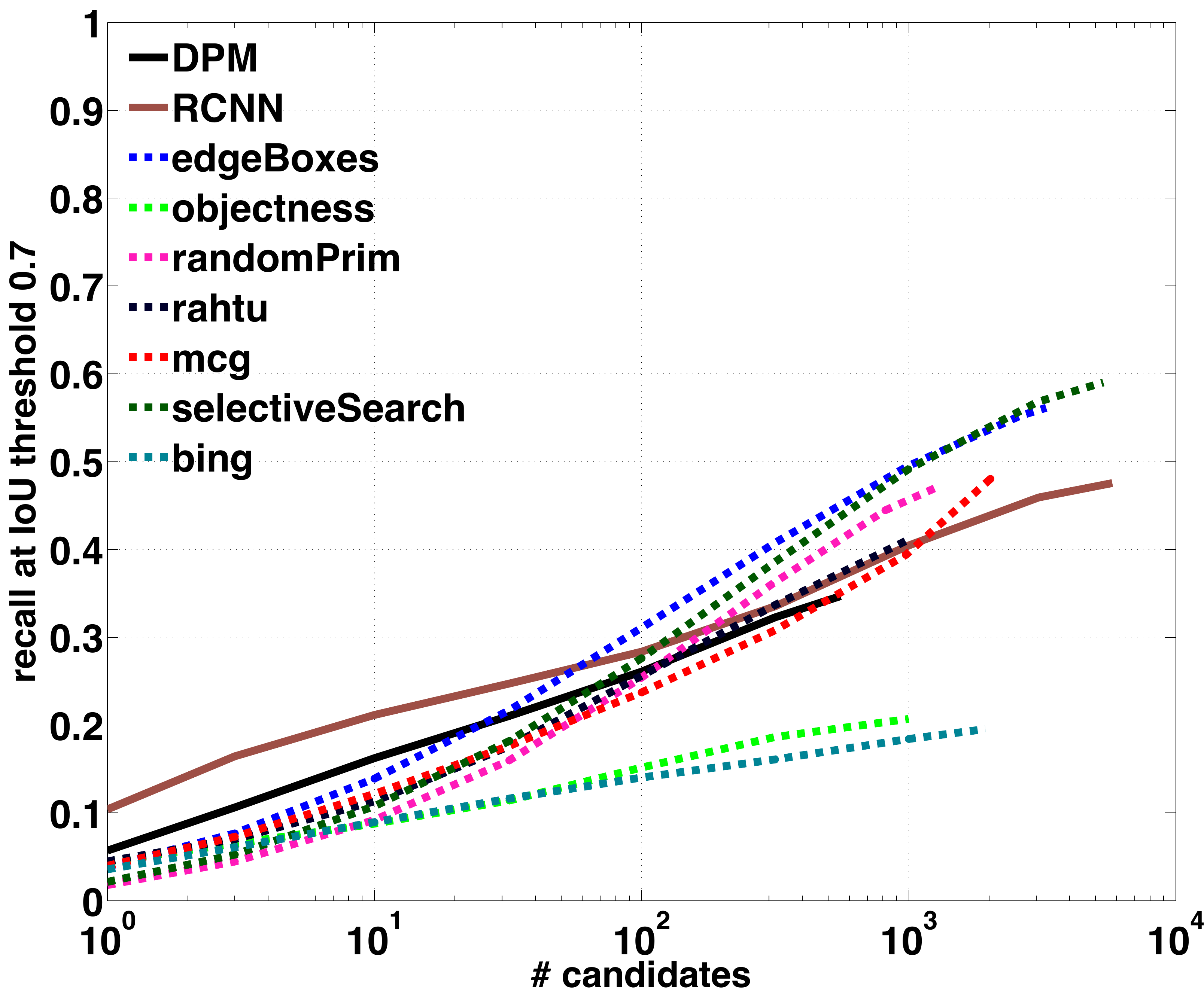}
		\subcaption{\label{80context07}Recall \vs number of proposals at 0.7 IOU for all categories annotated in PASCAL Context dataset} 
	\end{subfigure}
	\caption{\label{allcat_context}Performance of various object proposal methods on different evaluation metrics when evaluated on PASCAL Context dataset}
\end{figure*}
\begin{figure*}
	\centering
	\begin{subfigure}[b]{0.30\textwidth}
		\includegraphics[width=1\columnwidth]{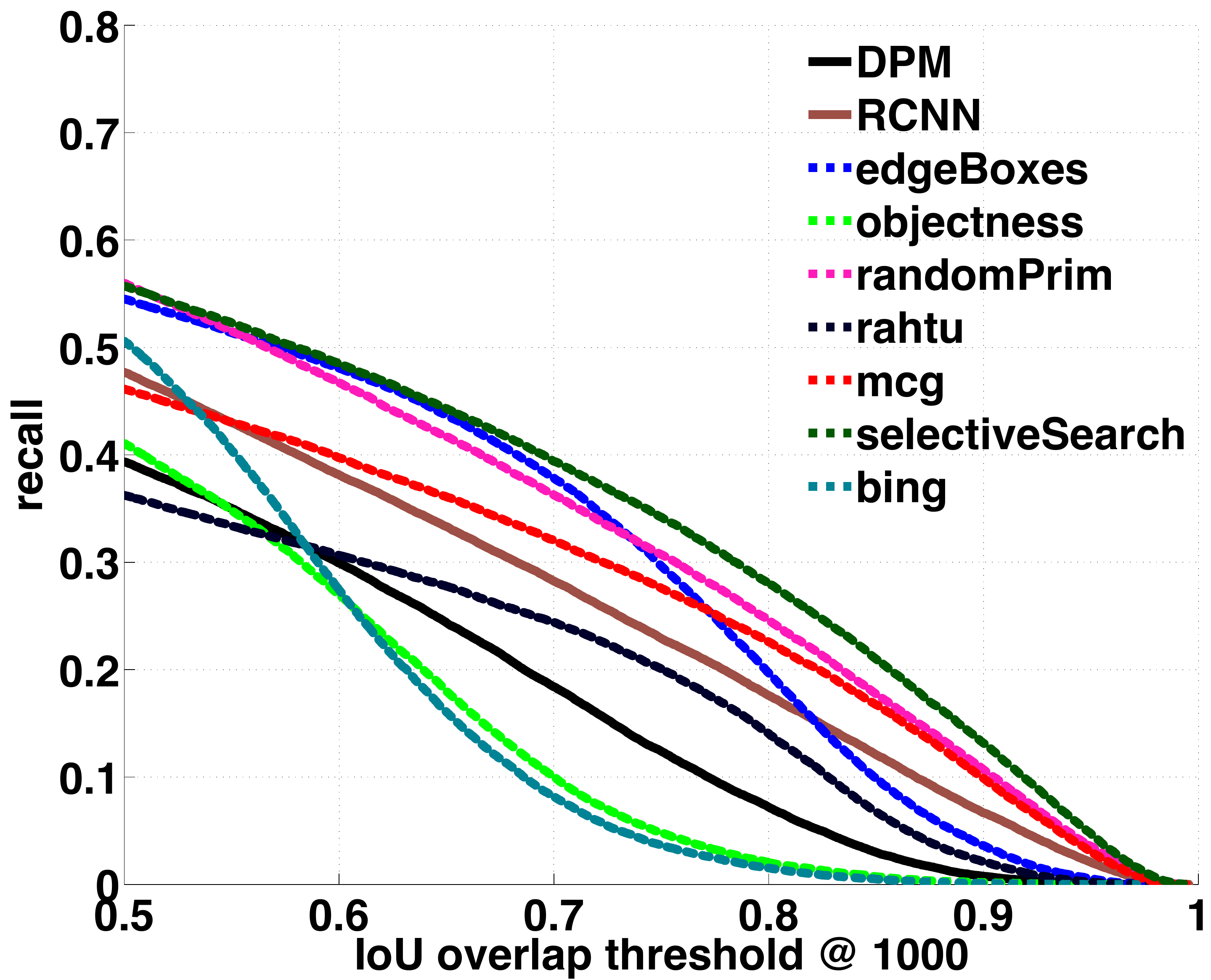}
		\subcaption{\label{nyu21000}Recall vs IOU at 1000 proposals for all categories annotated in the NYU2 dataset}
	\end{subfigure}
	\,\,\,
	\begin{subfigure}[b]{0.30\textwidth}
		\includegraphics[width=1\columnwidth]{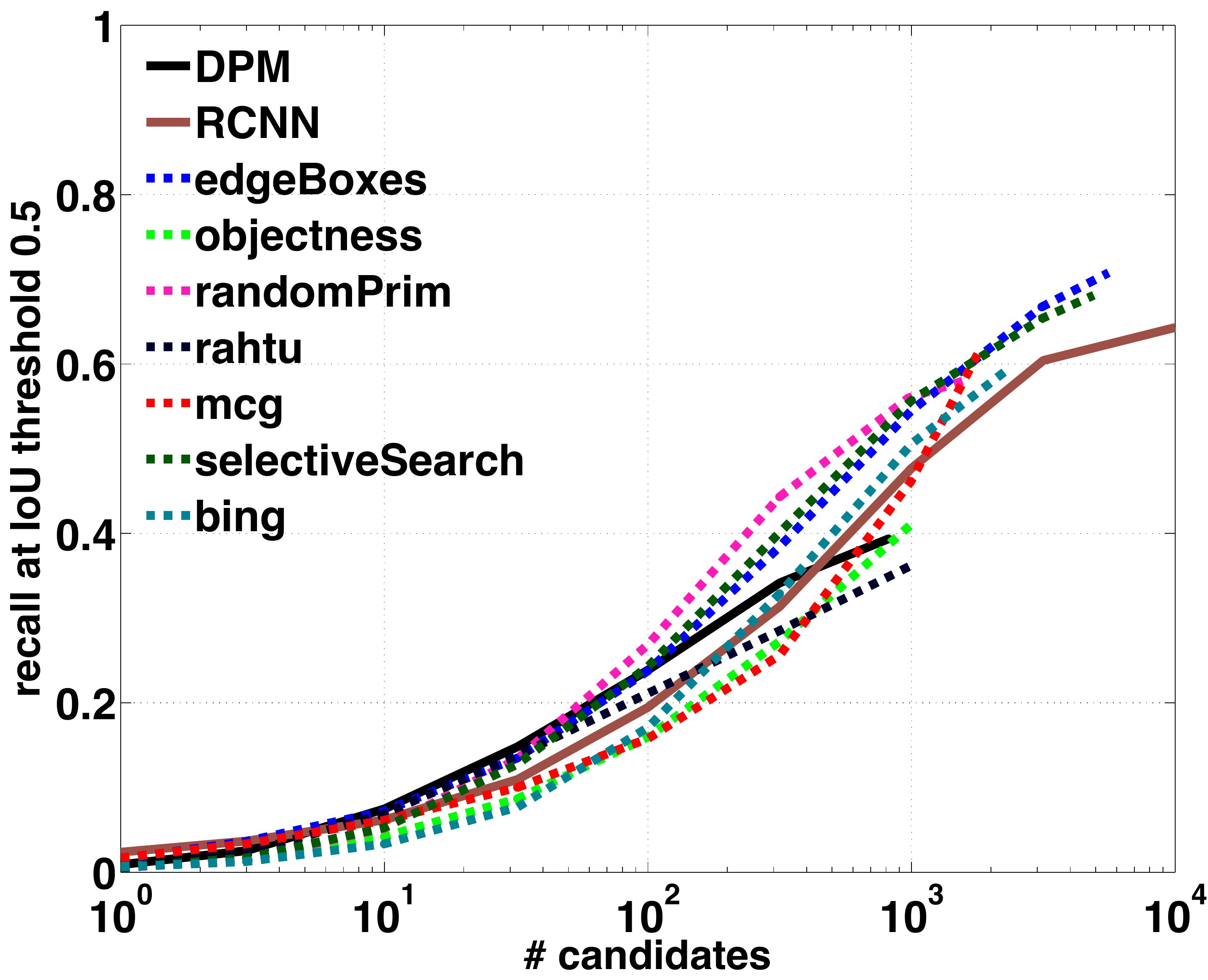}
		\subcaption{\label{nyu205}Recall \vs number of proposals at 0.5 IOU for all categories annotated in the NYU2 dataset} 
	\end{subfigure}
	\,\,\,
	\begin{subfigure}[b]{0.30\textwidth}
		\includegraphics[width=1\columnwidth]{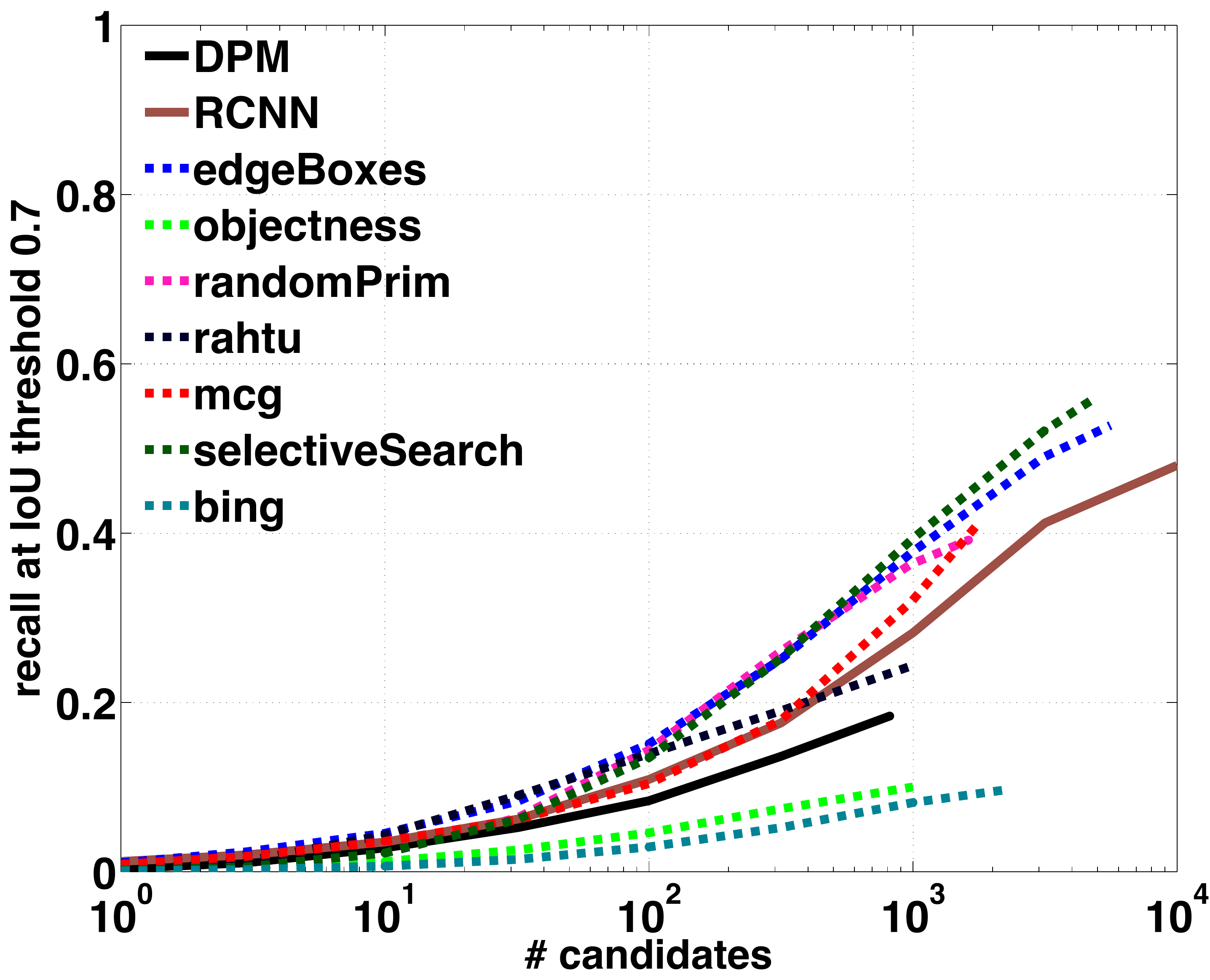}
		\subcaption{\label{nyu207}Recall \vs number of proposals at 0.7 IOU for all categories annotated in the NYU2 dataset} 
	\end{subfigure}

	\caption{\label{nyu2}Performance of various object proposal methods on different evaluation metrics when evaluated on NYU2 dataset containing annotations for all categories}
\end{figure*}
\label{more_results}
\vspace{\sectionReduceTop}
\subsection{Measuring Fine-Grained Recall}
\vspace{\sectionReduceTop}
\begin{figure*}[htb]
	\begin{subfigure}[b]{0.3\textwidth}
		\includegraphics[width=1.3\columnwidth]{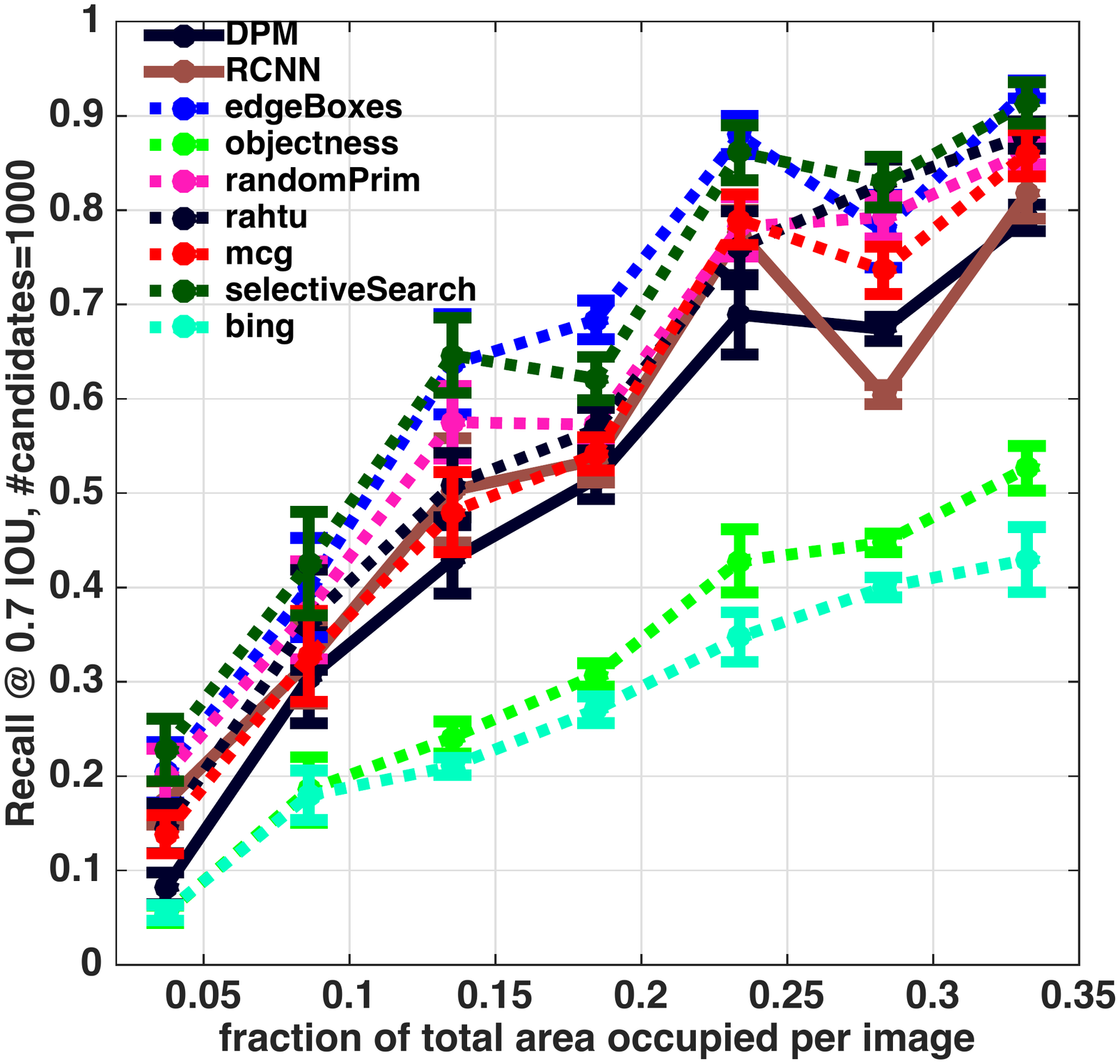}
		\subcaption{Sorted by size.
		\label{newexparea1}}
	\end{subfigure}
	\,\,\,
	\begin{subfigure}[b]{0.3\textwidth}
		\includegraphics[width=1.3\columnwidth]{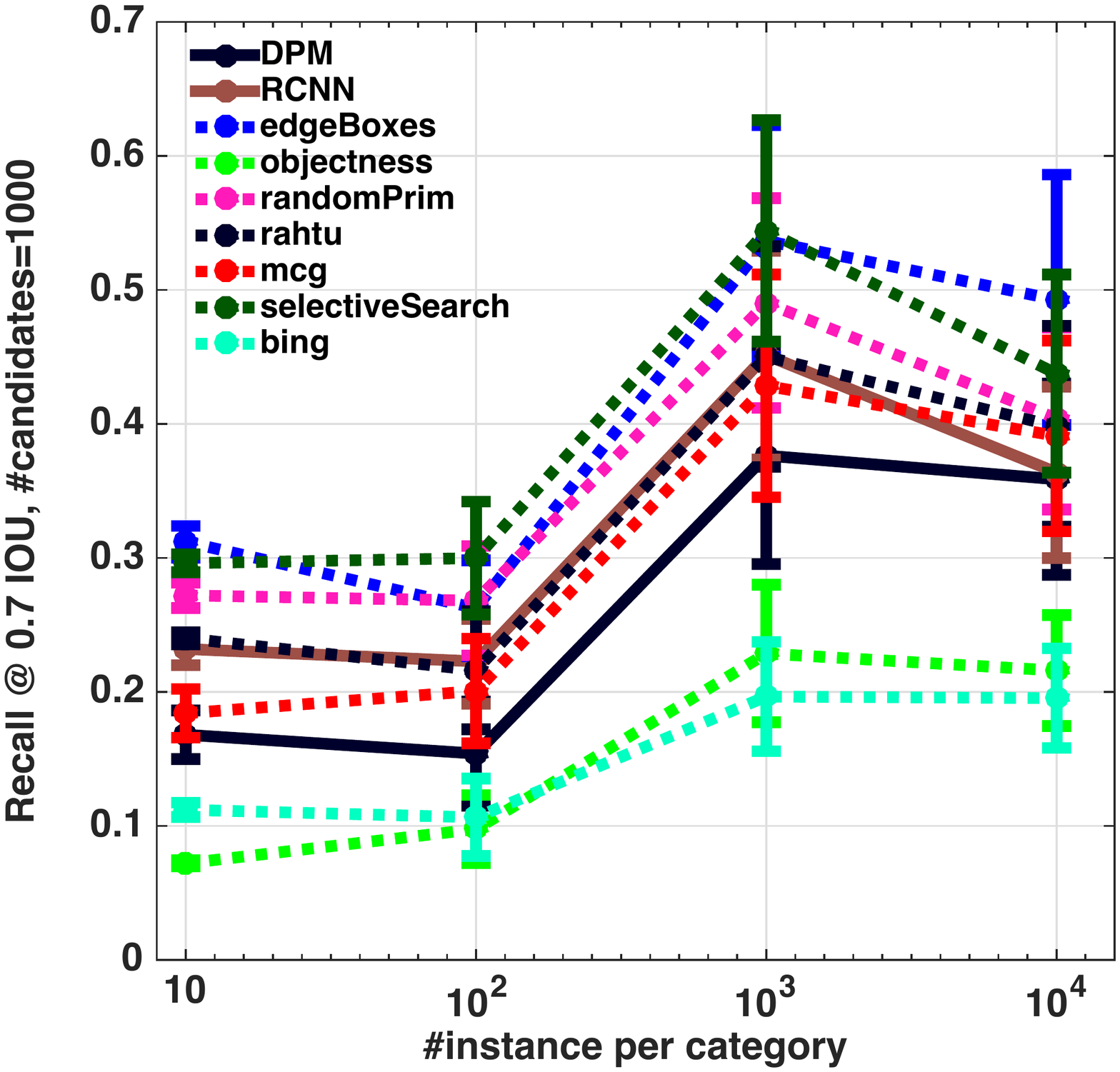}
		\subcaption{Sorted by the number of instances.} 
		\label{newexpinst1} 
	\end{subfigure}
	\,\,\,
	\begin{subfigure}[b]{0.3\textwidth}
		\includegraphics[width=1.3\columnwidth]{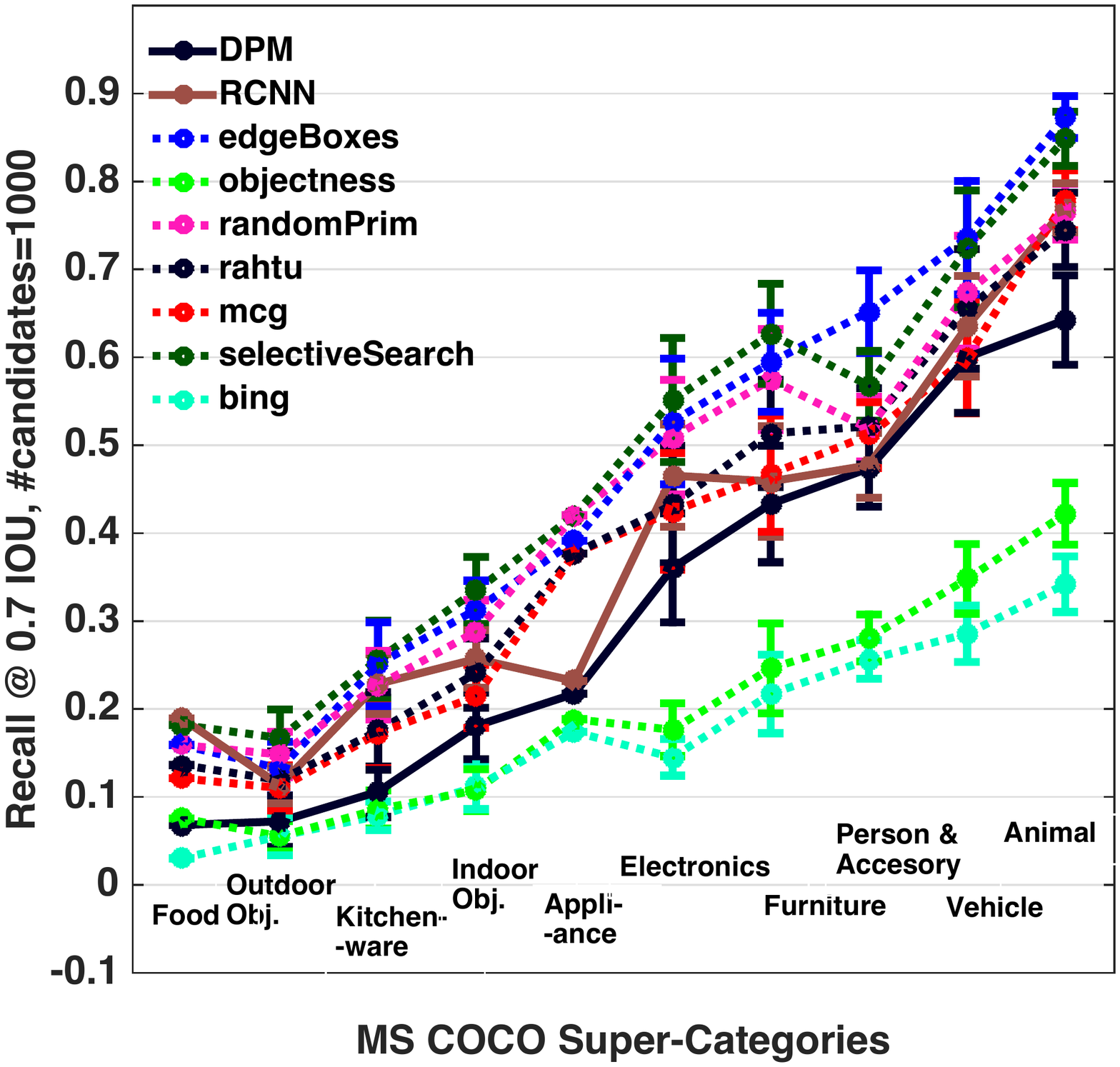}
		\subcaption{MS COCO super-categories.} 
		\label{newexpcoco1} 
	\end{subfigure}
	\caption{Recall at 0.7 IOU for categories sorted/clustered by (a) size, (b) number of instances, and 
		(c) MS COCO `super-categories' evaluated on PASCAL Context.\label{percategoryplot1}\\}
	
	\begin{subfigure}[b]{0.3\textwidth}
		\includegraphics[width=1.3\columnwidth]{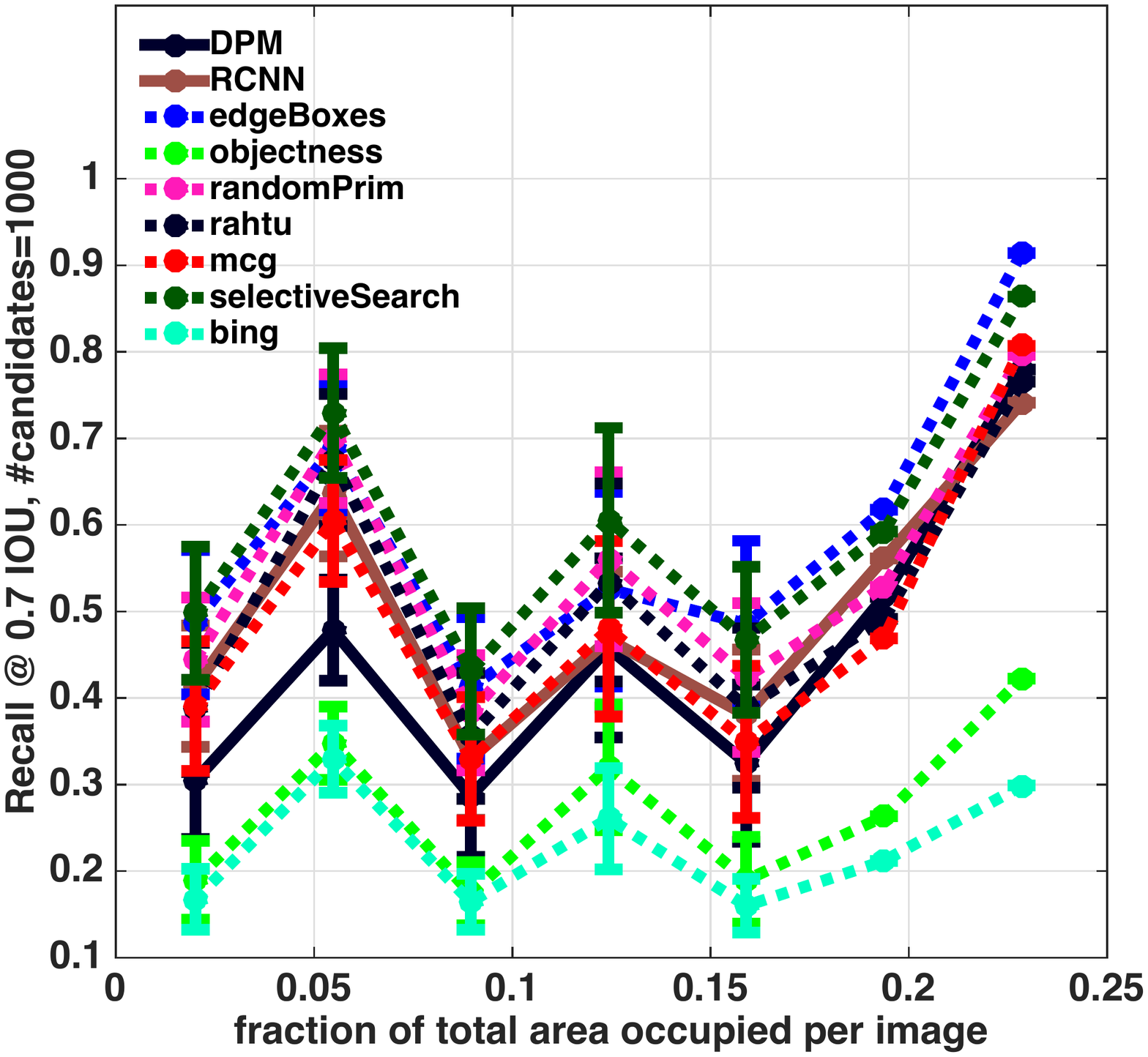}
		\subcaption{{Sorted by size.}}
		\label{newexparea}
	\end{subfigure}
	\,\,\,
	\begin{subfigure}[b]{0.3\textwidth}
		\includegraphics[width=1.3\columnwidth]{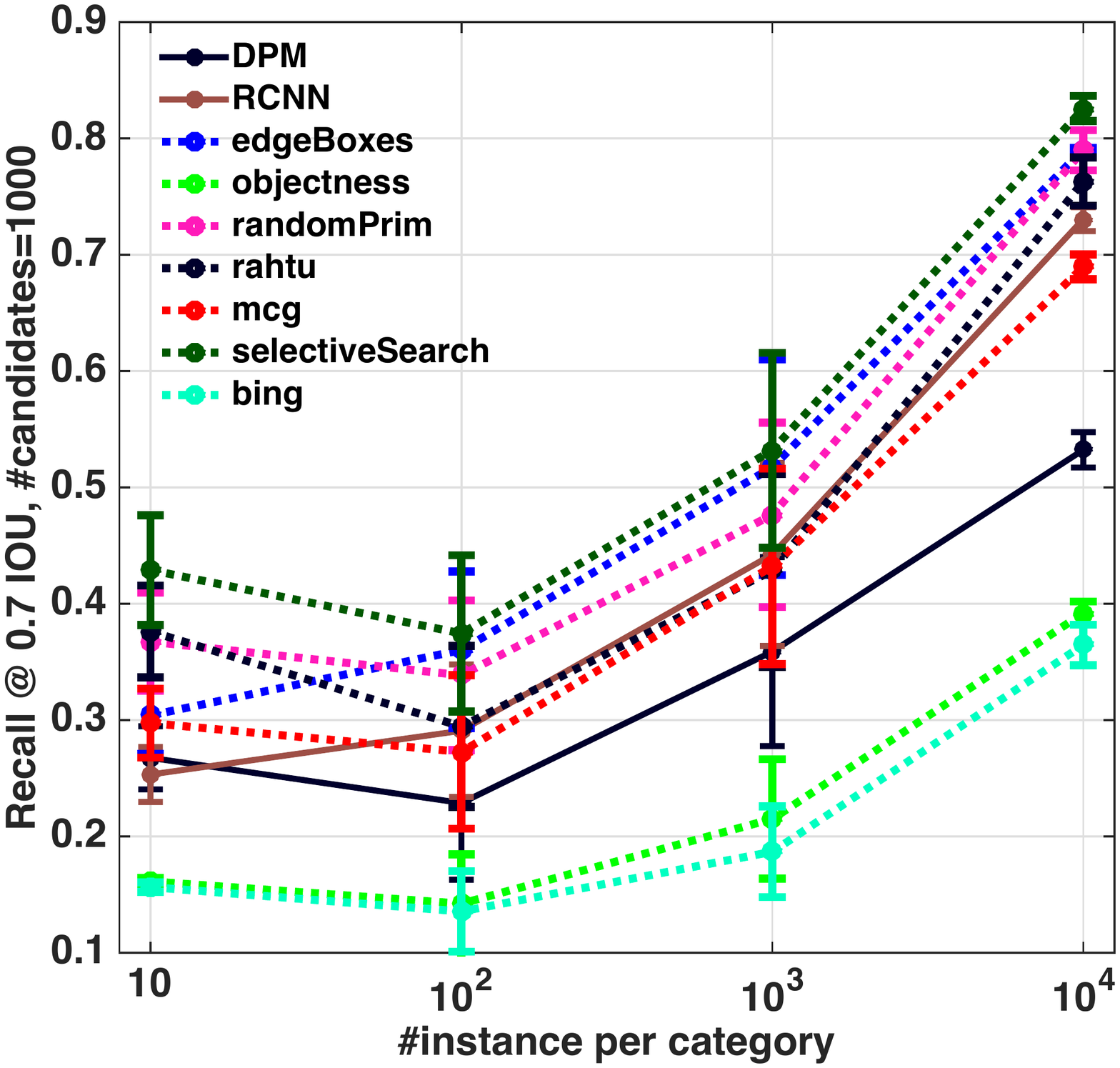}
		\subcaption{{Sorted by the number of instances.}} 
		\label{newexpinst} 
	\end{subfigure}
	\,\,\,
	\begin{subfigure}[b]{0.3\textwidth}
		\includegraphics[width=1.3\columnwidth]{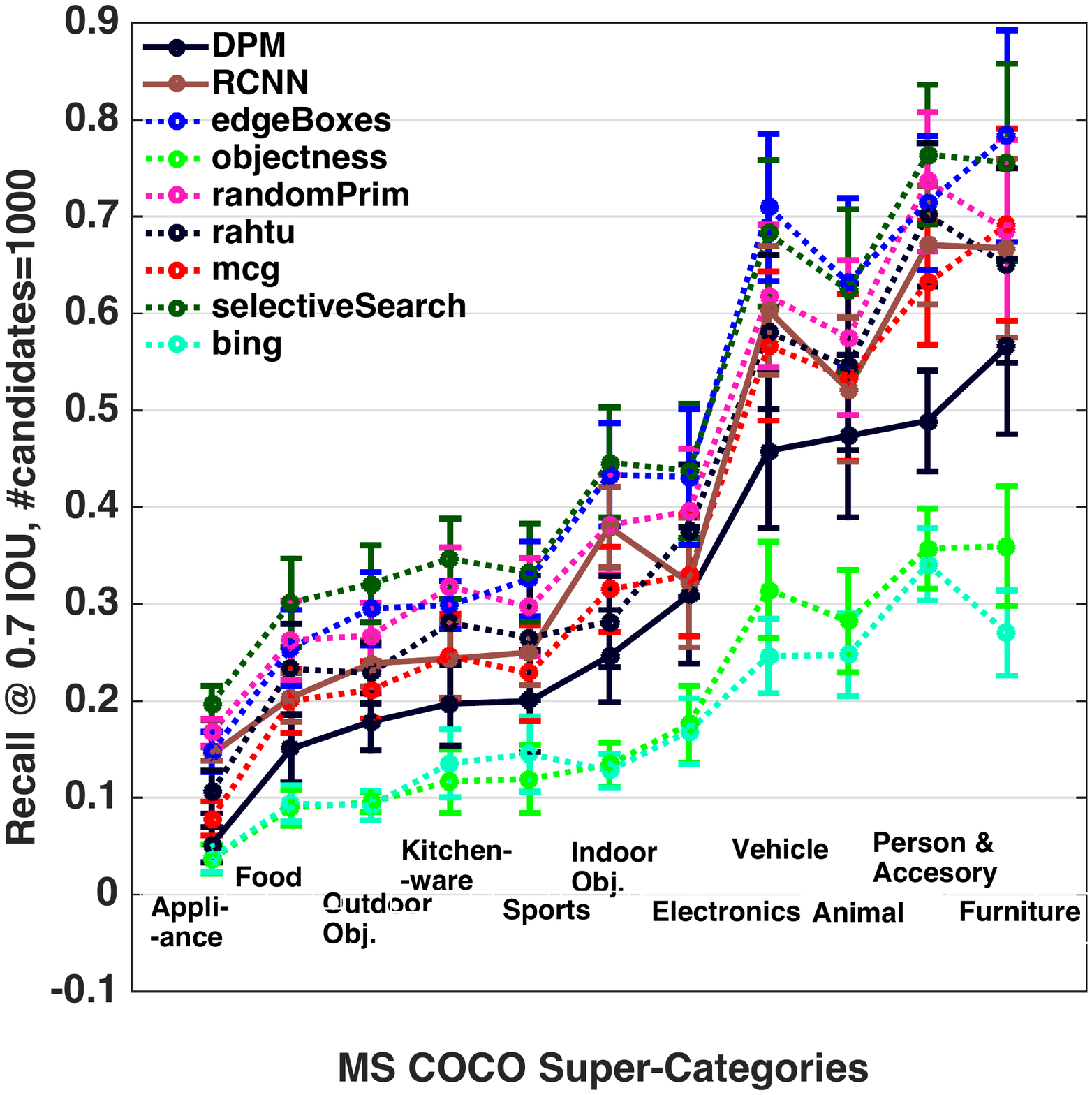}
		\subcaption{{MS COCO super-categories.}} 
		\label{newexpcoco} 
	\end{subfigure}
	\caption{{Recall at 0.7 IOU for categories sorted/clustered by (a) size, (b) number of instances, and 
			(c) MS COCO `super-categories' evaluated on PASCAL Context and MS COCO.}}
	\label{percategoryplot}
\end{figure*}
We also looked at a more fine-grained per-category performance 
of proposal methods and DMPs. 
Fine grained recall can be used to answer if some proposal methods 
are optimized for larger or frequent categories i.e. if they
perform better or worse with respect to different object attributes like area, kinds of objects, etc. It is also easier to 
observe the change in performance of a particular method on frequently occurring category 
\vs rarely occurring category. We performed this experiment on instance level PASCAL Context and MS COCO datasets. We sorted/clustered all categories on the basis of:
\begin{compactitem}
	\item Average size (fraction of image area) of the category, 
	\item Frequency (Number of instances) of the category, 
	\item Membership in `super-categories' defined in MS COCO dataset 
	(electronics, animals, appliance, \etc).
	10 pre-defined clusters of objects of different kind 
	(These clusters are the subset of 11 super-categories defined in MS COCO dataset 
	for classifying individual classes in groups of similar objects.)
\end{compactitem}
Now, we present the plots of  recall for all 80  (20 PASCAL + 60 non-PASCAL) categories for the modified PASCAL Context dataset and MS COCO. Note that the non-PASCAL 60 categories are different for both the datasets.\\
\textbf{Trends:} 
\figref{percategoryplot1} shows the performance of different proposal methods 
and DMPs along each of these dimensions. \\
In \figref{newexparea1}, we see that recall steadily improves 
 perhaps as expected, bigger objects are typically easier to find than smaller objects. 
In \figref{newexpinst1}, we see 
that the recall generally increases as the number of instances increase except for one outlier category. This category was found to be `pole' which appears to be quite difficult to 
recall, since poles are often occluded and have a long elongated shape, it is not surprising that 
this number is pretty low. 
Finally, in \figref{newexpcoco1} we observe that some super-categories (\eg outdoor objects) are hard to recall while others (\eg animal, electronics) are relatively easier to recall. 
It can be seen in \figref{percategoryplot}, the trends on MS COCO are almost similar to PASCAL Context.
\label{percat}
\subsection{Change Log}
This section tracks major changes in the paper.

\textbf{v1:} Initial version.\\
\textbf{v2,v3:} Minor modifications in text.\\
\textbf{v4:} Current version (more details in section 6.1). 


{\footnotesize
\bibliographystyle{ieeetr}
\bibliography{strings,bib_file}
}

\end{document}